\documentclass[conference]{IEEEtran}

\usepackage[T2A]{fontenc}

\usepackage[utf8]{inputenc}

\usepackage[mathcal]{euscript}
\usepackage{booktabs}

\usepackage{graphicx, url}

\usepackage{amsmath, amsfonts, amssymb, amsthm}
\usepackage{mathptmx}
\usepackage{algorithm2e}
\usepackage{subcaption}
\usepackage{caption}
\usepackage{float}
\usepackage{lmodern}

\newcommand{\ex}{\mathop{\mathbb{E}}\nolimits}
\newcommand{\pr}{\mathop{\mathbb{P}}\nolimits}
\newcommand{\Bcal}{\mathcal{B}}
\newcommand{\Hcal}{\mathcal{H}}
\newcommand{\Ncal}{\mathcal{N}}
\newcommand{\Kcal}{\mathcal{K}}
\newcommand{\BigO}{\mathcal{O}}
\newcommand{\Xcal}{\mathcal{X}}
\newcommand{\Ycal}{\mathcal{Y}}
\newcommand{\Zcal}{\mathcal{Z}}
\newcommand{\Real}{\mathbb{R}}
\DeclareUnicodeCharacter{00A0}{~}

\title{Conformalized Kernel Ridge Regression}
\author{
\IEEEauthorblockN{
Burnaev\IEEEauthorrefmark{1}, E. V.,
Nazarov\IEEEauthorrefmark{2}, I. N.}
\IEEEauthorblockA{
\IEEEauthorrefmark{1}IITP RAS, Skoltech, e.burnaev@skoltech.ru}
\IEEEauthorblockA{\IEEEauthorrefmark{2} IITP RAS, ivan.nazarov@iitp.ru}
}

\begin{document}
\maketitle

\begin{abstract}
General predictive models do not provide a measure of confidence in predictions
without Bayesian assumptions. A way to circumvent potential restrictions is to
use conformal methods for constructing non-parametric confidence regions, that
offer guarantees regarding validity. In this paper we provide a detailed description
of a computationally efficient conformal procedure for Kernel Ridge Regression (KRR),
and conduct a comparative numerical study to see how well conformal regions perform
against the Bayesian confidence sets. The results suggest that conformalized KRR
can yield predictive confidence regions with specified coverage rate, which is
essential in constructing anomaly detection systems based on predictive models.
\end{abstract}

\section{Introduction} 
\label{sec:introduction}

In many applied situations, like anomaly detection in telemetry of some equipment,
online filtering and monitoring of potentially interesting events, or power grid
load balancing, it is necessary not only to make optimal predictions with respect
to some loss, but also to be able to quantify the degree of confidence in the obtained
forecasts. At the same time it is necessary to take into consideration exogenous
variables, that in certain ways affect the object of study. 

Practical importance and difficulty of anomaly detection in general spurred a great
deal of research, which resulted in a large volume of heterogeneous approaches and
methods to its solution (\cite{Alestra2014, Burnaev2015, Artemov2015}). There are many
approaches: probabilistic, which rely on approximating the generative distribution
of the observed data (\cite{aggarwal2008, scott2008}), metric-based anomaly detection,
that measure similarity between normal and abnormal observations (\cite{hautamaki2004,
breunig2000, kriegel2009}), predictive modelling approaches, which use the forecast
error to measure abnormality (\cite{augusteijn2002, hawkins2002, hoffmann2007, scholkopf1998}).

Predictive modelling is concerned with recovering an unobserved relation $x\mapsto f(x)$
from a sample $(x_i, y_i)_{i=1}^n$ of noisy observations
\begin{equation} \label{eq:signal_model}
  y_i = f(x_i) + \epsilon_i \,,
\end{equation}
where $\epsilon_i$ is iid with mean zero. One way of assigning a confidence measure
to a prediction $\hat{f}$ is by using assumptions on the geometric structure of the
manifold approximating the sample data which provide estimates of the likelihood of
the prediction error for the test data, \cite{Bernstein2015,Kuleshov2016}. Another
approach is to quantify estimate's, model's and observational uncertainty, by imposing
Bayesian assumptions (eq.~\ref{eq:signal_model}).

Consider Kernel Ridge Regression -- a model that combines ridge regression with the
kernel trick. It learns a function $\hat{f}$ which, given training sample $X = (x_i)_{i=1}^n$,
$X\in \Xcal^{n\times 1}$, solves
\begin{equation*}
  \|y - f(X)\|^2 + \lambda \|f\|^2 \to \min_{f \in \Hcal} \,,
\end{equation*}
with $f(X) = (f(x_i))_{i=1}^n \in \Real^{n\times 1}$, and $y=(y_i)_{i=1}^n \in \Real^{n\times 1}$.
Here $\Hcal$ is the canonical Reproducing Kernel Hilbert Space associated with
a Mercer-type kernel $\Kcal:\Xcal\times \Xcal\mapsto \Real$. The Representer theorem,
\cite{scholkopf2002}, states that the solution $\hat{f}:\Xcal\mapsto\Real$ is of the
form $f(x) = k_x' \beta$, for $k_x : x\mapsto \Phi(x)$ and $\Phi = (\phi(x_i))_{i=1}^n \in \Hcal^{n\times 1}$.
If $e_j$ is the $j$-th unit vector in $\Real^{n\times 1}$, and $K_{XX}$ is the Gram
matrix of $\Kcal$ over $(x_i)_{i=1}^n$, then $f(x_j) = k_{x_j}' \beta = e_j' K_{XX} \beta$
and  $\| f \|^2 = \beta' K_{XX} \beta$. Thus the kernel ridge regression problem is
equivalent to this finite-dimensional convex minimization problem:
\begin{equation*}
  \|y - K_{XX} \beta \|^2 + \lambda \beta' K_{XX} \beta
    \to \min_{\beta\in \Real^{n\times 1}} \,,
\end{equation*}
which yields the optimal weight vector $\hat{\beta}$ and prediction at $x^*\in \Xcal$
given by $\hat{\beta} = (\lambda I_n + K_{XX})^{-1} y$ and $\hat{y}(x^*) = k_{x^*}'\hat{\beta}$,
respectively.

Bayesian Kernel Ridge Regression views the model eq.~\ref{eq:signal_model} as a sample
path of the underlying Gaussian Process, \cite{rasmussen2006}, which makes predictive
confidence intervals readily available. A Gaussian Process $(y_x)_{x\in \Xcal}$, with
mean $m : \Xcal \mapsto \Real$ and covariance kernel  $\Kcal : \Xcal \times \Xcal \mapsto \Real$
is a random process such that for any $n\geq1$ and any $X = (x_i)_{i=1}^n \in \Xcal$
the $n\times 1$ vector $y_X = (y(x_i))_{i=1}^n$ is Gaussian, $y_X \sim \Ncal_n(m_X, K_{XX})$,
where $m_X = (m(x_i))_{i=1}^n$. The conditional distribution of targets in a test sample
$y_{X^*} = (y_{x^*_j})_{j=1}^l$ with respect to the train sample $y_X = (y_{x_i})_{i=1}^n$
is given by
\begin{equation} \label{eq:cond_distr}
  y_{X^*}\vert_{y_X}
    \sim \Ncal_l\bigl(
      m_{X^*} + K_{X^*X} Q_X (y_X - m_X),
      \Sigma_K(X^*)
    \bigr)
    \,,
\end{equation}
where $\Sigma_K(X^*) = K_{X^*X^*} - K_{X^*X} Q_X K_{XX^*}$, $Q_X = \bigl(K_{XX}\bigr)^{-1}$,
and $K_{XX^*} = (\Kcal(x_i, x^*_j))\in \Real^{n\times l}$. Gaussian Process Regression,
or Kriging, generalizes both linear and kernel regression and assumes linearity of the
mean function with respect to $x$ and external factors $h$.

Bayesian KRR assumes a prior on functions $f\sim GP(0, \sigma^2 \Kcal)$ and independent
Gaussian white noise $\epsilon_x\sim \Ncal(0, \sigma^2 \lambda)$ in model~\ref{eq:signal_model},
for $\sigma^2 > 0$. In this setting, eq.~\ref{eq:cond_distr} implies that the distribution
of a yet unobserved target $y_{x^*}$ at $x^*\in \Xcal$, conditional on the train data $(X, y_X)$,
is
\begin{equation} \label{eq:gp_cond_dist}
{y_{x^*}}_{|y_X}
  \sim \Ncal\bigl(\hat{y}_{y_X}(x^*), \sigma^2 \sigma_K^2(x^*)\bigr) \,,
\end{equation}
with $\hat{y}_{y_X}(x^*) = K_X(x^*)' Q_X y_X$, and
\begin{equation*}
\sigma_K^2(x^*) = \lambda + K(x^*,x^*) - K_X(x^*)' Q_X K_X(x^*) \,,
\end{equation*}
where $Q_X = \bigl(\lambda I_n + K_{XX}\bigr)^{-1}$, $K_{XX} = (K(x_i,x_j))_{ij}$,
and $K_X = (K(x_i, \cdot))_{i=1}^n: \Xcal \mapsto \Real^{n\times1}$. Thus, the $1 - \alpha$
confidence interval is thus given by
\begin{equation} \label{eq:gp_conf_int}
\Gamma^\alpha_{y_X}(x^*)
  = \hat{y}_{y_X}(x^*)
  + \sigma \sqrt{\sigma_K^2(x^*)}
  \times [z_{\frac{\alpha}{2}}, z_{1-\frac{\alpha}{2}}]
  \,,
\end{equation}
where $z_\gamma$ is the $\gamma$ quantile of $\Ncal(0, 1)$. Additionally, this version
naturally permits estimation of parameters of the underlying kernel $\Kcal$ through
maximization of the joint likelihood of the train data $(X, y_X)$:
\begin{equation} \label{eq:bkrr_likelihood}
  \mathcal{L}
    = -\frac{n}{2} \log 2\pi
    - \frac{n}{2}\log \sigma^2
    - \frac{1}{2}\log \lvert R_X \rvert
    - \frac{1}{2\sigma^2} y' R_X^{-1} y
    \,,
\end{equation}
where $R_X = \lambda I_n + K_{XX}$, and $K_{XX}$ depends on the hyper-parameters of
$\Kcal$ (shape, precision et c.).
Other approaches to estimating the covariance function's hyper-parameters are reported
in \cite{Burnaev2014}, properties of posterior parameter distribution in Bayesian KRR
are studied in \cite{Zaitsev2013}, methods of estimating Gaussian Process Regression
on large structured datasets are considered in \cite{Belyaev2015, Belyaev2016}, and
the problem of estimating in non-stationary case with regularization is considered in 
\cite{Burnaev2016}.


It is desirable to have distribution-free method that measures confidence of predictions
of a machine learning algorithm. One such method is ``Conformal prediction'' -- an
approach developed in \cite{vovk2005}, which under standard independence assumptions
yields a set in the space of targets, that contains yet unobserved data with a pre-specified
probability. In this study, we provide empirical evidence supporting the claim that
when model assumptions do hold, the conformal confidence sets, constructed over the
Kernel Ridge Regression with isotropic Gaussian kernel do not perform worse than
the prediction confidence intervals of a Bayesian version of the KRR. The paper is
structured as follows: in section~\ref{sec:conformal_prediction} a concise overview
of what conformal prediction is and what is required to construct such kind of confidence
predictor is given. Section~\ref{sec:conformalized_krr} describes the particular steps
needed to build a conformal predictor atop the kernel ridge regression. The main
empirical study is reported in section~\ref{sec:numerical_study}, where we study the
properties of the predictor in a batch learning setting for a KRR with specific kernel. 


\section{Conformal prediction} 
\label{sec:conformal_prediction}

Conformal prediction is a distribution-free technique designed to yield a statistically
valid confidence sets for predictions made by machine learning algorithms. The key
advantage of the method is that it offers coverage probability guarantees under
standard IID assumptions, even in cases when assumptions of the underlying prediction
algorithm fail to be satisfied. The method was introduced in \cite{vovk2005} for online
supervised and unsupervised learning.

Let $\Zcal$ denote the object-target space $\Xcal \times \Ycal$. At the core of a
conformal predictor is a measurable map $A: \Zcal^*\times \Zcal \mapsto \Real$, a
Non-Conformity Measure (NCM), which quantifies how much different $z_{n+1} \in \Zcal$
is relative to a sample $Z_{:n} = (z_i)_{i=1}^n\in\Zcal$. A conformal predictor over
$A$ is a procedure, which for every sample $Z_{:n}$, a test object $x_{n+1} \in \Xcal$,
and a level $\alpha \in (0,1)$, gives a confidence set $\Gamma_{Z_{:n}}^\alpha(x^*)$
for the target value $y_{n+1}$:
\begin{equation} \label{eq:conf_pred_set}
  \Gamma_{Z_{:n}}^\alpha(x_{n+1})
    = \bigl\{ y\in \Ycal \,:\, p_{Z_{:n}}(\tilde{z}^y_{n+1}) \geq \alpha \bigr\} \,,
\end{equation}
where $\tilde{z}^y_{n+1} = (x_{n+1}, y)$ a synthetic test observation with target
label $y$. The function $p:\Zcal^*\times (\Xcal\times \Ycal)\mapsto [0,1]$ measures
the likelihood of $\tilde{z}$ based on its non-conformity with $Z_{:n}$, and is
\begin{equation} \label{eq:conf_p_value}
  p_{Z_{:n}}(\tilde{z})
    = (n+1)^{-1} \bigl\lvert\{ i \,:\,
      \eta_i^{\tilde{z}} \geq \eta_{n+1}^{\tilde{z}} \}\bigr\rvert \,,
\end{equation}
where $i=1,\ldots, n+1$, and $\eta_i^{\tilde{z}} = A(S^{\tilde{z}}_{-i}, S^{\tilde{z}}_i)$
-- the non-conformity of the $i$-th observation with respect to the augmented sample
$S^{\tilde{z}} = (Z_{:n}, {\tilde{z}}^y_{n+1}) \in \Zcal^{n+1}$. For any $i$, $S^{\tilde{z}}_i$
is the $i$-th element of the sample, and $S^{\tilde{z}}_{-i}$ is the sample with the $i$-th
observation omitted.


For every possible value $z$ of an object $Z_{n+1}$ the conformal procedure tests
$H_0: Z_{n+1} = z$, and then inverts the test to get a confidence region. The hypothesis
tests are designed to have a fixed empirical type-I error rate $\alpha$ based on
the observed sample $Z_{:n}$ and hypothesized $z$.

In \cite{vovk2005}, chapter 2, is has been shown, that for sequences of iid examples
$(z_n)_{n \geq1} \sim P$, the coverage probability of the prediction set $\Gamma^\alpha$,
\ref{eq:conf_pred_set}, is at least $1-\alpha$ and successive errors are independent
in online learning and prediction setting. The procedure guarantees unconditional validity: 
for any $\alpha \in (0,1)$
\begin{equation} \label{eq:conservative_coverage}
  \pr_{Z_{:n}\sim P} \bigl(
    y_n \notin \Gamma^\alpha_{Z_{:(n-1)}}(x_n)
  \bigr) \leq \alpha \,,
\end{equation} 
where $(x_n, y_n) = z_n$.
Intuitively, the event $y_n \notin \Gamma^\alpha_{Z_{:(n-1)}}(x_n)$ is equivalent
to $\eta_n = A(Z_{-n}, Z_n)$ being among the largest $\lfloor n\alpha\rfloor$ values
of $\eta_i = A(Z_{-i}, Z_i)$, which is equal to $\frac{\lfloor n\alpha\rfloor}{n}$,
due to independence of $Z_{:n}$ (for a rigorous proof see \cite{vovk2005}, ch.~8).

The choice of \textbf{NCM} affects the size of the confidence sets and the computational
burden of the conformal procedure. In the general case computing eq.~\ref{eq:conf_pred_set}
requires exhaustive search through the target space $\Ycal$, which is infeasible in general
regression setting. However, for specific non-conformity measures it is possible to
come up with efficient procedures for computing the confidence region as demonstrated
in \cite{vovk2005} and sec.~\ref{sec:conformalized_krr} of this work.


\section{Conformalized kernel ridge regression} 
\label{sec:conformalized_krr}

In this section we describe the construction of confidence regions of the conformal
procedure eq.~\ref{eq:conf_pred_set} for the case of the non-conformity measures
based on kernel ridge regression. We consider two NCMs defined in terms of regression
residuals: the one used in constructing a ``Ridge Regression Confidence Machine'',
proposed in \cite{vovk2005}, chapter 2, and ``two-sided'' NCM, proposed in \cite{burnaevV14}.

\subsection{Residuals} 
\label{sub:residuals}

In each NCM it is possible to use any kind of prediction error, but we focus on two:
the in-sample and \textbf{l}eave-\textbf{o}ne-\textbf{o}ut (or deleted) residuals.
Consider a sample $(X, y) = (x_i, y_{x_i})_{i=1}^n$, and for any $i=1\ldots, n$ put
$X = (X_{-i}, x_i)$, and $y = (y_{-i}, y_i)$. In-sample residuals, $\hat{r}_{\text{in}}(X, y)$,
are defined for each $i$ as
\begin{equation} \label{eq:ins_resid}
  e_i' \hat{r}_{\text{in}}(X, y) = y_i - \hat{y}_{|(X, y)}(x_i) \,,
\end{equation}
and LOO $\hat{r}_{\text{loo}}(X, y)$ are given by
\begin{equation} \label{eq:loo_resid}
  e_i' \hat{r}_{\text{loo}}(X, y) = y_i - \hat{y}_{|(X_{-i}, y_{-i})}(x_i) \,,
\end{equation}
where $\hat{y}_{|(X, y)}$ and $\hat{y}_{|(X_{-i}, y_{-i})}$ denote predictions of
a KRR fit on the whole sample $(X, y)$, and a sample $(X_{-i}, y_{-i})$ with the
$i$-th observation knocked-out, respectively. For any $i$ the residuals are related
by
\begin{equation}
  e_i' \hat{r}_{\text{in}}(X, y)
    = \lambda m_i^{-1} e_i' \hat{r}_{\text{loo}}(X, y)
    \,,
\end{equation}
where $\lambda m_i^{-1} = \lambda e_i'Q_X e_i$ is the KRR ``leverage'' score of
the $i$-th observation, and
\begin{equation} \label{eq:krr_leverage}
  m_i = \lambda + \Kcal(x_i, x_i) - k_{-i}(x_i)' Q_{-i} k_{-i}(x_i) \,,
\end{equation}
with $k_{-i}(x_i)$ -- the $n-1\times 1$ vector of $(\Kcal(x_j, x_i))_{i\neq j}$,
$Q_{-i} = (K_{-i} + \lambda I_{n-1})^{-1}$, and $K_{-i}$ is the Gram matrix of
the kernel $\Kcal$ over subsample $X_{-i}$.


\subsection{Ridge Regression Confidence Machine} 
\label{sub:ridge_regression_confidence_machine}

In this section we describe a conformal procedure for the NMC proposed in \cite{vovk2005},
chapter 2, and focus on its ``in-sample'' version, bearing in mind that residuals
(\ref{eq:loo_resid}) and (\ref{eq:ins_resid}) are interchangeable.

The Ridge Regression Confidence Machine (RRCM) constructs an non-conformity measure
from the absolute value of the regression residual: the ``in-sample'' NCM, $A_{\text{in}}$,
is given by
\begin{equation} \label{eq:ins_ncm}
  A_{\text{in}}\bigl((X_{-i}, y_{-i}), (x_i, y_i)\bigr) = |e_i' \hat{r}_{\text{in}}(X, y)| \,,
\end{equation}
and the ``LOO'' NCM, $A_{\text{loo}}$ is defined similarly using eq.~\ref{eq:loo_resid}.
For the NCM $A$ the $1 - \alpha$ conformal confidence interval for the $n$-th observation
is given by
\begin{equation} \label{eq:conf_ci}
  \Gamma_{X_{-n}, y_{-n}}^\alpha(x_n)
    = \bigl\{ z\in \Real \,:\, p_n\bigl((X, \tilde{y}_n^z)\bigr) \geq \alpha \bigr\}
    \,,
\end{equation}
where $\tilde{y}_i^z = (y_{-i}, z)$ -- the augmented target sample $y$ with the
$i$-th value replaced by $z$. The ``conformal likelihood'' of the $i$-th observation
in some sample $(X, y)$ is given by
\begin{equation*}
  p_j\bigl((X, y)\bigr)
    = n^{-1} \bigl\lvert \bigl\{
        j = 1,\ldots, n \, : \,
        \eta_j \geq \eta_i
    \bigr\} \bigr\rvert
    \,,
\end{equation*}
for $\eta_i = A\bigl((X_{-i}, y_{-i}), (x_i, y_i)\bigr)$.

Efficient construction of the confidence set for the NCM (\ref{eq:ins_ncm}) for
in-sample (and deleted) residuals relies on linear dependence on the target of the
$n$-th observation:
\begin{equation} \label{eq:krr_in_resid}
  \hat{r}_i^z
    = e_i' \hat{r}_{\text{in}}(X, \tilde{y}_n^z)
    = \lambda c_i + \lambda b_i z
    \,,
\end{equation}
with $c_i = e_i' C_{-n}\bigl((X, y), x_n\bigr)$ and $C_{-n}\bigl((X, y), x_n\bigr)$
given by
\begin{equation*}
  \begin{pmatrix} Q_{-n} y_{-n} \\ 0 \end{pmatrix}
    - B_{-n}(x_n) K_{-n}(x_n)' Q_{-n} y_{-n}
    \,,
\end{equation*}
where $0$ is scalar and the vector $B_{-n}(x_n)\in\Real^{n\times 1}$ is
\begin{equation} \label{eq:krr_in_resid_B}
  B_{-n}(x_n)
    = \begin{pmatrix} - Q_{-n} K_{-n}(x_n) \\ 1 \end{pmatrix} m_n^{-1}
    \,.
\end{equation}
Since absolute values of the residuals are compared, it is possible to consistently
change the signs of each element of $C$ and $B$ to ensure that $e_i'B\geq 0$ for
all $i$.

The conformal p-value for $(x_n, y)$, eq.~\label{eq:conf_p_value}, is can be re-defined in terms
of regions $S_i = \{z\in\Real\,:\, |\hat{r}_i^z| \geq |\hat{r}_n^z|\}$, for $i=1,\ldots, n$:
\begin{equation} \label{eq:rrcm_conf_p_value}
  p_{X_{-n}, y_{-n}}(x_n, y) = n^{-1} \bigl\lvert\{ i \,:\, y \in S_i \}\bigr\rvert \,.
\end{equation}
These regions are either closed intervals, complements of open intervals, one-side
closed half-rays in $\Real$, depending on the values of $C$ and $B$. In particular,
with $p_i$ and $q_i$ denoting $-\frac{c_i+c_n}{b_i+b_n}$ and $\frac{c_i-c_n}{b_n-b_i}$,
respectively (whenever each is defined), each region $S_i$ has one of the following
representations:
\begin{enumerate}
  \item $b_i=b_n=0$: $S_i = \Real$ if $|c_i| \geq |c_n|$, or $S_i = \emptyset$
  otherwise;
  \item $b_n = b_i > 0$: $S_i$ is either $(-\infty, p_i]$ if $c_i < c_n$, $[p_i, +\infty)$ if
  $c_i > c_n$, or $\Real$ otherwise;
  \item $b_n > b_i \geq 0$: $S_i$ is either $[p_i, q_i]$ if $c_i b_n \geq c_n b_i$,
  or $[q_i, p_i]$ otherwise;
  \item $b_i > b_n \geq 0$: $S_i$ is $\Real\setminus (q_i, p_i)$ when $c_i b_n \geq c_n b_i$,
  or $\Real\setminus (p_i, q_i)$ otherwise.
\end{enumerate}
Let $P$ and $Q$ be the sets of all well-defined $p_i$ and $q_i$ respectively, and
let $(g_j)_{j=0}^{J+1}$ enumerate distinct values of $\{\pm\infty\} \cup P \cup Q$,
so that $g_j < g_{j+1}$ for all $j$. Then the confidence region is a closed subset
of $\Real$ constructed from sets $G^m_j = [g_j, g_{j+m}]\cap \Real$ for $m=0, 1$:
\begin{equation} \label{eq:rrcm_conf_ci}
  \Gamma_{X_{-n}, y_{-n}}^\alpha(x_n)
    = \bigcup_{m\in\{0,1\}} \bigcup_{j\,:\, N^m_j \geq n \alpha} G^m_j
    \,,
\end{equation}
where $N^m_j = |\{i \,:\, G^m_j \subseteq S_i\}|$ is the coverage frequency of $G^m_j$,
eq.~\ref{eq:rrcm_conf_p_value}. In general, the resulting confidence set might contain
isolated singletons $G^0_j$.

This set, can be constructed efficiently in $\BigO(n \log{} n)$ time with $\BigO(n)$
memory footprint. Indeed, it is necessary to sort at most $J\leq 2n$ distinct endpoints
of $G_j$, then locate the values $p_i$ and $q_i$ associated with each region $S_i$
($\BigO(n \log{} n)$). Then, since the building blocks $G^m_j$ of $\Gamma^\alpha$
are either singletons ($m=0$), or intervals made up from adjacent singletons ($m=1$),
coverage numbers $N^m_j$ can be computed in at most $\BigO(n)$ time.


\subsection{Kernel two-sided confidence predictor} 
\label{sub:kernel_crr}

Another possibility is to use the two-sided conformal procedure, proposed in \cite{burnaevV14}.
The main result of that paper is that under relaxed Bayesian Ridge Regression assumptions
if a sequence $(x_n)_{n\geq1}\in\Xcal$ is i.i.d. with an non-singular second moment
matrix $\ex x_1x_1' \succeq 0$, then for all sufficiently large $n$ the conformal
confidence regions that lose little efficiency (the upper endpoints of the Bayesian
and conformal prediction intervals deviate as much as $\mathcal{O}_p\bigl(n^{-\frac{1}{2}}\bigr)$).

The ``two-sided'' procedure of \cite{burnaevV14}, denoted by \textbf{CRR} for short,
uses a conformity measure
\begin{equation} \label{eq:crr_ncm}
  A(Z_{-i}, Z_i)
    = \bigl\lvert\{j\,:\, \hat{r}_j \geq \hat{r}_i \} \bigr\rvert \wedge
       \bigl\lvert\{j\,:\, \hat{r}_j \leq \hat{r}_i \} \bigr\rvert \,,
\end{equation}
where $(\hat{r}_i)_{i=1}^n$ are the in-sample ridge regression residuals.
In that paper it was also shown that for any $\alpha \in (0,1)$ the confidence
region $\Gamma^\alpha$ produced by CRR procedure for the conformity measure in
eq.~\ref{eq:crr_ncm} is equivalent to the intersection of confidence sets yielded
by conformal procedures with non-conformity measures given by $\eta_i = \hat{r}_i$
and $\eta_i = -\hat{r}_i$ at significance levels $\frac{\alpha}{2}$. Individually,
these NCMs define a \textbf{upper} and \textbf{lower} CRR sets respectively, and
together constitute a ``two-sided'' conformal procedure. Confidence regions based
on this NCM, much like RRCM, can use any kind of residual: leave-one-out, or in-sample.

For the upper CRR the regions $U_i = \{z\in\Real\,:\, \hat{r}_i^z \geq \hat{r}_n^z\}$,
$i=1,\ldots, n$, are either empty, full $\Real$ or one-side closed half-rays. Since
$\hat{r}_i^z = \lambda c_i + \lambda b_i z$, $U_i$ takes one of the following forms:
\begin{enumerate}
  \item $b_i=b_n$: $U_i = \Real$ if $c_i\geq c_n$, and $\emptyset$ otherwise;
  \item $b_i\neq b_n$: $U_i = [q_i, +\infty)$ if $b_i>b_n$, or
  $U_i = (-\infty, q_i]$ otherwise;
\end{enumerate}
with $q_i = \frac{c_i-c_n}{b_n-b_i}$. The forms of regions $L_i$ for the lower CRR
are computed similarly, but with the signs of $c_i$ and $b_i$ flipped for each $i=1, \ldots, n$.

Both upper and lower confidence regions are built similarly to the kernel RRCM region
eq.~\ref{eq:rrcm_conf_ci} in sec.~\ref{sub:ridge_regression_confidence_machine}. The
final Kernel CRR confidence set is given by
\begin{equation} \label{eq:crr_conf_ci}
  \Gamma_{X_{-n}, y_{-n}}^\alpha(x_n)
    = \Gamma_{X_{-n}, y_{-n}}^{\alpha,\text{u}}(x_n)
    \cap \Gamma_{X_{-n}, y_{-n}}^{\alpha,\text{l}}(x_n)
    \,.
\end{equation}
This intersection can be computed efficiently in $\BigO(n \log{} n)$, since the regions
are built form sets anchored at a finite set $Q$ with at most $n+2$ values. Therefore,
the CRR confidence set for a fixed significance level $\alpha$ has $\BigO(n\log{} n)$
complexity.



\section{Numerical study} 
\label{sec:numerical_study}

Validity of conformal predictors in the online learning setting has been shown in
\cite{vovk2005}, chapter 2, however, no result of this kind is known in the batch
learning setting. Our experiments aim to evaluate the empirical performance of the
conformal prediction in this setting: with dedicated train and test datasets. In
this section we conduct a set of experiments to examine the validity of the regions,
produced by the conformal Kernel Ridge Regression and compare its efficiency to
the Bayesian confidence intervals. We use the isotropic Gaussian kernel with the
precision parameter $\theta>0$, $\Kcal(x,x') = \mathop{\text{exp}}\bigl
\{-\theta \|x - x'\|^2\bigr\}$, for both the Conformal Kernel ridge regression and
the Gaussian Process Regression. We experiment on a compact set $\Xcal\subset \Real^{d\times 1}$,
since the validity of conformal region is by design unaffected by the NCM $A$, which
can be an arbitrary computable function and is oblivious to the structure of the
domain. The dimensionality of the input data, however, may impact the width of
the constructed confidence region.

The off-line validity and efficiency of Bayesian and Conformal confidence regions
is studied in two settings: the fully Gaussian case, and the non-Gaussian cases.
In the first case the Bayesian assumptions hold and experiments are run on a sample
path of $GP(0, \Kcal + \delta_{x,x'} \gamma)$. In the second the assumptions of
Gaussian Process Regression are partially valid: a deliberately non-Gaussian $f$
in eq.~\ref{eq:signal_model} is contaminated by moderate Gaussian white noise with
variance $\gamma$.

The following hyper-parameters are controlled: the true noise-to-signal ratio $\gamma
\in \{10^{-6}, 10^{-1}\}$, the covariance kernel precision $\theta \in \{10, 10^2, 10^3\}$,
the train sample size $n$ (from $25$ up to $1600$), the NCM (eq.~\ref{eq:ins_ncm}
RRCM, or eq.~\ref{eq:crr_ncm} CRR), the residual eq.~\ref{eq:ins_resid}, or eq.~\ref{eq:loo_resid},
the regularization parameter $\lambda \in \{10^{-1}, 10^{-6}\}$, and either fixed
$\theta$ or $\theta$ that minimizes eq.~\ref{eq:bkrr_likelihood}.

For a given test function $f:\Xcal \mapsto \Real$ and a set of hyper-parameters
each experiment consists of the following steps:
\begin{enumerate}
  \item The test inputs, $X^*$, are given by a regular grid in $\Xcal$ with constant
  spacing;
  \item Train inputs, $X$, are sampled from a uniform distribution over $\Xcal$;
  \item For all $x\in X_{\text{pool}} = X \cup X^*$, target values $y_x = f(x)$
  are generated;
  \item For $l=1,\ldots, L$ independently
  \begin{enumerate}
    \item draw a random subsample of size $n$ from train dataset;
    \item fit a Gaussian Process Regression with zero mean and Gaussian kernel
    $\Kcal$ with the specified precision $\theta$ and $\lambda$;
    \item for each $x^* \in X^*$ construct the Bayesian (eq.~\ref{eq:gp_conf_int})
    and conformal (eq.~\ref{eq:conf_pred_set}) confidence regions using the NCM $A$
    and residuals $\hat{r}$ with MLE estimated $\sigma^2$;
    \item estimate the coverage rate and the width of the convex hull of the region
    over the test sample $X^*$: $p_l(R) = |X^*|^{-1}\sum_{x\in X^*} 1_{y_x\in R_x}$,
    and $w_l(R) = \inf\{b-a\,:\,R \subseteq [a, b]\}$, where $R$ is a confidence
    region;
  \end{enumerate}
\end{enumerate}
With the experimental procedure properly outlined, we proceed to summarizing the
results.

\subsection{Results: $1$-d} 
\label{sub:results_1_d}

\begin{figure}
  \centering
  \begin{subfigure}[b]{0.5\linewidth}
    \includegraphics[width=0.9\linewidth]{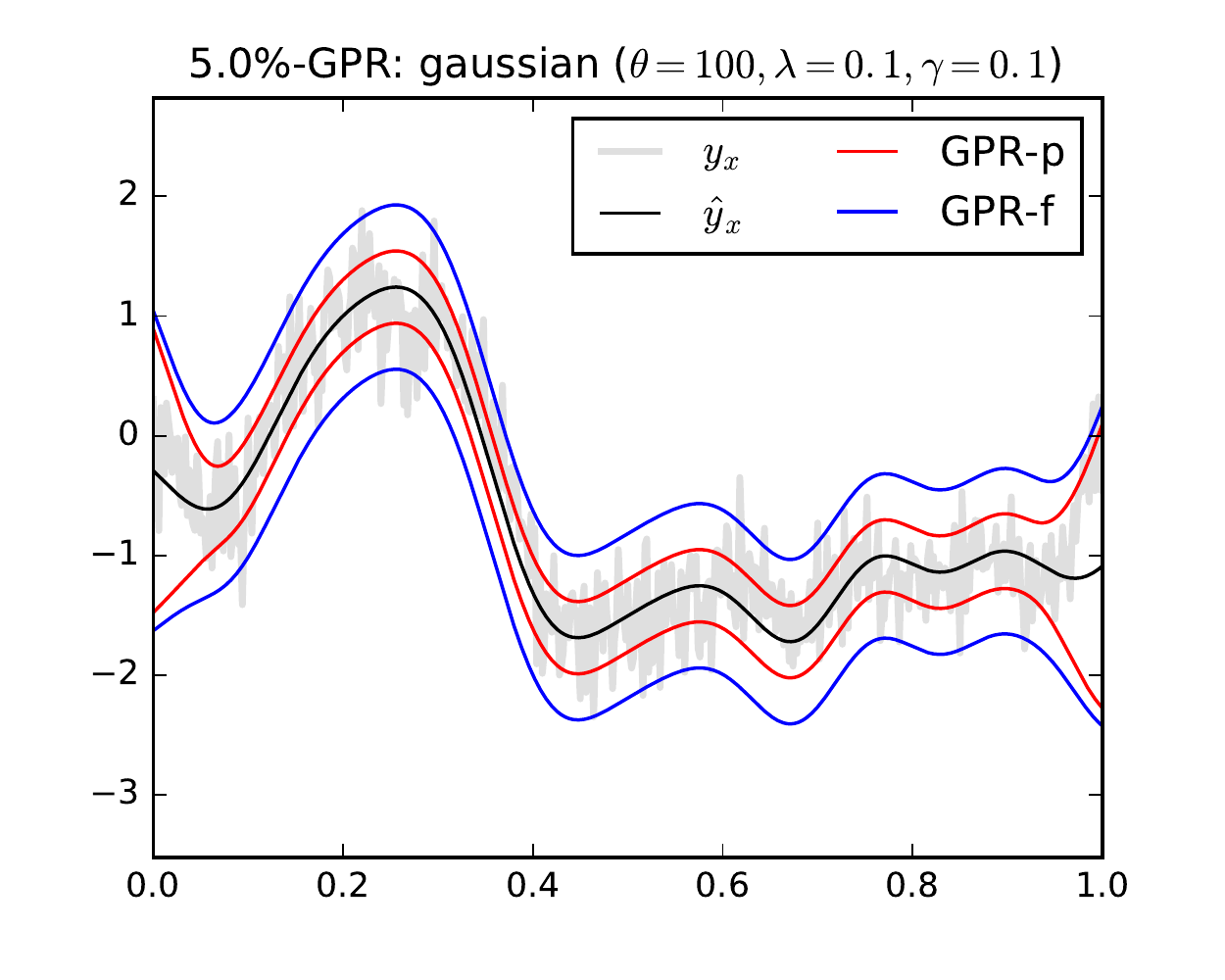}
  \end{subfigure}
  \begin{subfigure}[b]{0.5\linewidth}
    \includegraphics[width=0.9\linewidth]{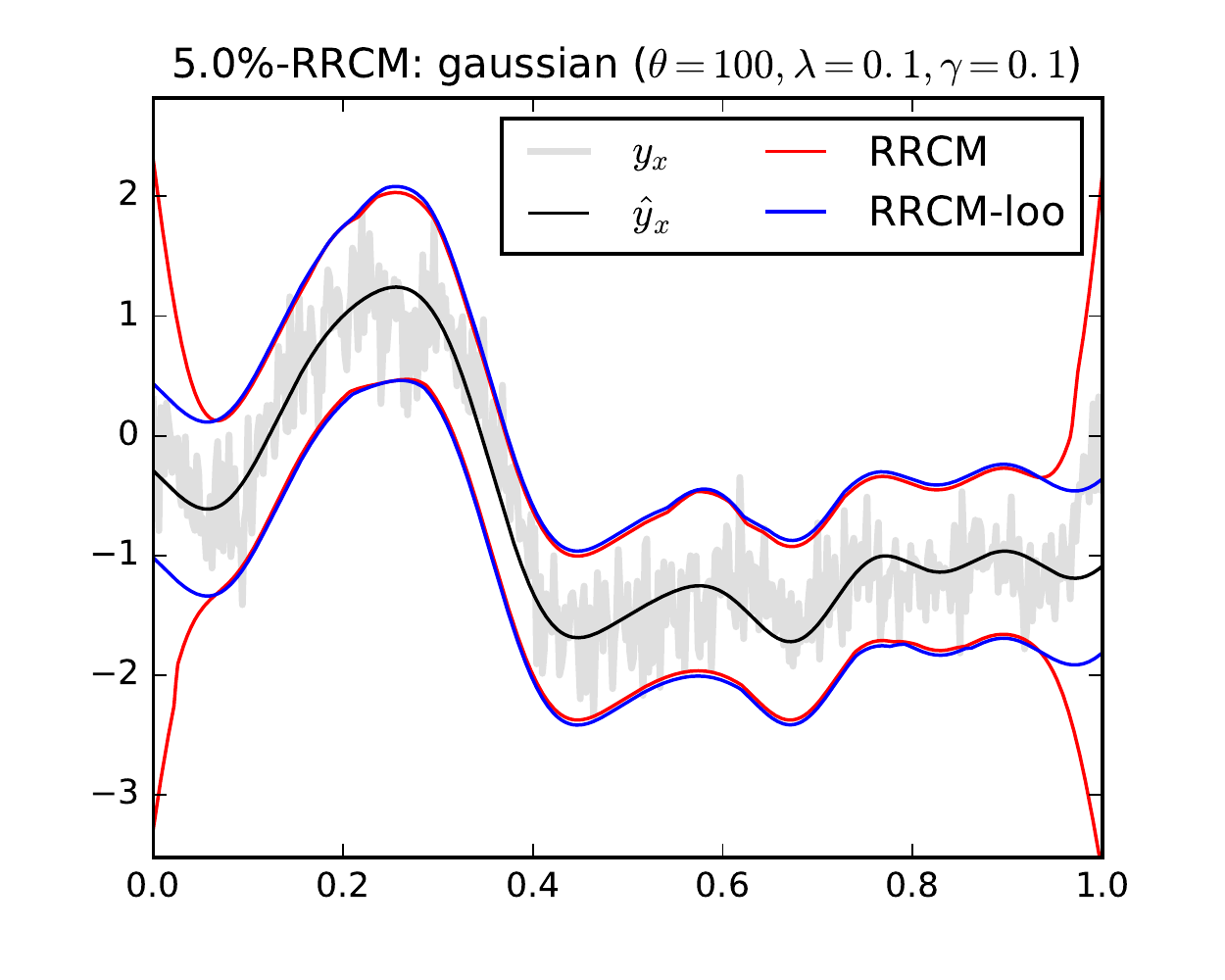}
  \end{subfigure}\\
  \begin{subfigure}[b]{0.5\linewidth}
    \includegraphics[width=0.9\linewidth]{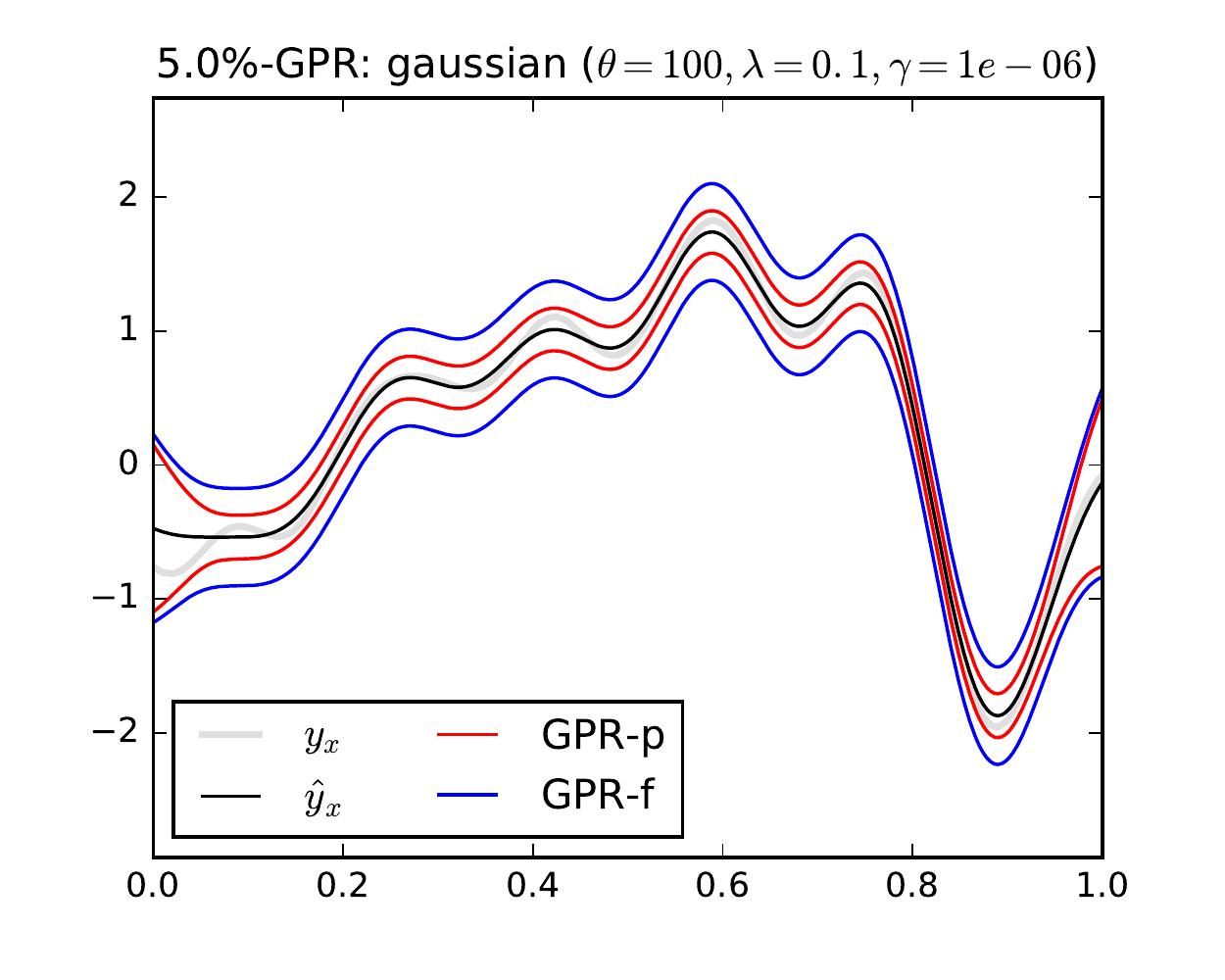}
  \end{subfigure}
  \begin{subfigure}[b]{0.5\linewidth}
    \includegraphics[width=0.9\linewidth]{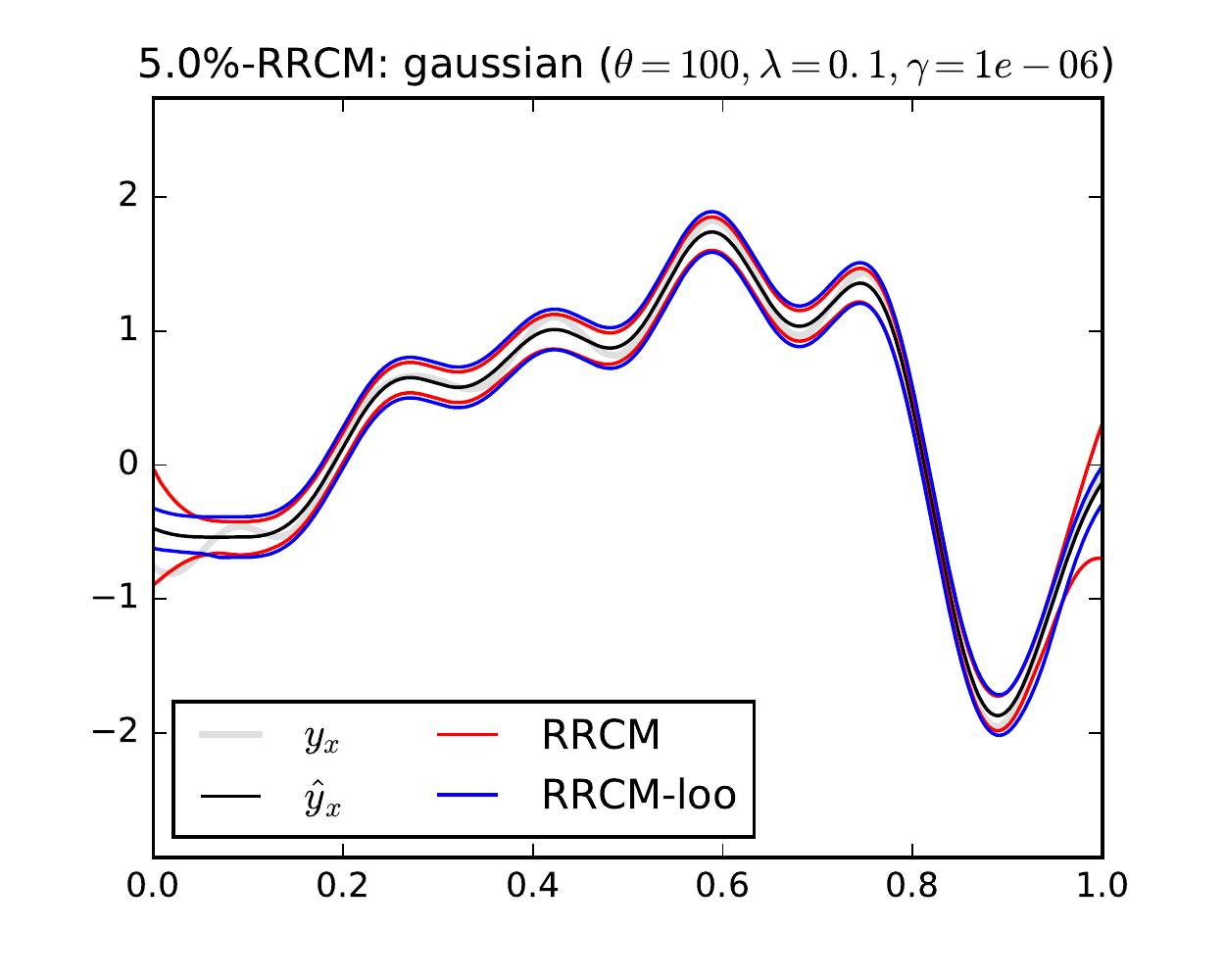}
  \end{subfigure}%
  \caption{A sample path of a $1$-d Gaussian Process with $\gamma=10^{-1}$ (\textit{top}),
  and $\gamma=10^{-6}$ (\textit{bottom}) constructed confidence intervals: the forecast
  ``GPR-f'' and prediction ``GPR-p'' (\textit{left}), and the ``RRCM'' confidence
  bands (\textit{right}).}
  \label{fig:gauss_1d_prof_gpr_conf}
\end{figure}

We begin with the examination of the fully-Gaussian setup with $\Xcal=[0, 1]$.
To illustrate the constructed confidence regions, we generated a sample path of
the $1$-d Gaussian process with isotropic Gaussian kernel on a regular grid of
$501$ knots, and use a subset of $51$ knots in $[0.05, 0.95]$ for constructing
the Bayesian (GPR) and conformal (RRCM) confidence regions. The confidence regions
are depicted in fig.~\ref{fig:gauss_1d_prof_gpr_conf}: conformal regions closely
track the Bayesian confidence bands (``GPR-f'', eq.~\ref{eq:gp_conf_int}), but the
latter are too wide in the low noise case. Near the endpoints the confidence regions
dramatically increase in width, reflecting increased uncertainty. In general, the
conformal regions necessarily cover the KRR prediction $\hat{y}^*_{|(X, y)}(x^*)$
but are not necessarily symmetric around it, where as GPR regions are (eq.~\ref{eq:gp_conf_int}).

\begin{figure}
  \centering
  \begin{subfigure}[b]{0.5\linewidth}
    \includegraphics[width=0.9\linewidth]{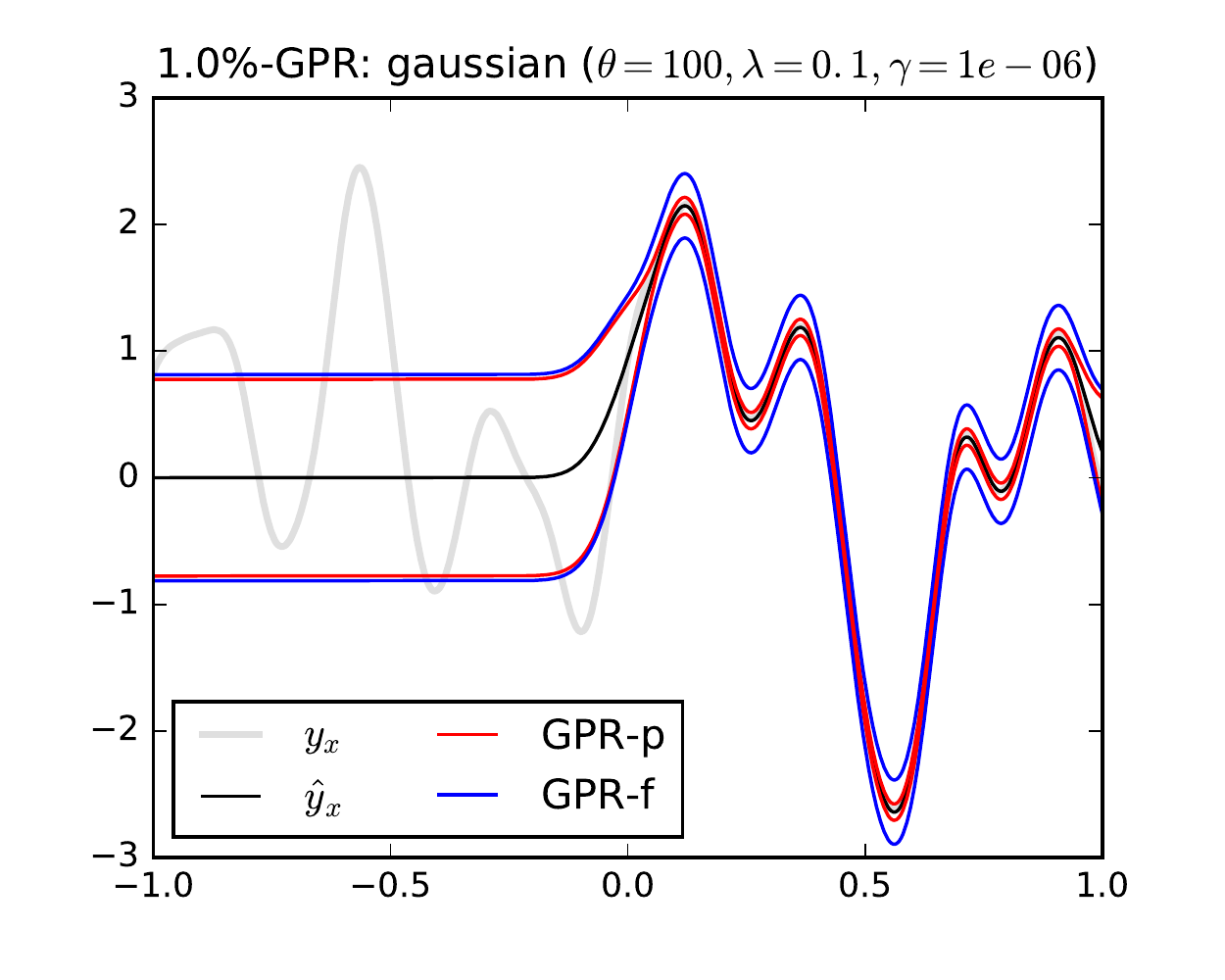}
  \end{subfigure}%
  \begin{subfigure}[b]{0.5\linewidth}
    \includegraphics[width=0.9\linewidth]{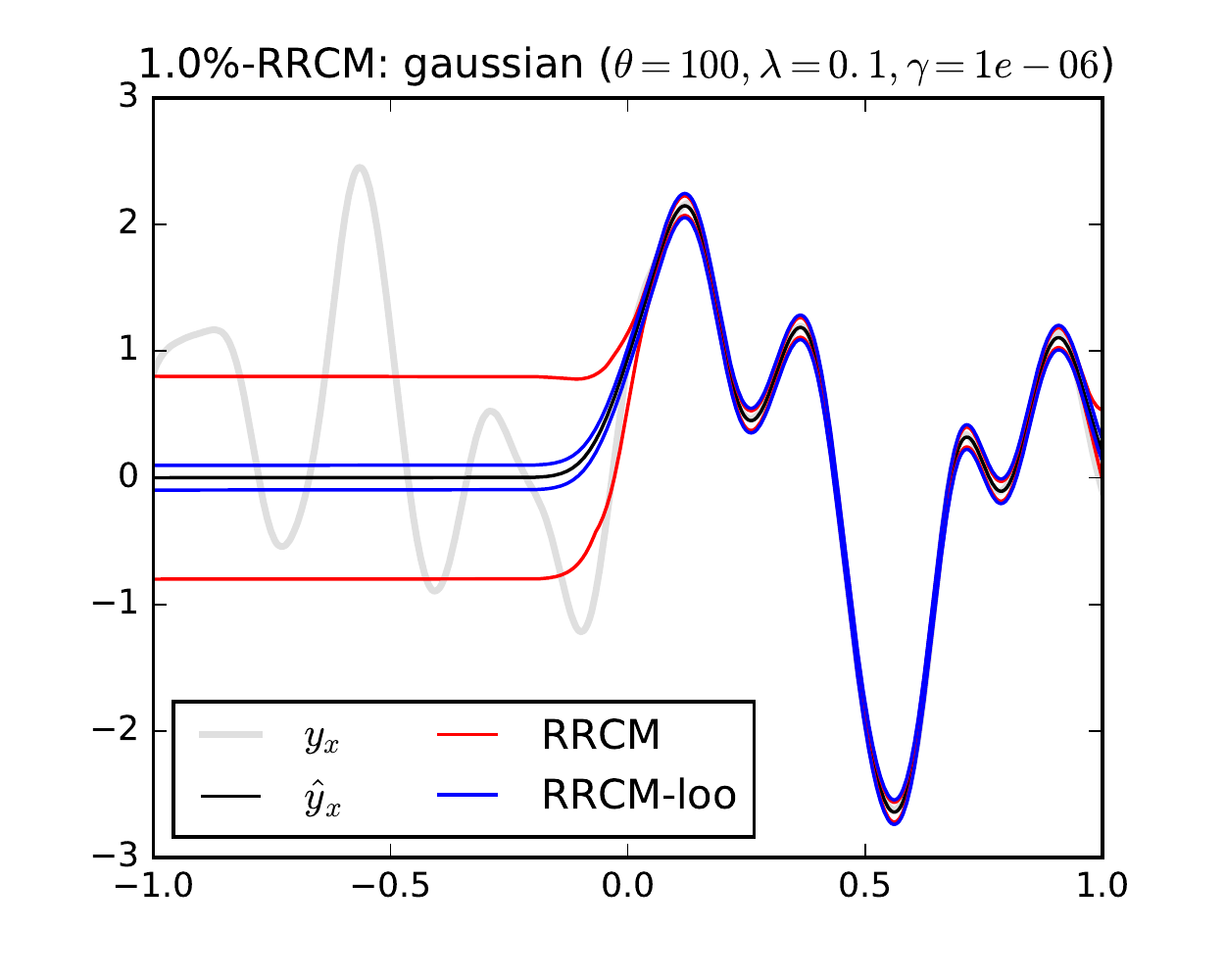}
  \end{subfigure}\\
  \begin{subfigure}[b]{0.5\linewidth}
    \includegraphics[width=0.9\linewidth]{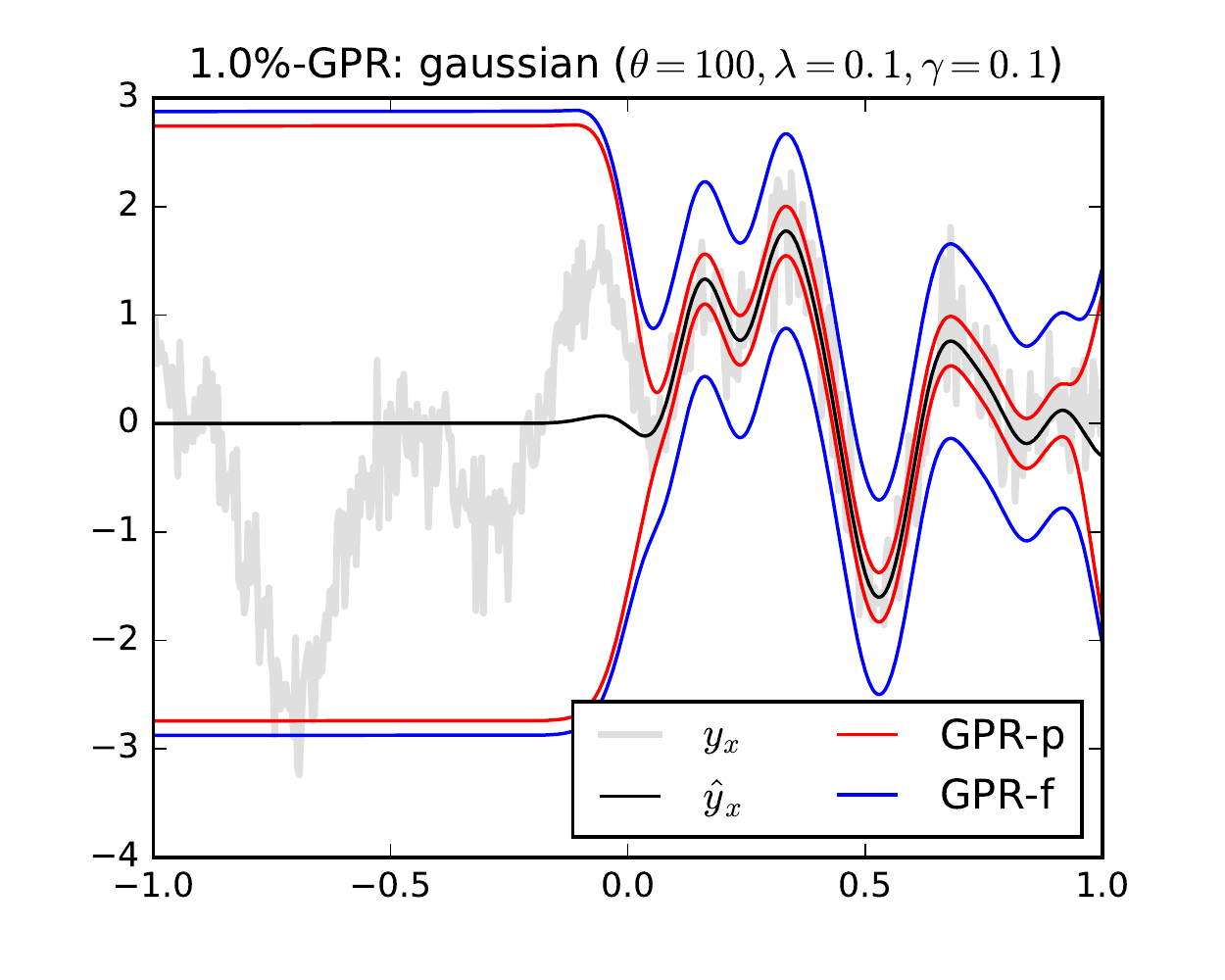}
  \end{subfigure}%
  \begin{subfigure}[b]{0.5\linewidth}
    \includegraphics[width=0.9\linewidth]{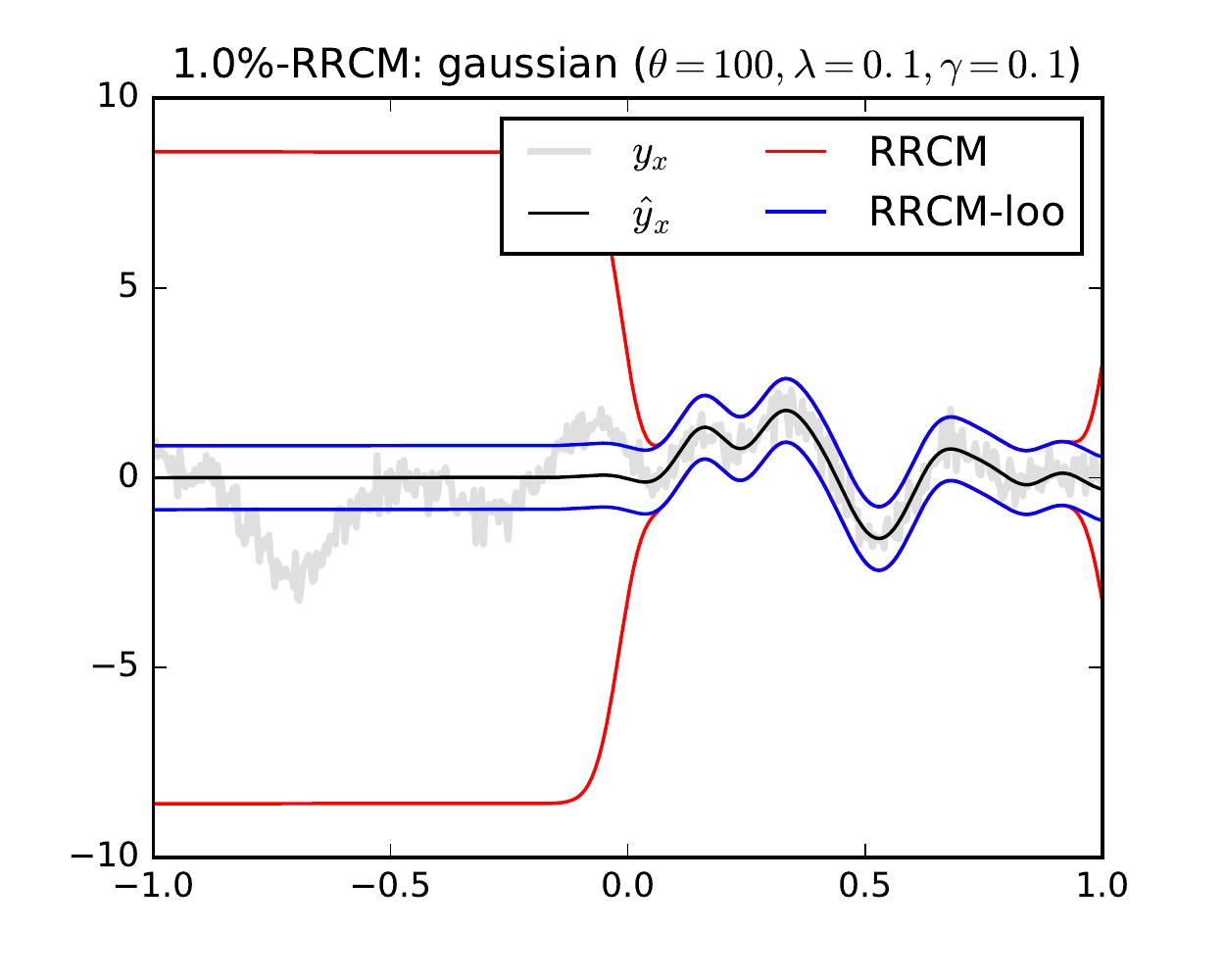}
  \end{subfigure}
  \caption{Limiting out-of-sample behaviour of GPR (\textit{left}) and RRCM (\textit{right})
    confidence regions for a sample path of a Gaussian process with negligible (\textit{top},
    $\gamma=10^{-6}$) and high (\textit{bottom}, $\gamma=10^{-1}$) noise-to-signal level.}
  \label{fig:limit_1d_ci_size}
\end{figure}

In can be argued, that confidence regions for any observation $x_n$ sufficiently
far away from the bulk of the training dataset have constant size, determined only
by the train sample (fig.~\ref{fig:limit_1d_ci_size}). Indeed, as $\|x_n\|^2\to \infty$
($n$-fixed) the vector $B_{-n}$ in (eq.~\ref{eq:krr_in_resid_B}) approaches the
$n$-th unit vector $e_n$, since for the Gaussian kernel the vector $\|K_{-n}(x_n)\|^2\to 0$.
Since the kernel is bounded, the value $m_n$ (eq.~\ref{eq:krr_leverage}) is a bounded
function of $x_n$, which, in turn, implies that eventually all RRCM (similarly, CRR)
regions $S_i$ assume the form of closed intervals $[-|q_i|, |q_i|]$, where
$q_i = m_n (e_i'Q_{-n}y_{-n}) + o(\|x_n\|^2)$, $i \neq n$. Therefore, the conformal
procedure essentially reverts to a constant-size confidence region, determined by
the $n^{-1}\lfloor n(1-\alpha)\rfloor$-th order statistic of $(|q_i|)_{i=1}^n$.
Analogous effects can be observed for the Gaussian Process confidence interval
(eq.~\ref{eq:gp_conf_int}).

By construction, conformal confidence region (eq.~\ref{eq:conf_pred_set}) allows
for some uncertainty. Indeed, the residuals (eq.~\ref{eq:krr_in_resid}) and hence
the vector of non-conformity scores $(\eta^z_i)_{i=1}^n$ are continuous functions
of targets $y=(y_i)_{i=1}^n$. Thus for small perturbations of $y$ the relative
ordering of $\eta^z_i$ is kept, and the interval remains unchanged. Therefore,
eq.~\ref{eq:conf_pred_set} tends to capture more points $y$ for the same fixed
significance level.

\begin{figure}
  \centering
  \begin{subfigure}[b]{0.25\linewidth}
    \includegraphics[width=0.95\linewidth]{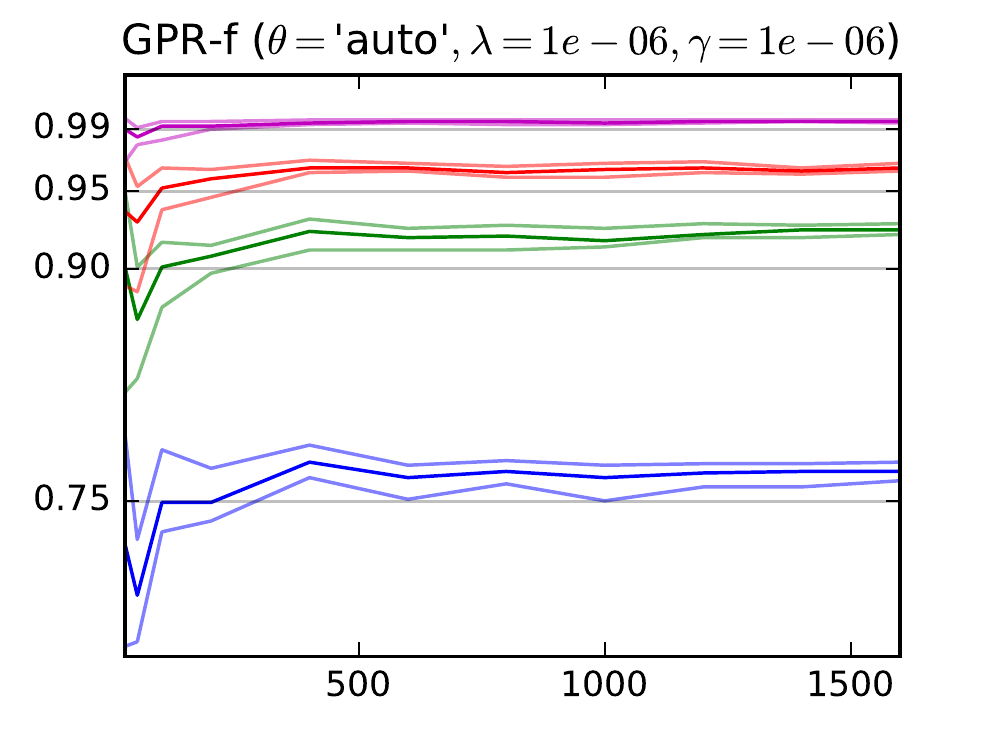}
  \end{subfigure}%
  \begin{subfigure}[b]{0.25\linewidth}
    \includegraphics[width=0.95\linewidth]{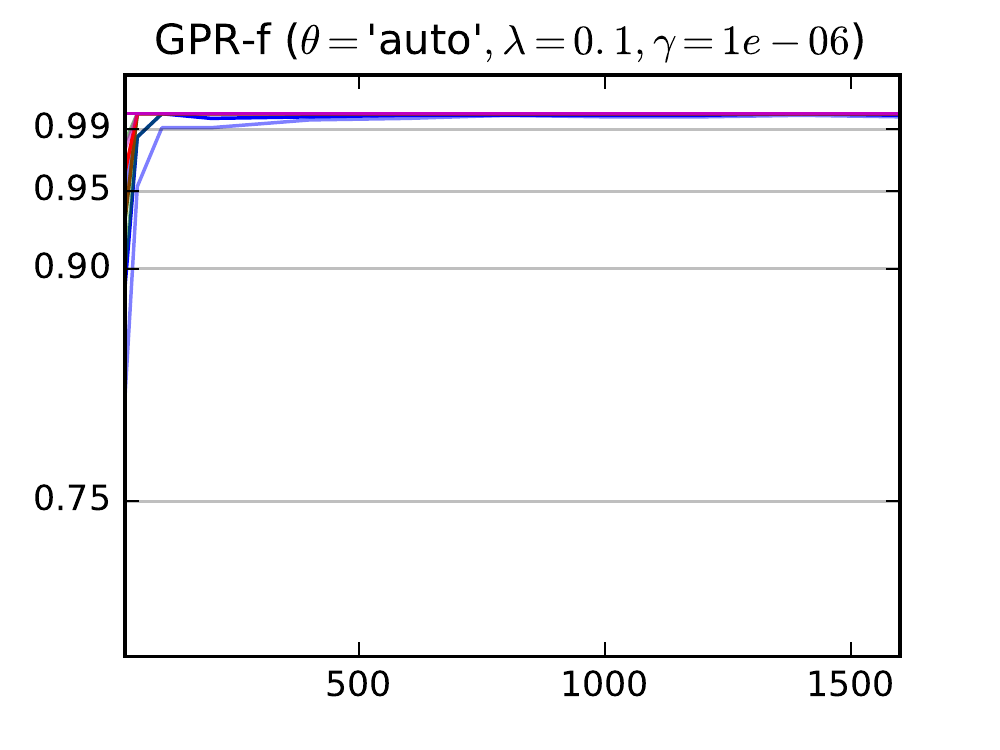}
  \end{subfigure}%
  \begin{subfigure}[b]{0.25\linewidth}
    \includegraphics[width=0.95\linewidth]{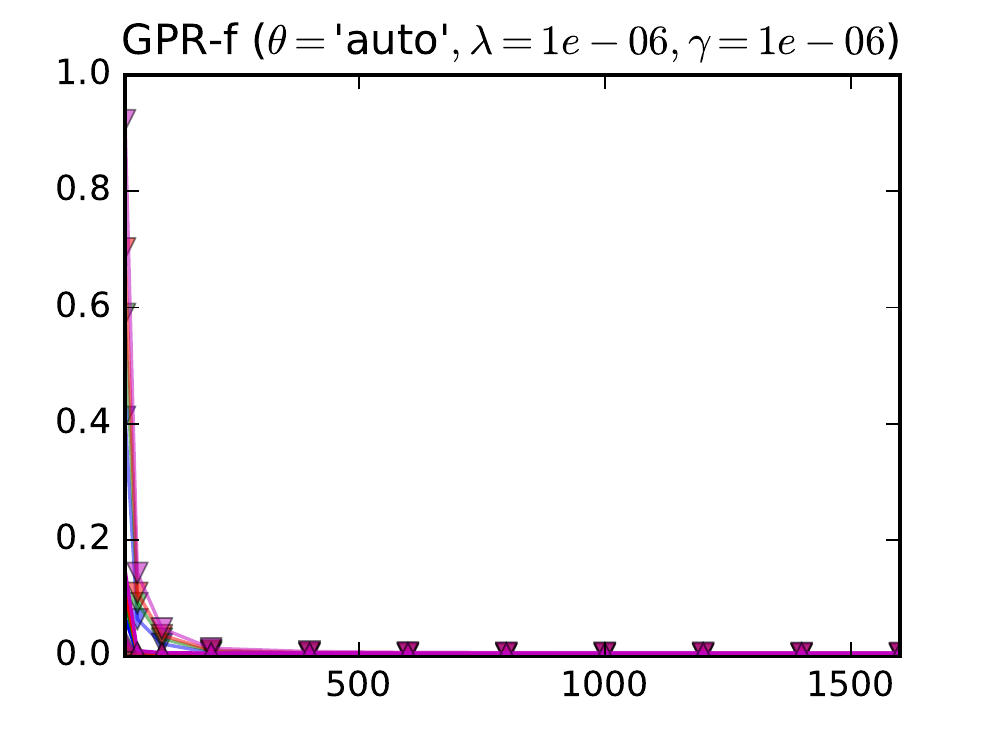}
  \end{subfigure}%
  \begin{subfigure}[b]{0.25\linewidth}
    \includegraphics[width=0.95\linewidth]{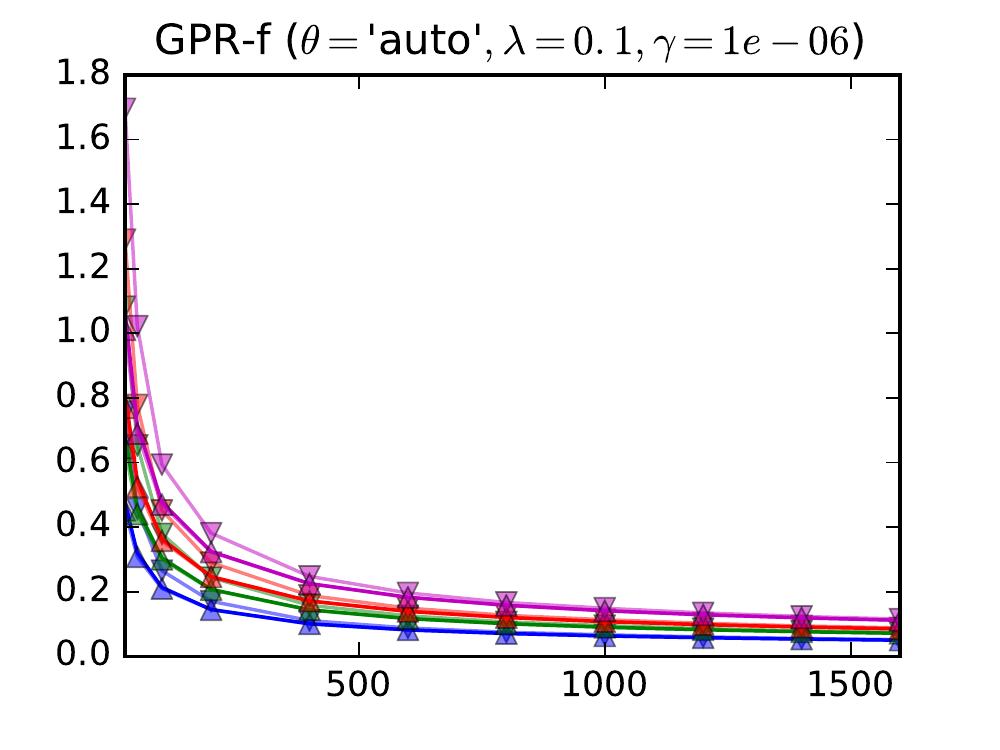}
  \end{subfigure}\\
  \begin{subfigure}[b]{0.25\linewidth}
    \includegraphics[width=0.95\linewidth]{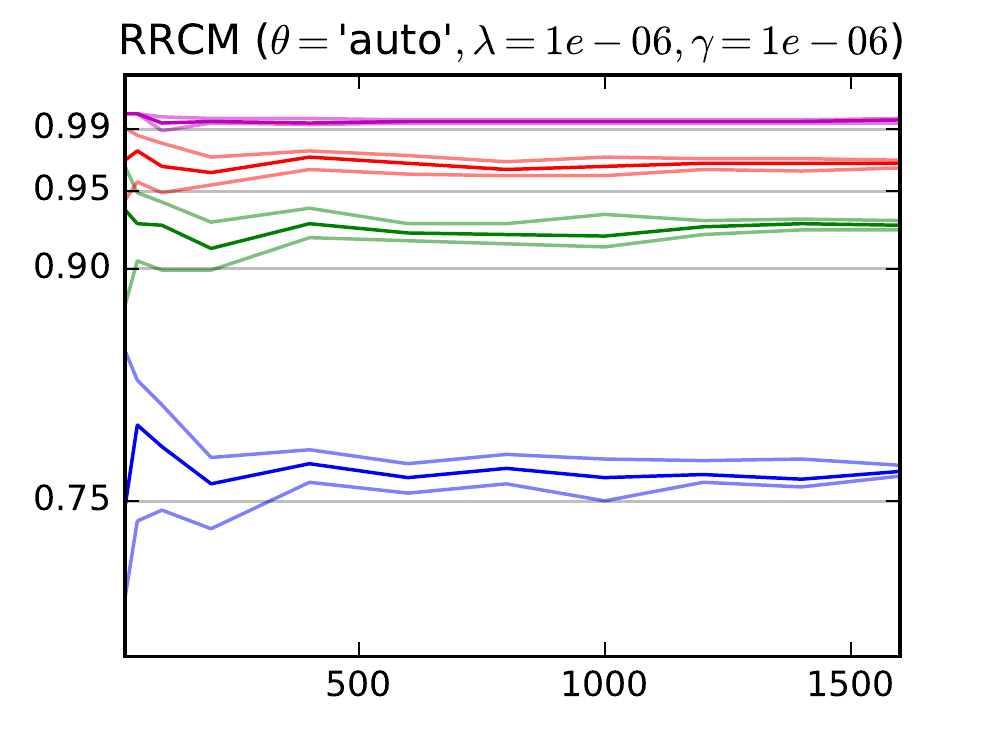}
  \end{subfigure}%
  \begin{subfigure}[b]{0.25\linewidth}
    \includegraphics[width=0.95\linewidth]{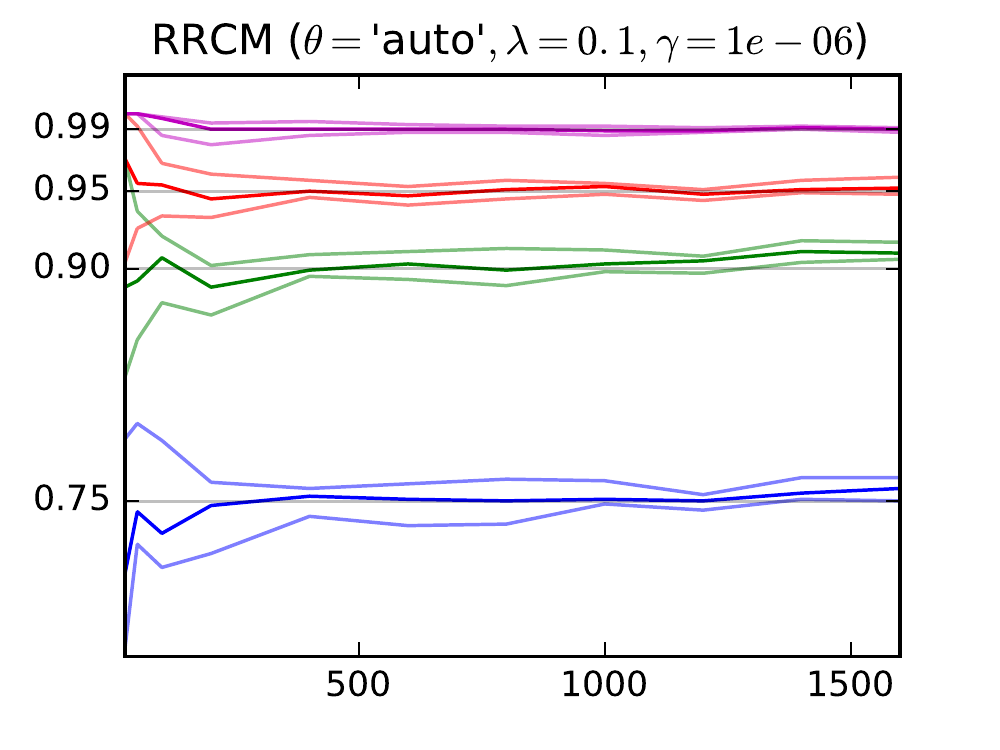}
  \end{subfigure}%
  \begin{subfigure}[b]{0.25\linewidth}
    \includegraphics[width=0.95\linewidth]{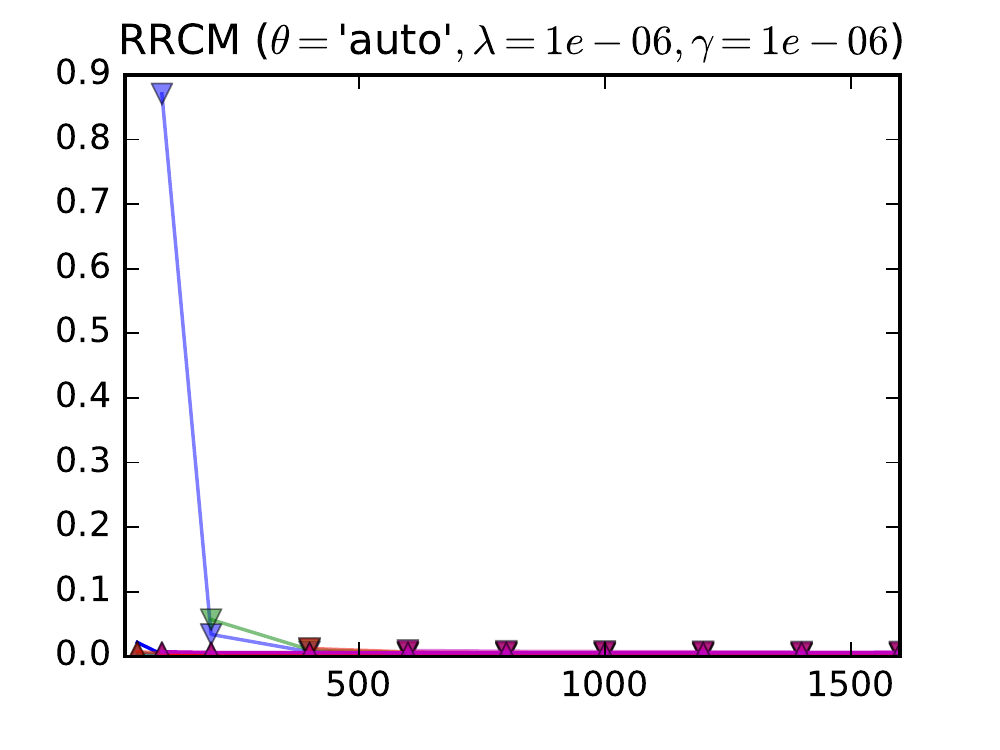}
  \end{subfigure}%
  \begin{subfigure}[b]{0.25\linewidth}
    \includegraphics[width=0.95\linewidth]{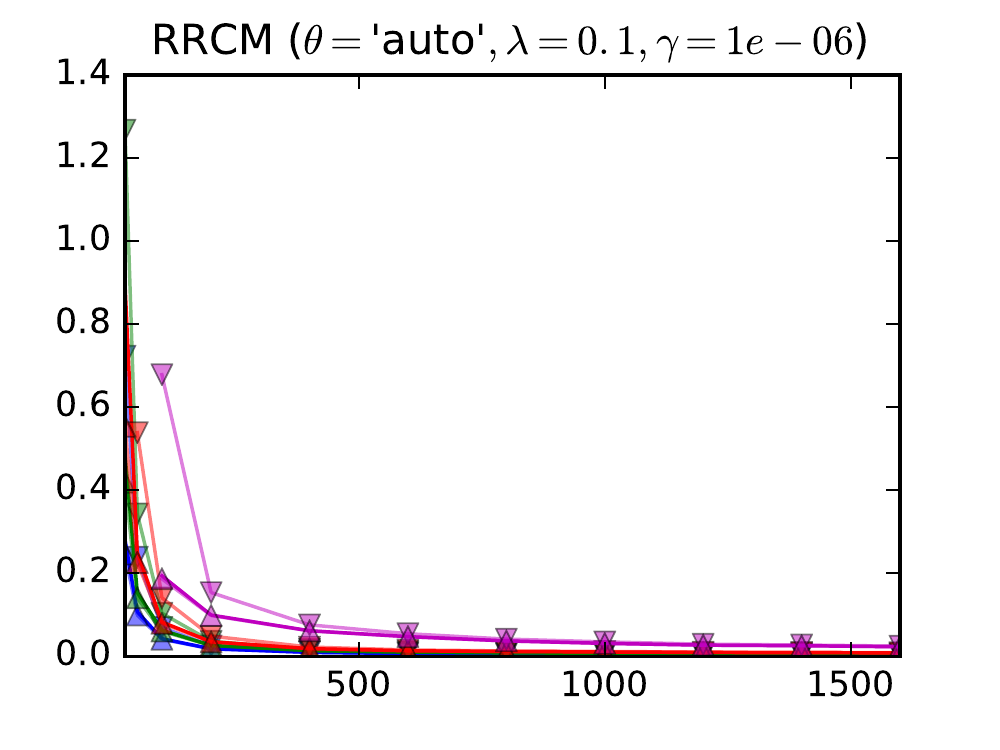}
  \end{subfigure}\\
  \begin{subfigure}[b]{0.25\linewidth}
    \includegraphics[width=0.95\linewidth]{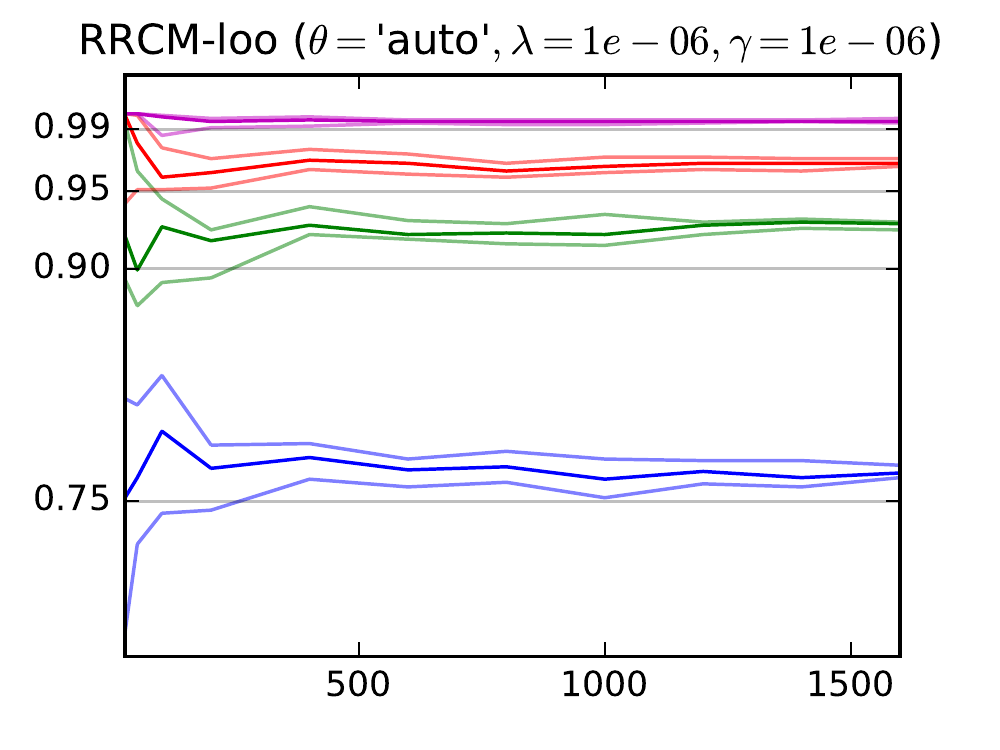}
  \end{subfigure}%
  \begin{subfigure}[b]{0.25\linewidth}
    \includegraphics[width=0.95\linewidth]{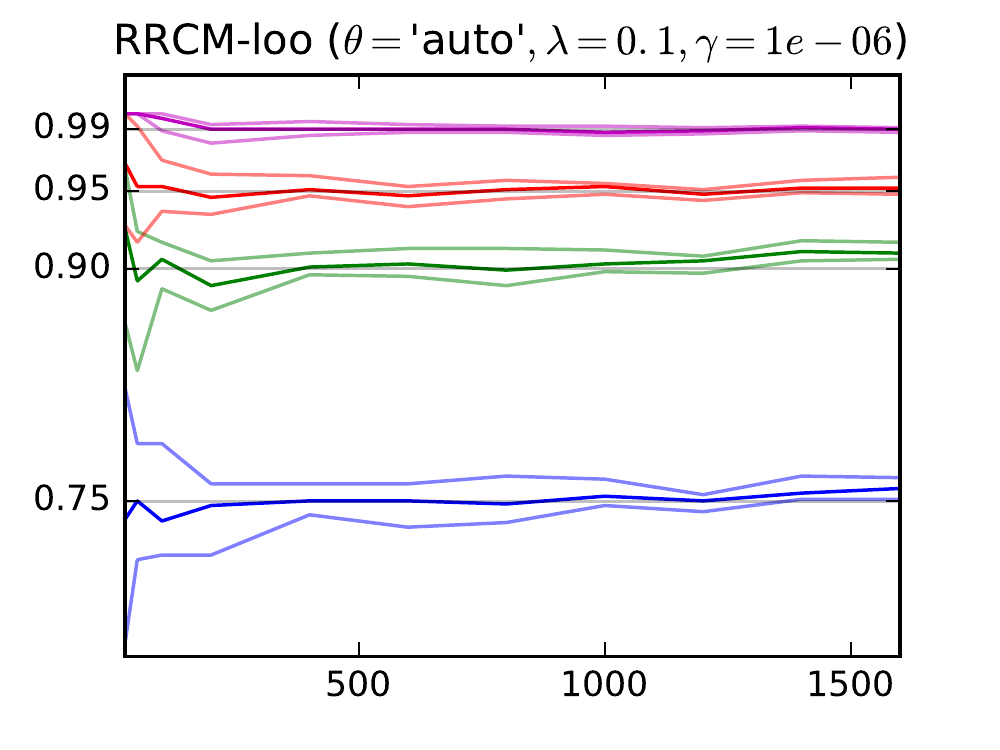}
  \end{subfigure}%
  \begin{subfigure}[b]{0.25\linewidth}
    \includegraphics[width=0.95\linewidth]{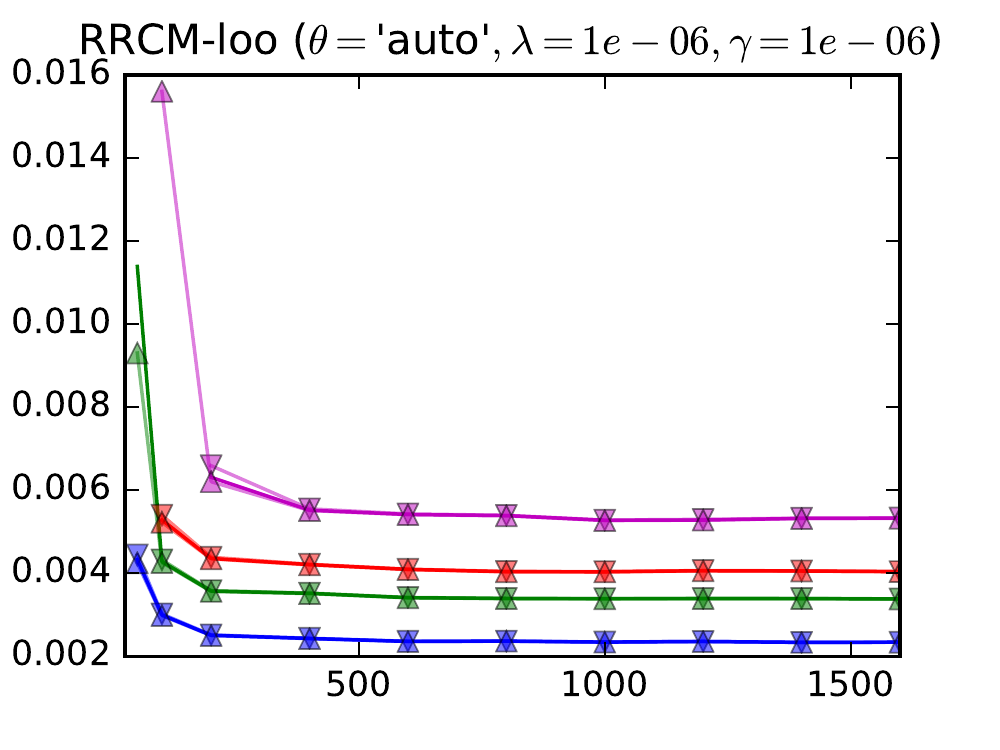}
  \end{subfigure}%
  \begin{subfigure}[b]{0.25\linewidth}
    \includegraphics[width=0.95\linewidth]{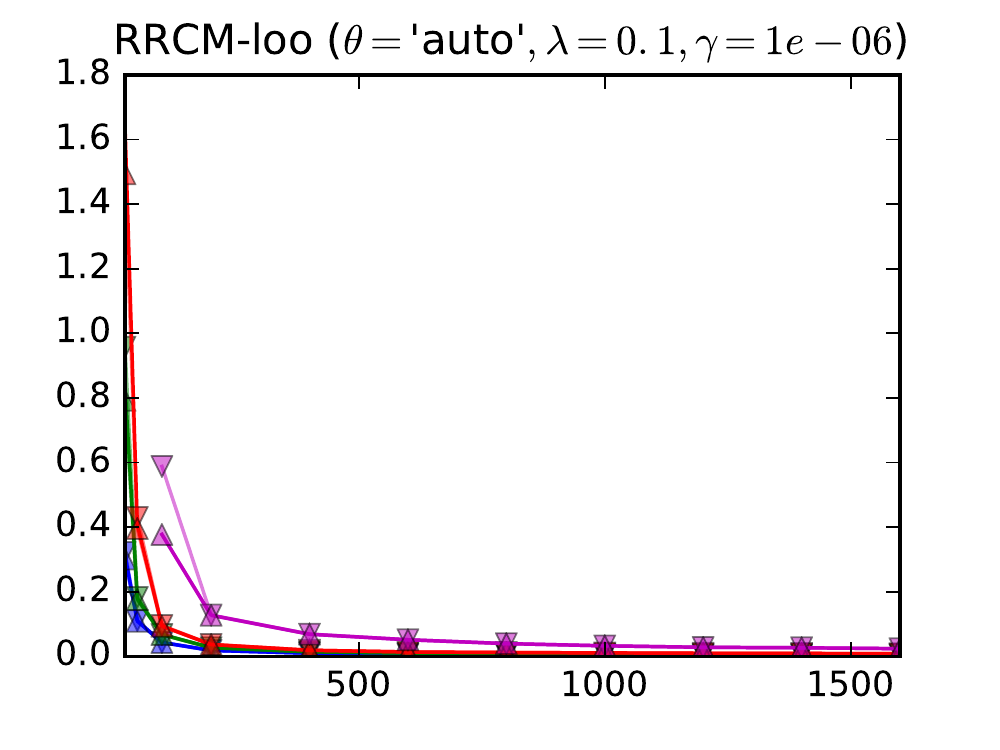}
  \end{subfigure}\\
  \begin{subfigure}[b]{0.25\linewidth}
    \includegraphics[width=0.95\linewidth]{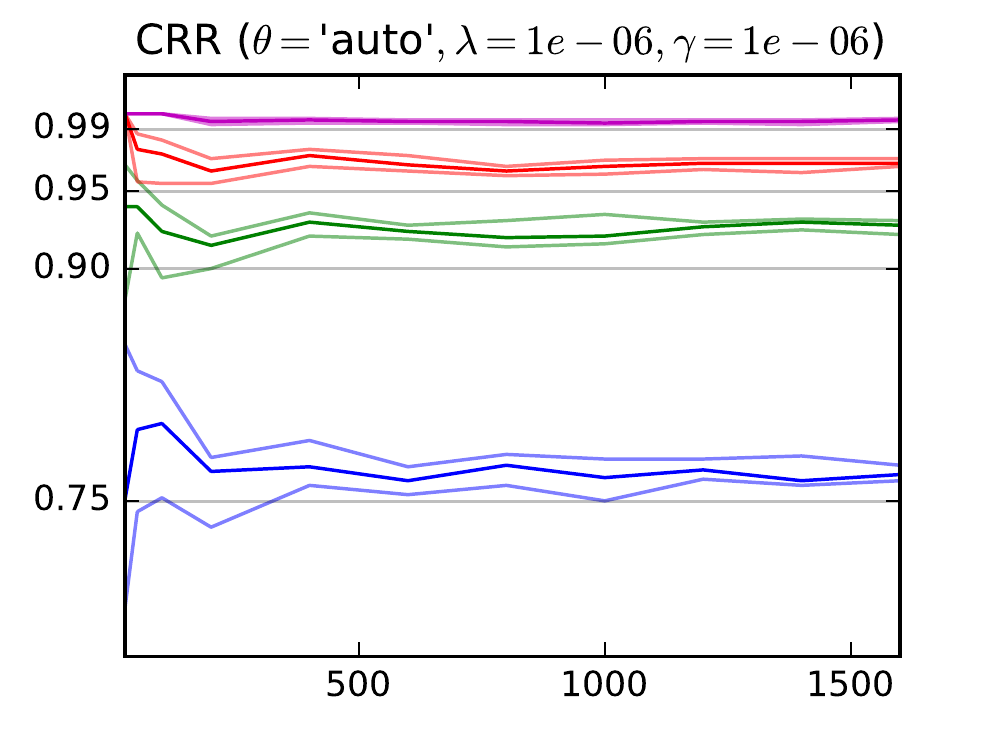}
  \end{subfigure}%
  \begin{subfigure}[b]{0.25\linewidth}
    \includegraphics[width=0.95\linewidth]{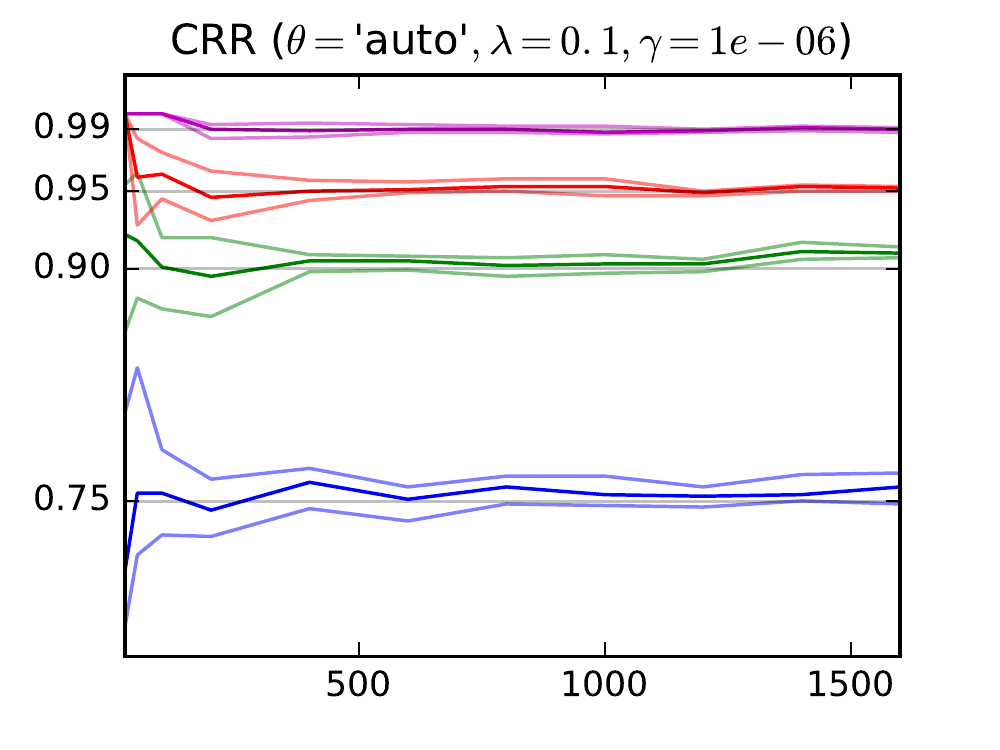}
  \end{subfigure}%
  \begin{subfigure}[b]{0.25\linewidth}
    \includegraphics[width=0.95\linewidth]{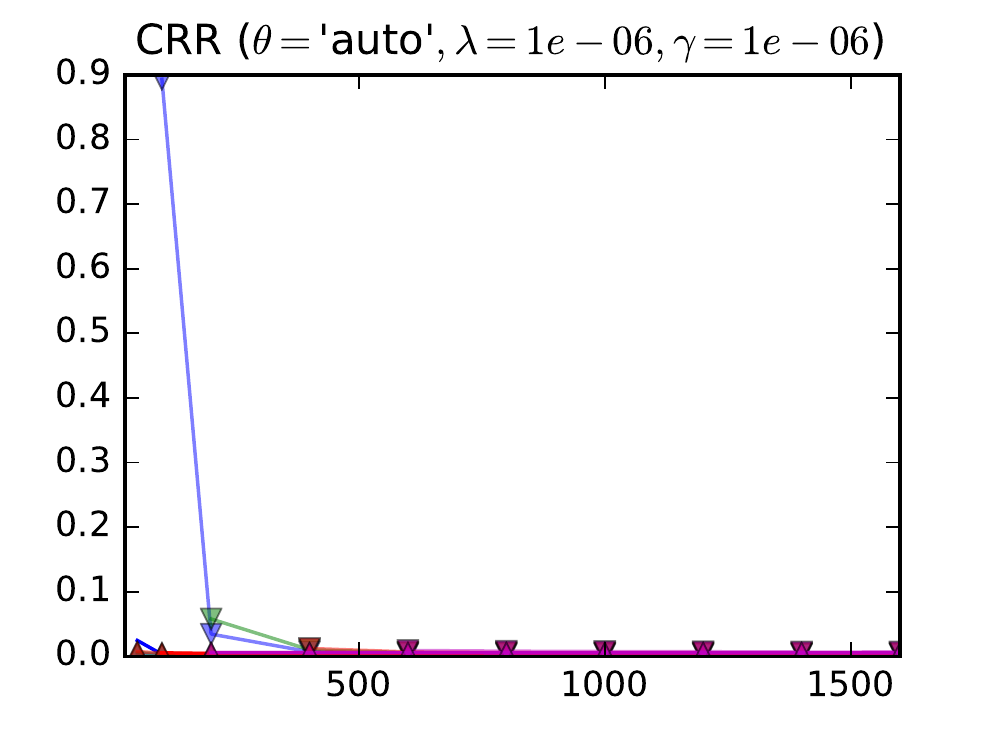}
  \end{subfigure}%
  \begin{subfigure}[b]{0.25\linewidth}
    \includegraphics[width=0.95\linewidth]{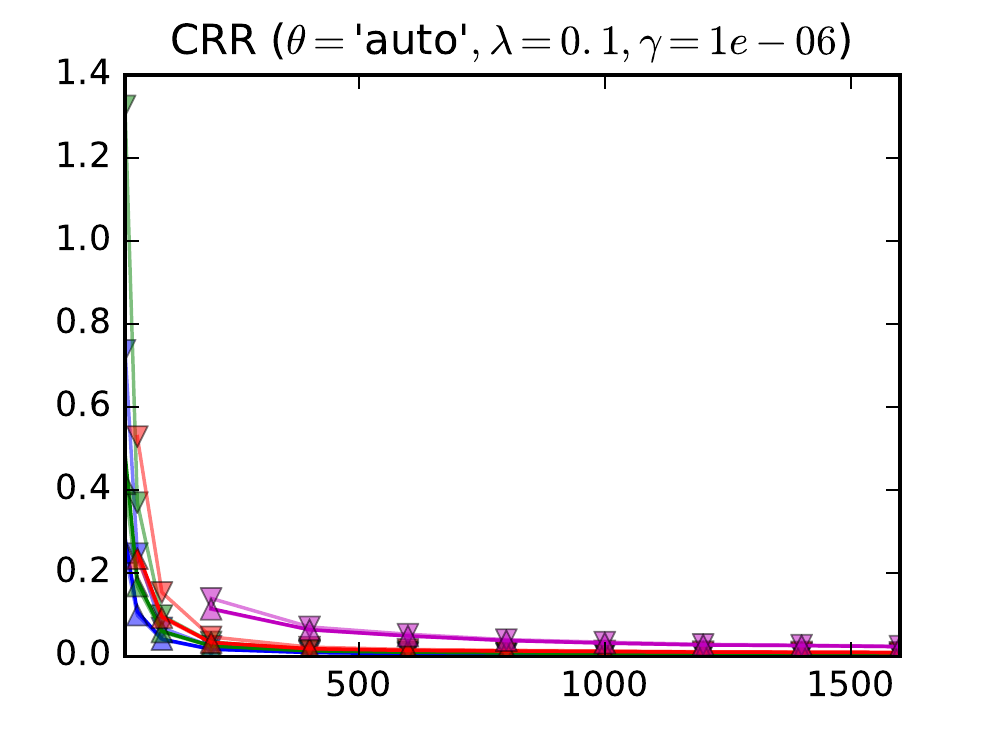}
  \end{subfigure}\\
  \begin{subfigure}[b]{0.25\linewidth}
    \includegraphics[width=0.95\linewidth]{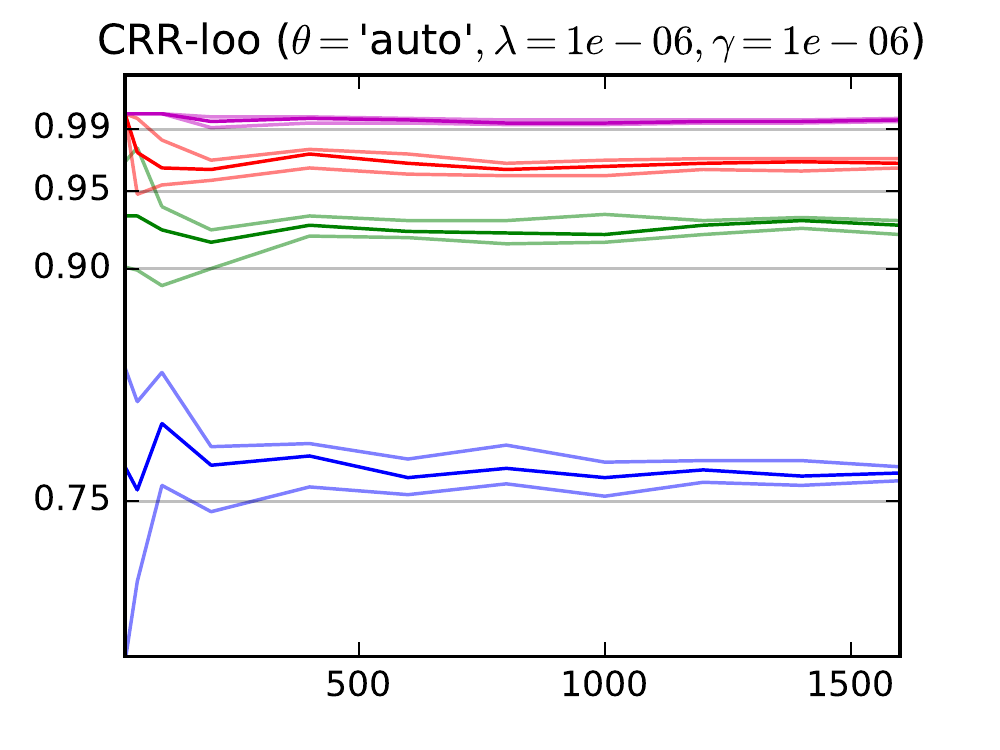}
    \caption{} \label{fig:gaussian_1d_low_noise_c3}
  \end{subfigure}%
  \begin{subfigure}[b]{0.25\linewidth}
    \includegraphics[width=0.95\linewidth]{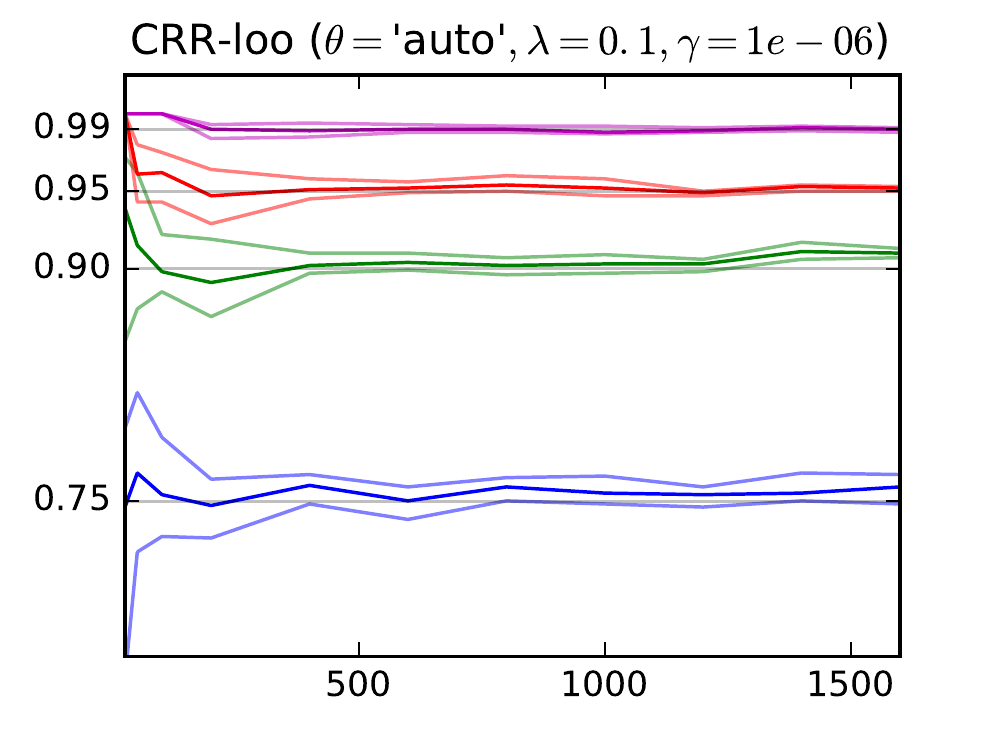}
    \caption{} \label{fig:gaussian_1d_low_noise_c4}
  \end{subfigure}%
  \begin{subfigure}[b]{0.25\linewidth}
    \includegraphics[width=0.95\linewidth]{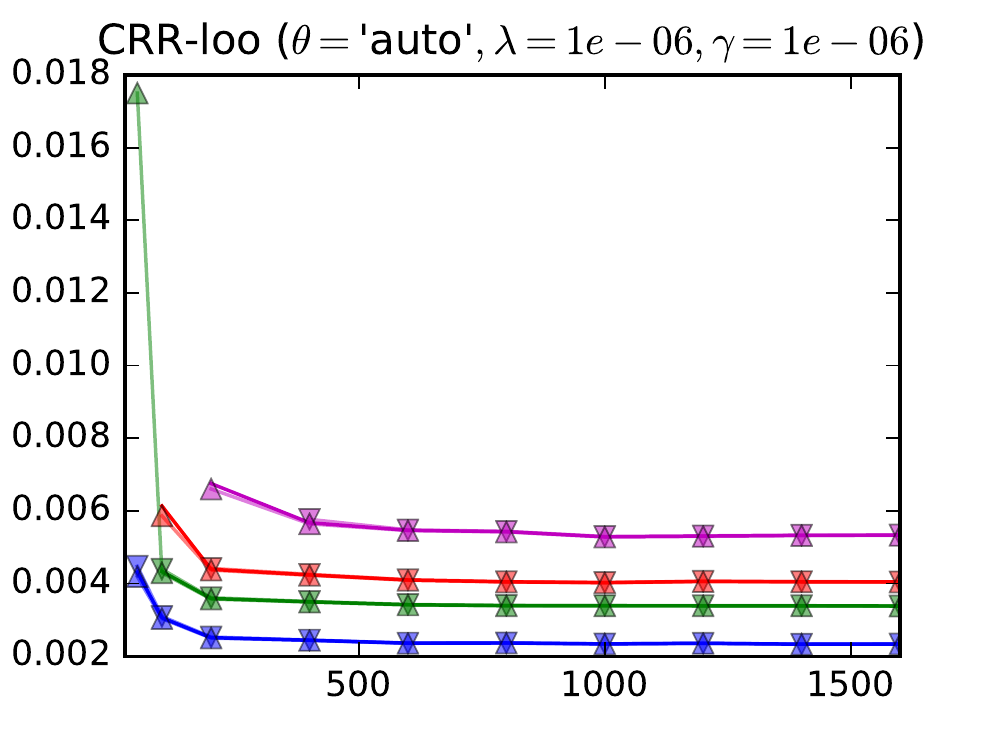}
    \caption{} \label{fig:gaussian_1d_low_noise_width_c3}
  \end{subfigure}%
  \begin{subfigure}[b]{0.25\linewidth}
    \includegraphics[width=0.95\linewidth]{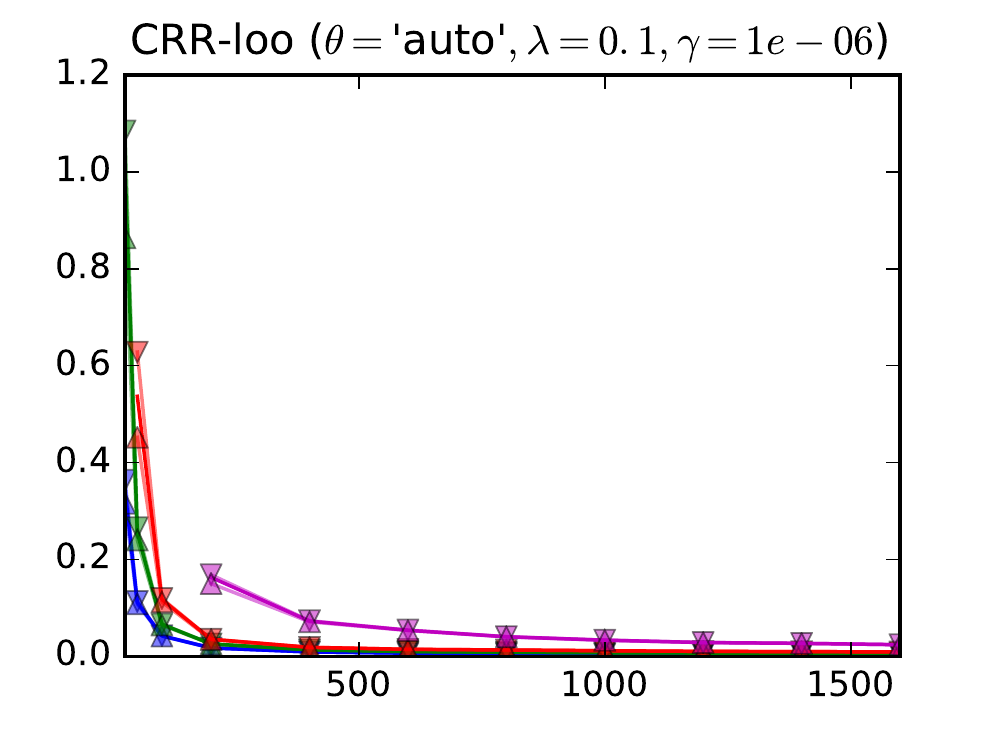}
    \caption{} \label{fig:gaussian_1d_low_noise_width_c4}
  \end{subfigure}%
  \caption{Coverage rate dynamics (\subref{fig:gaussian_1d_low_noise_c3}, \subref{fig:gaussian_1d_low_noise_c4})
  and asymptotic width (\subref{fig:gaussian_1d_low_noise_width_c3}, \subref{fig:gaussian_1d_low_noise_width_c4})
  of the confidence regions in the fully Gaussian low-noise case $\gamma=10^{-6}$
  for $\theta=\hat{\theta}_\text{ML}$ and $\lambda = 10^{-6}$ (\subref{fig:gaussian_1d_low_noise_c3},
  \subref{fig:gaussian_1d_low_noise_width_c3}), and $\lambda = 10^{-1}$ (\subref{fig:gaussian_1d_low_noise_c4},
  \subref{fig:gaussian_1d_low_noise_width_c4}). Rows from \textit{top} to \textit{bottom}: ``GPR-f'', ``RRCM'',
  ``RRCM-loo'', ``CRR'', ``CRR-loo''. In columns \subref{fig:gaussian_1d_low_noise_width_c3} and
  \subref{fig:gaussian_1d_low_noise_width_c4} upward triangles indicate the $5\%$ sample
  quantile across the whole test sample, downward triangles indicate the maximal width,
  the median width is drawn with a slightly thicker line.}
  \label{fig:gaussian_1d_low_noise}
\end{figure}

Experimental results in the perfectly noiseless case show that both confidence regions
are conservative (see fig.~\ref{fig:gaussian_1d_low_noise}). In this picture we consider
the ML estimate of $\theta$, but the results for other fixed choices are qualitatively
similar. By construction, the procedure (eq.~\ref{eq:conf_pred_set}) adapts to the noise
level in observations through the distribution of the non-conformity scores, rather
than the regularization parameter $\lambda$, which makes Bayesian confidence intervals
usually wider than conformal regions (fig.~\ref{fig:gaussian_1d_low_noise_c4}). Nevertheless
for higher $\lambda$ the coverage rate of conformal regions gets closer to the specified
confidence level.

\begin{figure}
  \centering
  \begin{subfigure}[b]{0.5\linewidth}
    \includegraphics[width=0.9\linewidth]{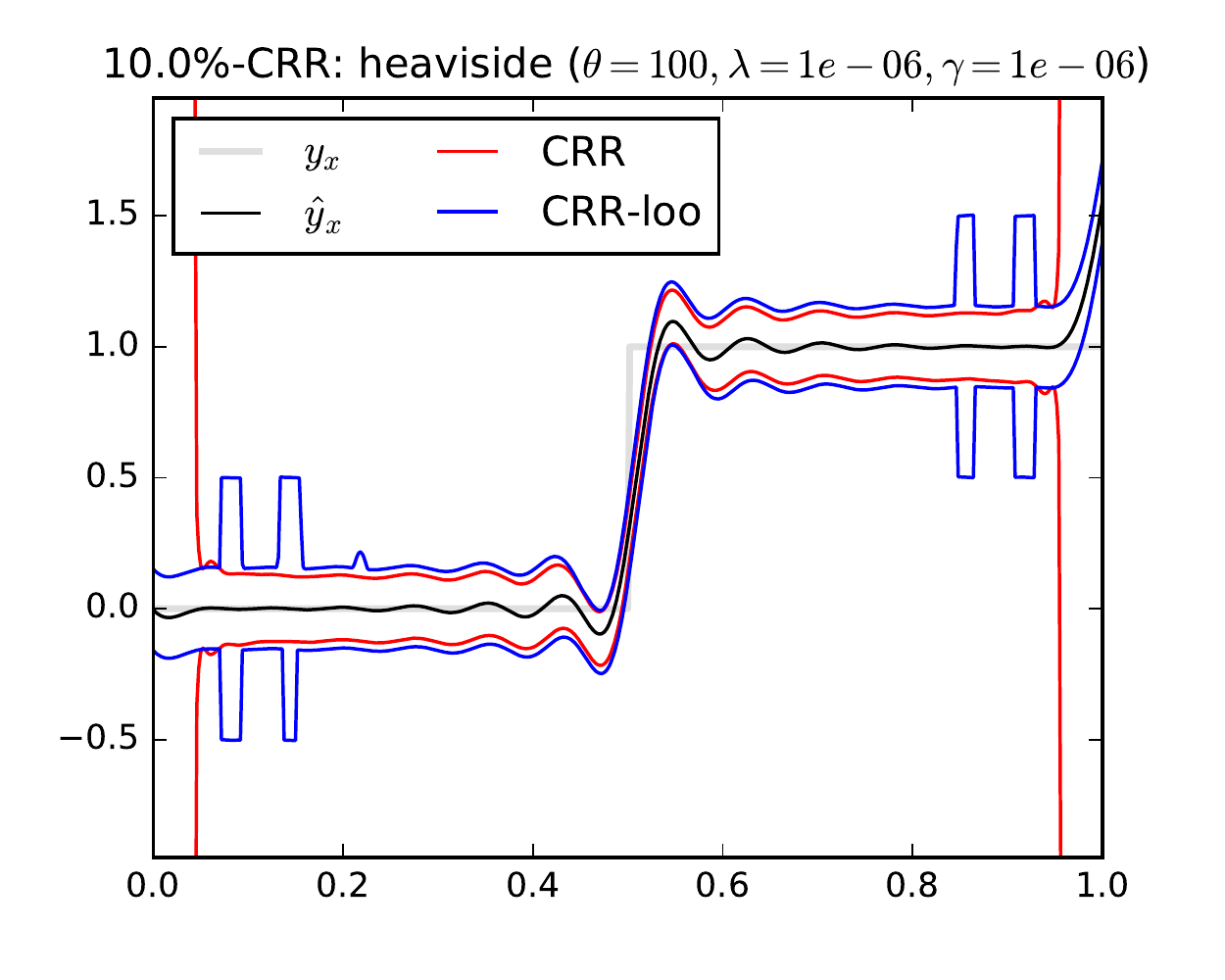}
  \end{subfigure}%
  \begin{subfigure}[b]{0.5\linewidth}
    \includegraphics[width=0.9\linewidth]{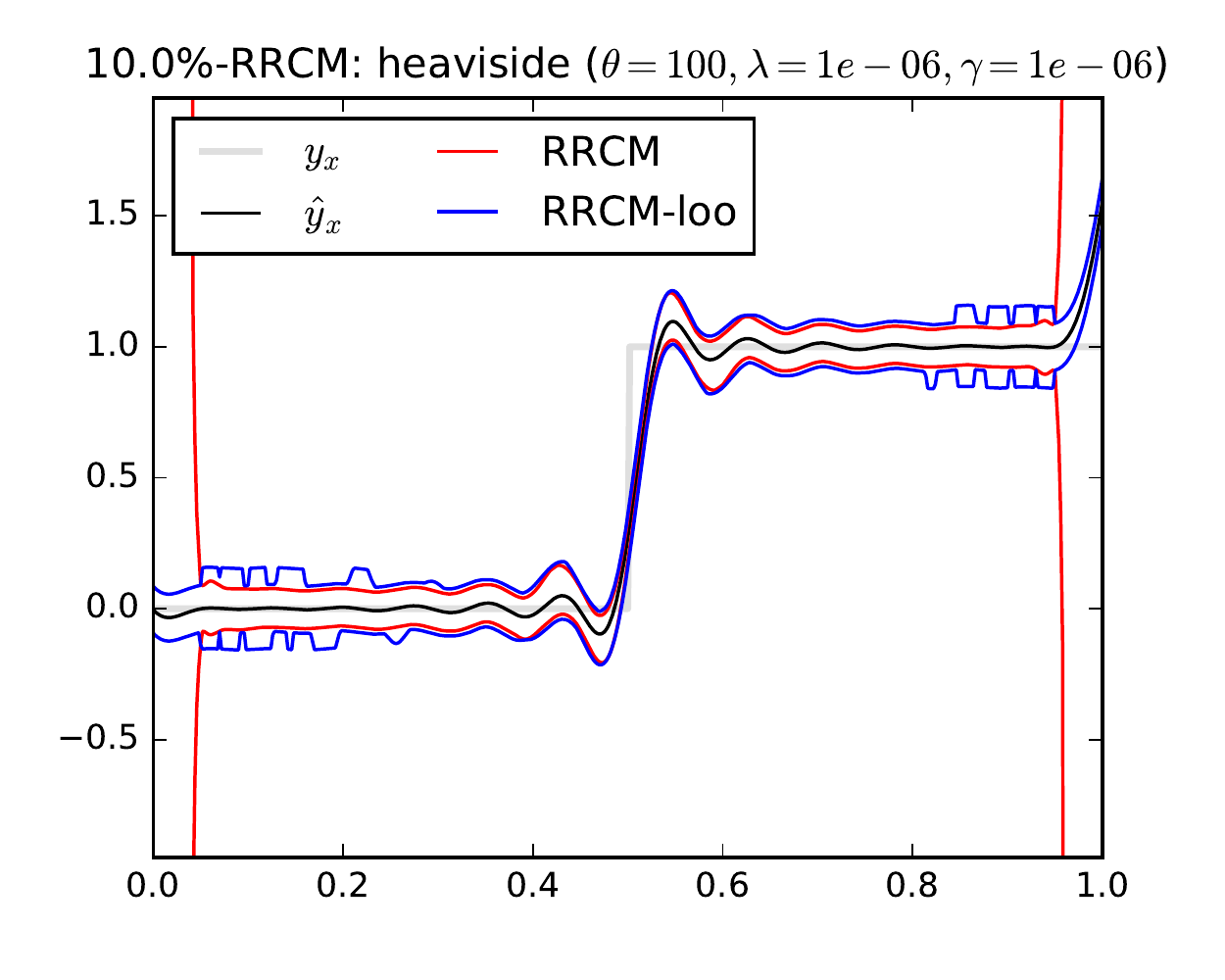}
  \end{subfigure}
  \caption{Typical conformal confidence bands for the ``Heaviside'' step function
  (train sample size $n=50$): \textit{left} -- CRR, and \textit{right} -- RRCM.}
  \label{fig:nongauss_1d_heaviside}
\end{figure}

\begin{figure}
  \centering
  \begin{subfigure}[b]{0.25\linewidth}
    \includegraphics[width=0.95\linewidth]{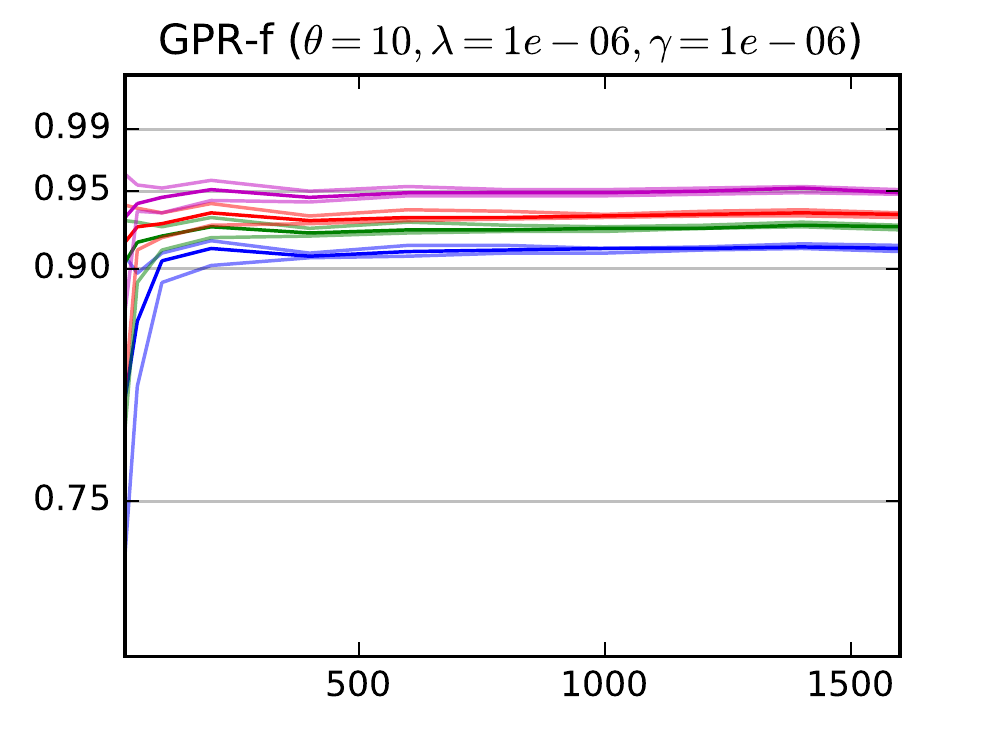}
  \end{subfigure}%
  \begin{subfigure}[b]{0.25\linewidth}
    \includegraphics[width=0.95\linewidth]{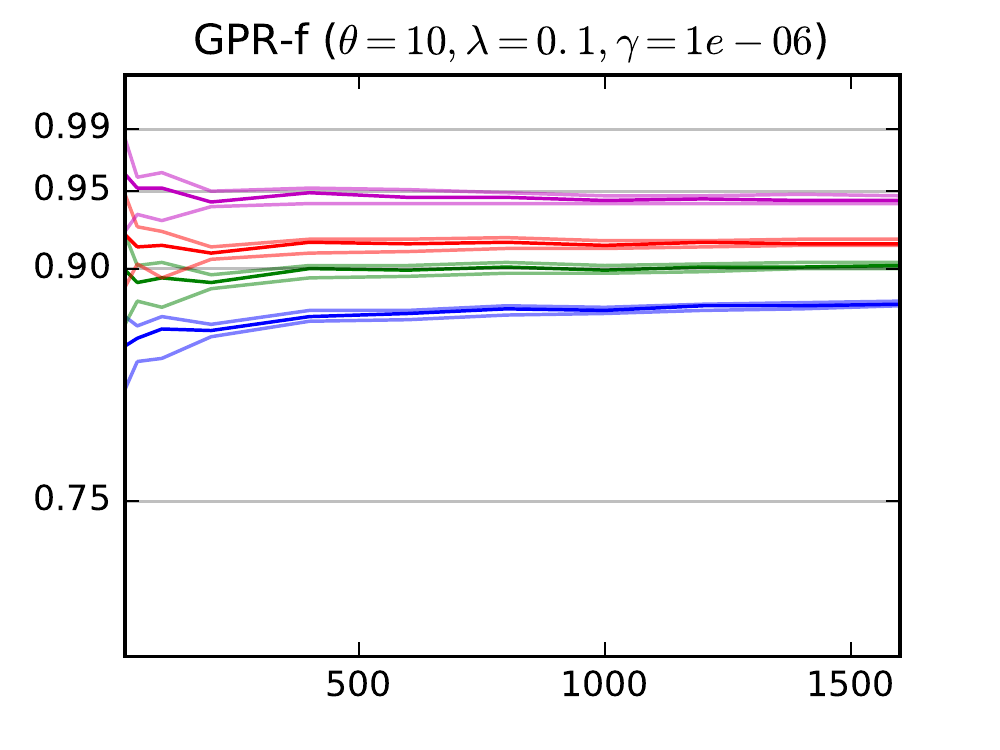}
  \end{subfigure}%
  \begin{subfigure}[b]{0.25\linewidth}
    \includegraphics[width=0.95\linewidth]{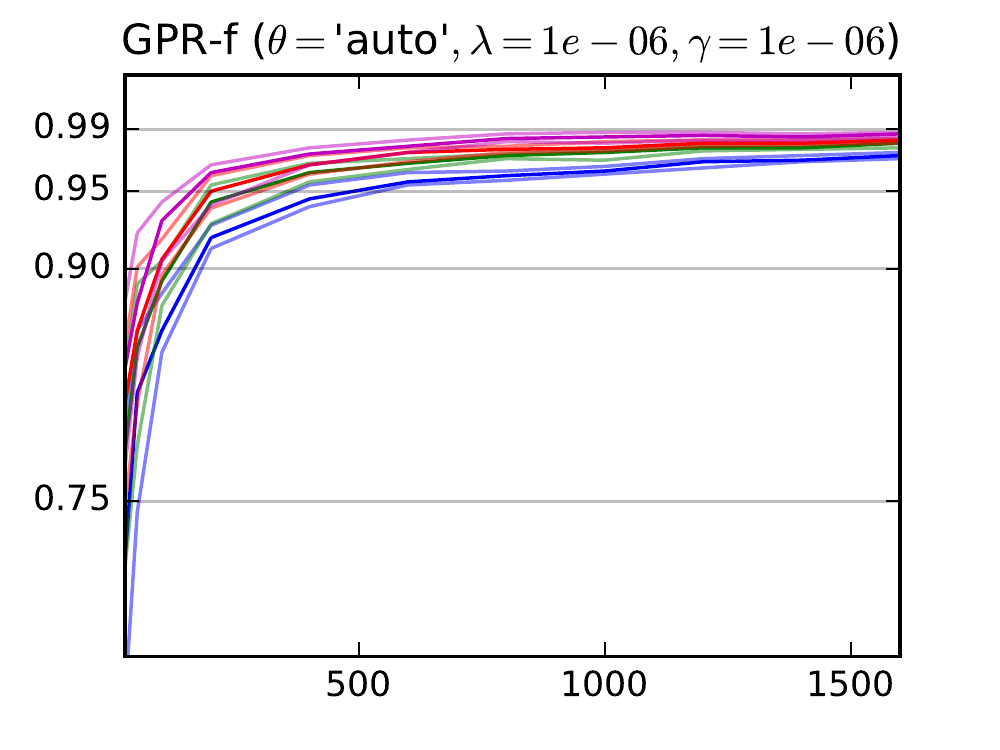}
  \end{subfigure}%
  \begin{subfigure}[b]{0.25\linewidth}
    \includegraphics[width=0.95\linewidth]{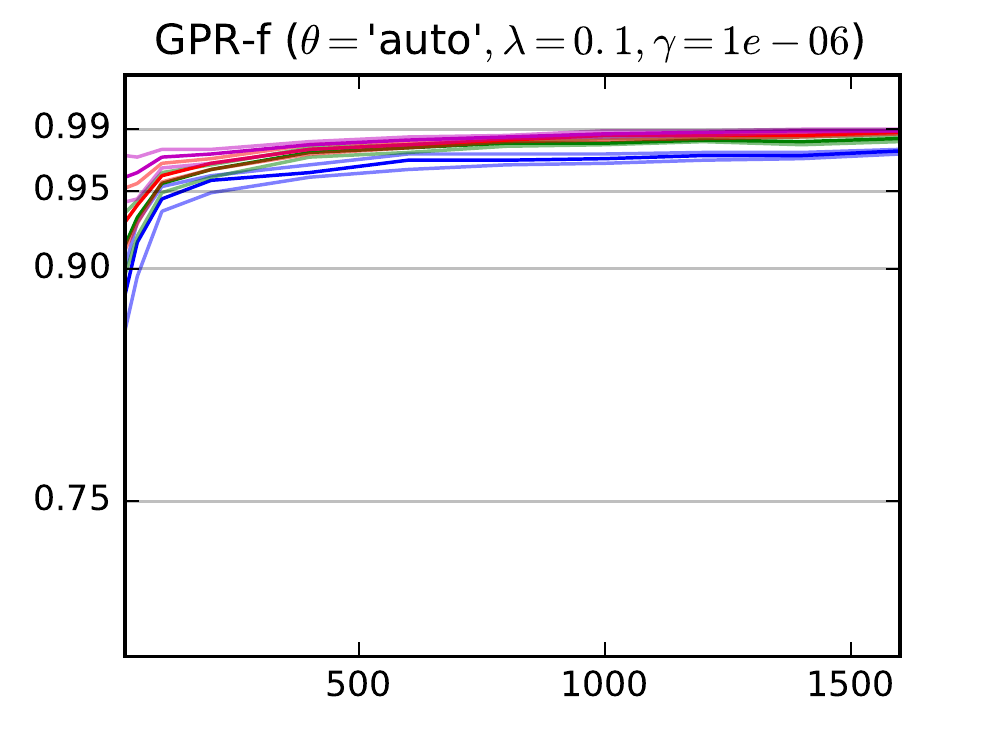}
  \end{subfigure}\\
  \begin{subfigure}[b]{0.25\linewidth}
    \includegraphics[width=0.95\linewidth]{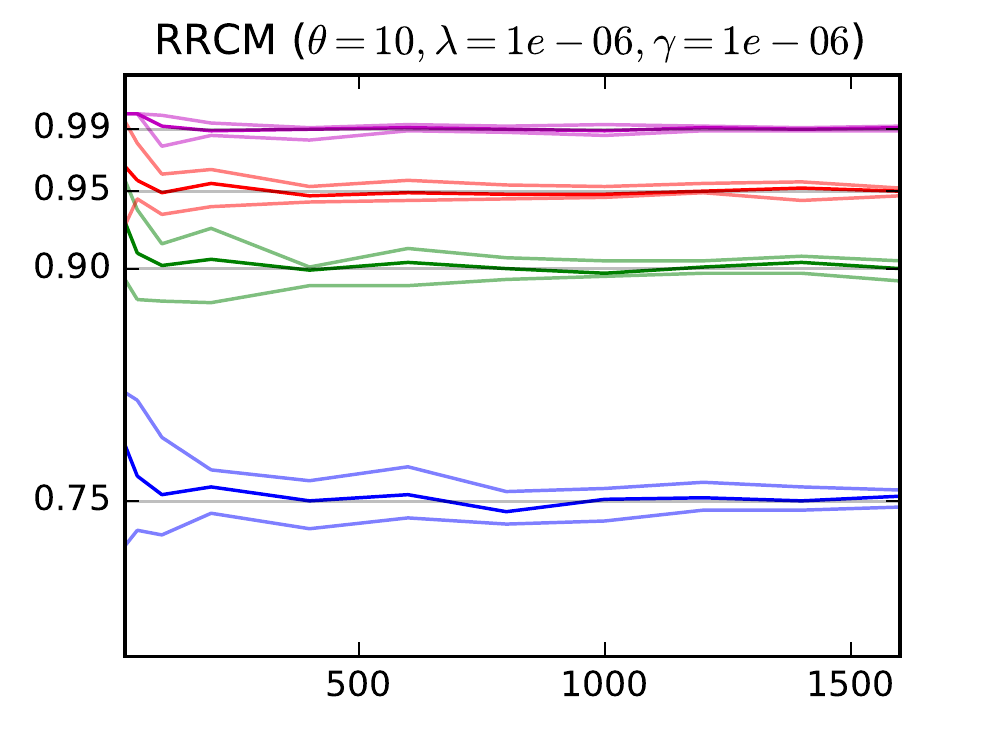}
  \end{subfigure}%
  \begin{subfigure}[b]{0.25\linewidth}
    \includegraphics[width=0.95\linewidth]{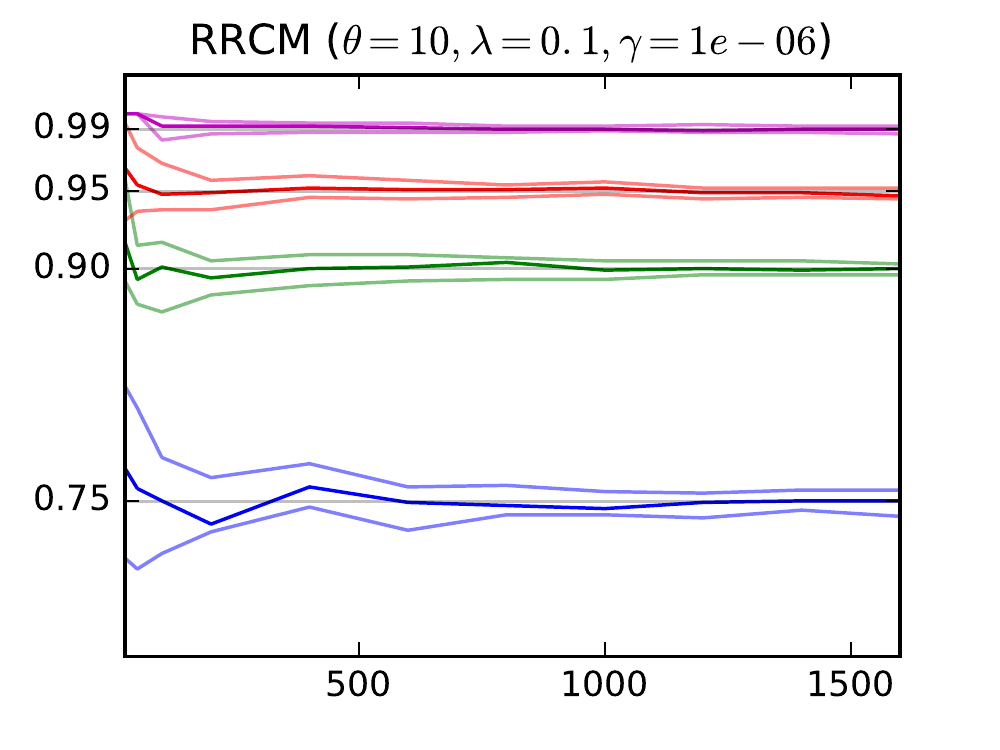}
  \end{subfigure}%
  \begin{subfigure}[b]{0.25\linewidth}
    \includegraphics[width=0.95\linewidth]{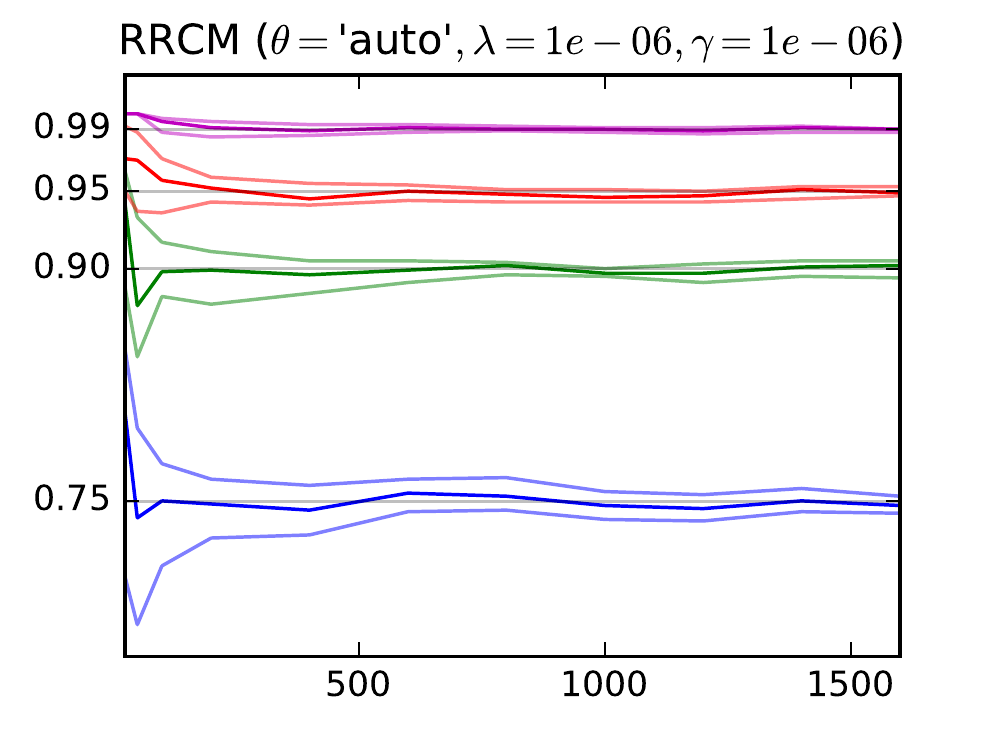}
  \end{subfigure}%
  \begin{subfigure}[b]{0.25\linewidth}
    \includegraphics[width=0.95\linewidth]{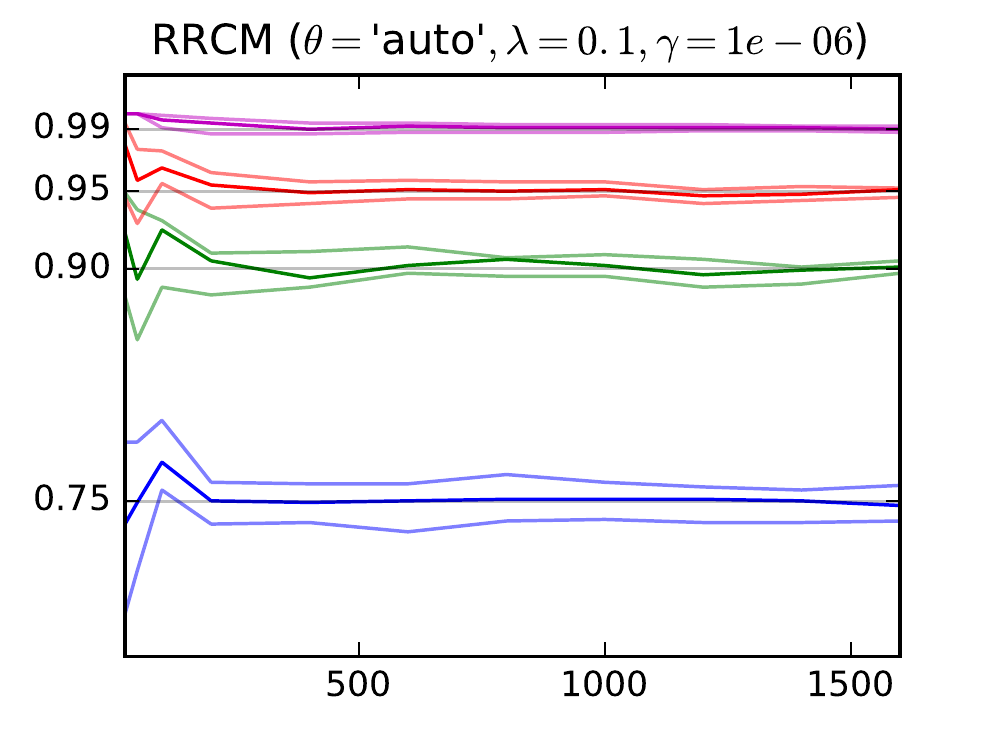}
  \end{subfigure}
  \caption{Coverage dynamics for the ``Heaviside'' ($\lambda=10^{-6}$).}
  \label{fig:heaviside_1d_low_noise}
\end{figure}

In the non-Gaussian noiseless experiments the coverage rate of conformal confidence
intervals maintains approaches the specified confidence levels and all conformal
procedures demonstrate very similar asymptotic validity. For the ``Heaviside''
step function typical confidence bands are shown in fig.~\ref{fig:nongauss_1d_heaviside},
and the asymptotic coverage rate of various confidence bands are presented in
fig.~\ref{fig:heaviside_1d_low_noise}.

In the non-Gaussian setting the GPR confidence intervals are not consistently valid,
as is evident from coverage rate dynamics for $\lambda=10^{-1}$. At the same time
conformal procedures show no significant departures from claimed validity (results
for other measures and residuals were qualitatively similar). The main conclusion
is that in the negligible noise case the conformal confidence intervals for the KRR
with the Gaussian kernel perform reasonably well both in terms of validity in a
non-Gaussian setting and efficiency in fully Gaussian setting.

\begin{figure}
  \centering
  \begin{subfigure}[b]{0.25\linewidth}
    \includegraphics[width=0.95\linewidth]{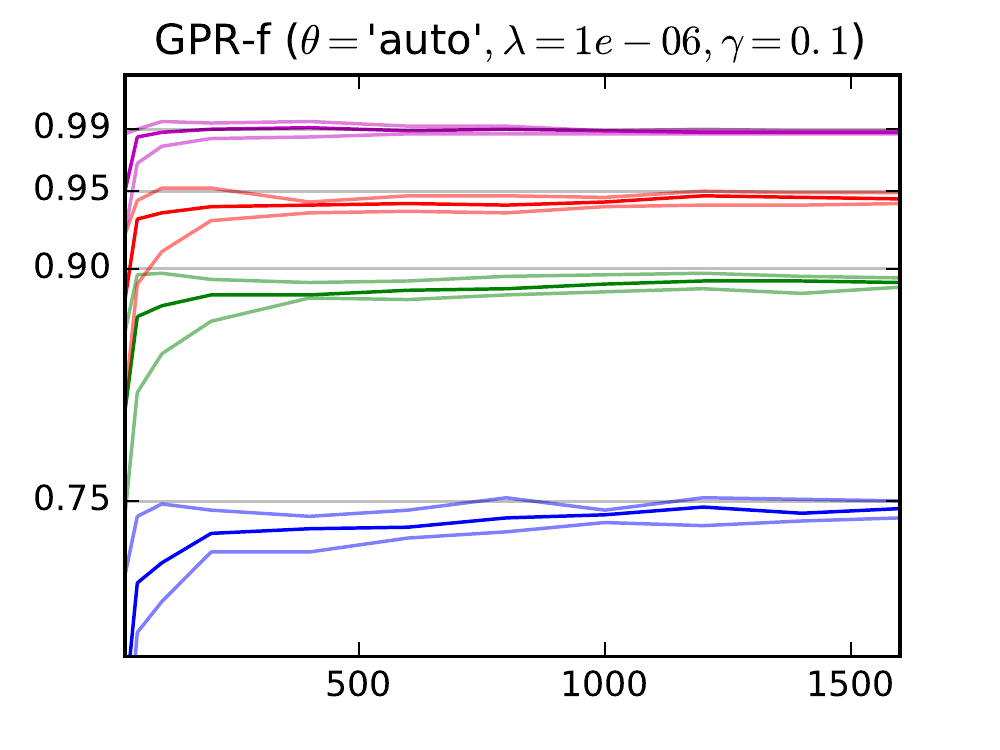}
  \end{subfigure}%
  \begin{subfigure}[b]{0.25\linewidth}
    \includegraphics[width=0.95\linewidth]{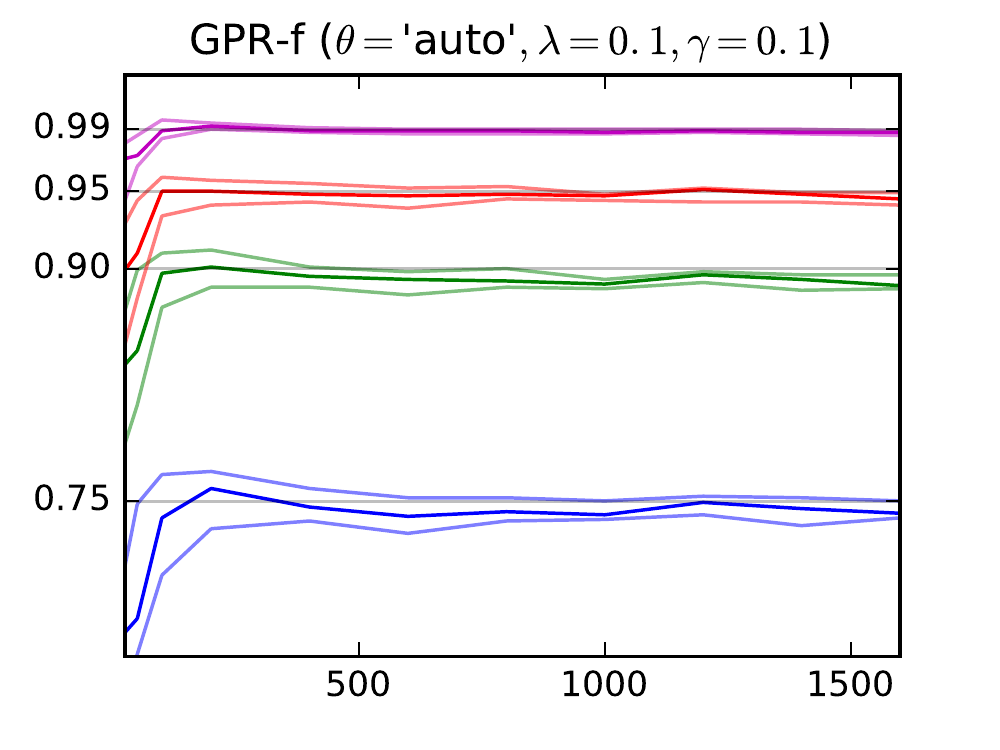}
  \end{subfigure}%
    \begin{subfigure}[b]{0.25\linewidth}
    \includegraphics[width=0.95\linewidth]{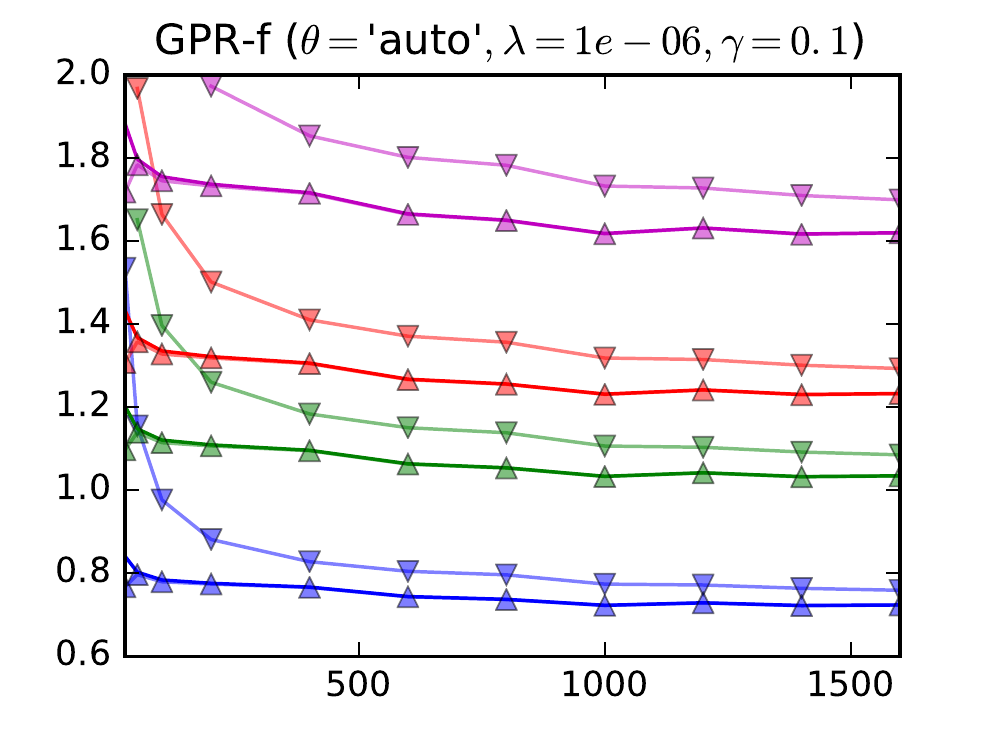}
  \end{subfigure}%
  \begin{subfigure}[b]{0.25\linewidth}
    \includegraphics[width=0.95\linewidth]{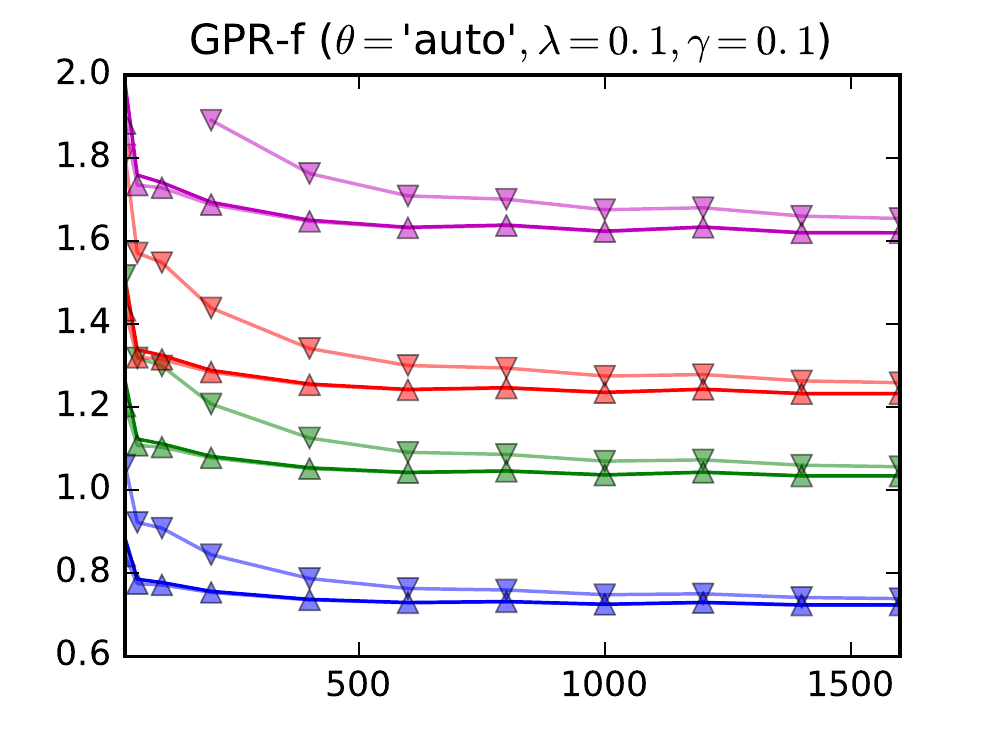}
  \end{subfigure}\\
  \begin{subfigure}[b]{0.25\linewidth}
    \includegraphics[width=0.95\linewidth]{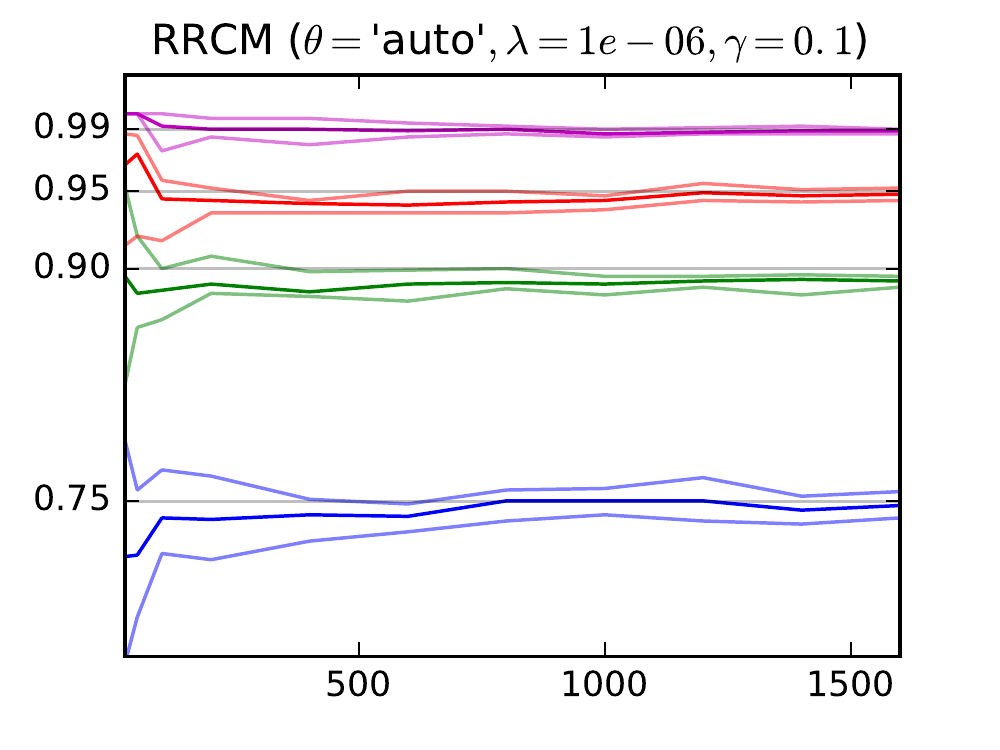}
  \end{subfigure}%
  \begin{subfigure}[b]{0.25\linewidth}
    \includegraphics[width=0.95\linewidth]{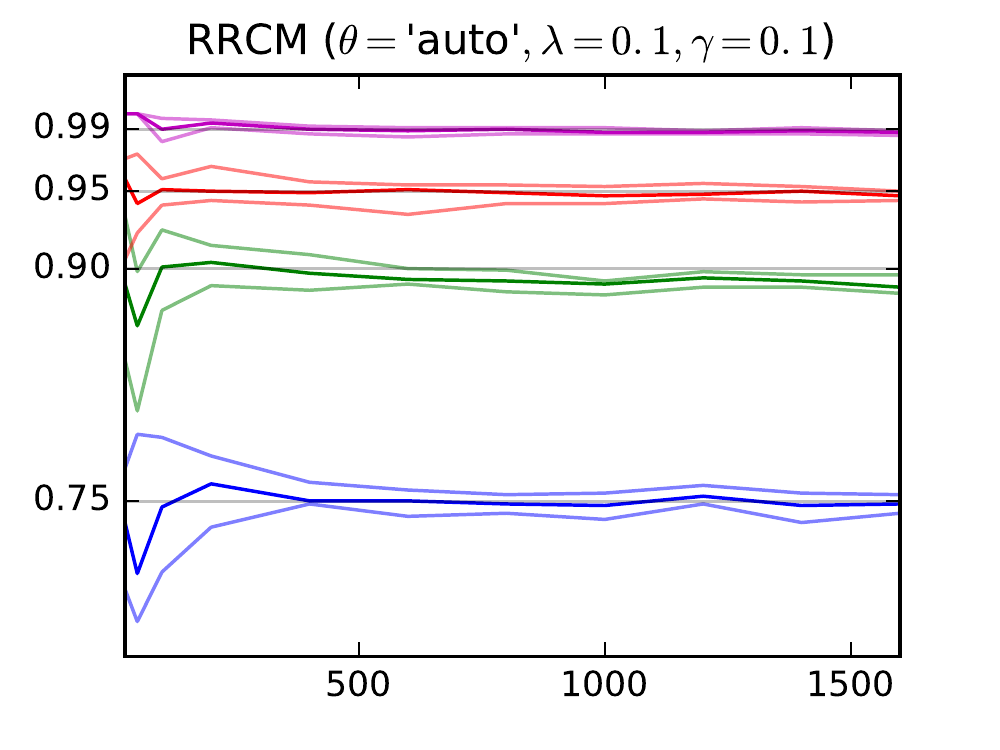}
  \end{subfigure}%
  \begin{subfigure}[b]{0.25\linewidth}
    \includegraphics[width=0.95\linewidth]{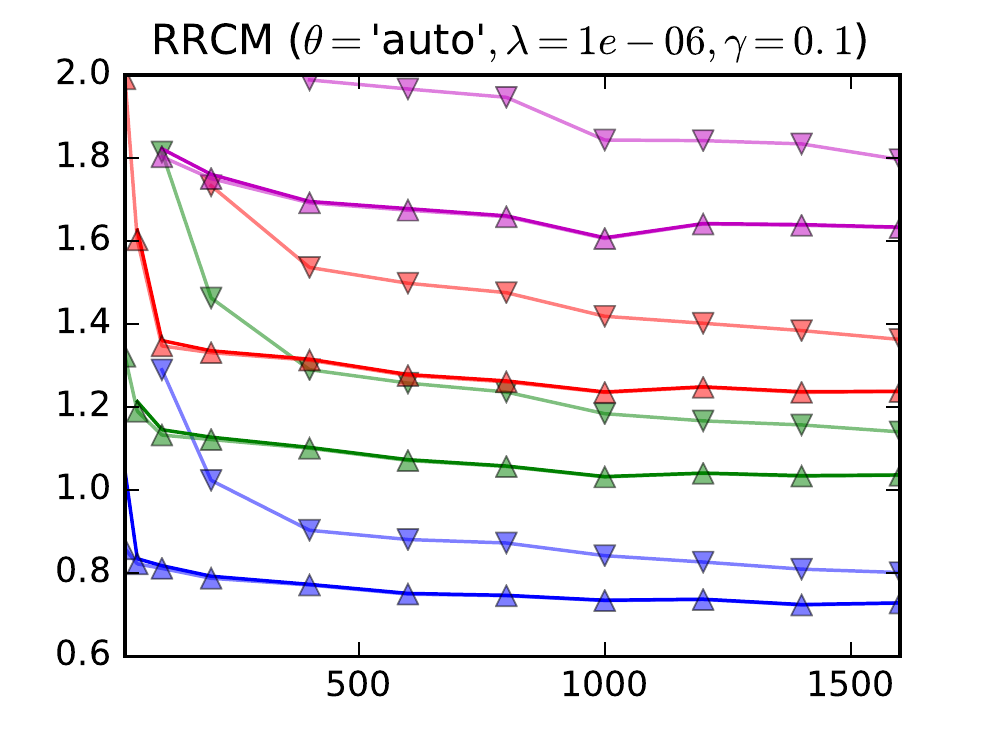}
  \end{subfigure}%
  \begin{subfigure}[b]{0.25\linewidth}
    \includegraphics[width=0.95\linewidth]{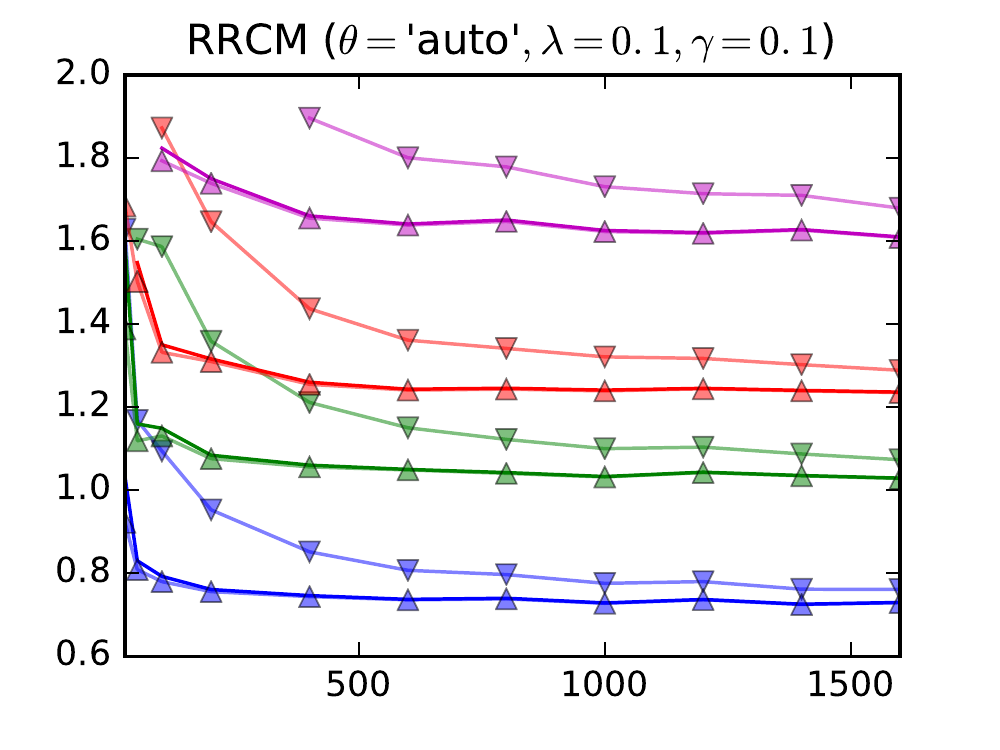}
  \end{subfigure}\\
  \begin{subfigure}[b]{0.25\linewidth}
    \includegraphics[width=0.95\linewidth]{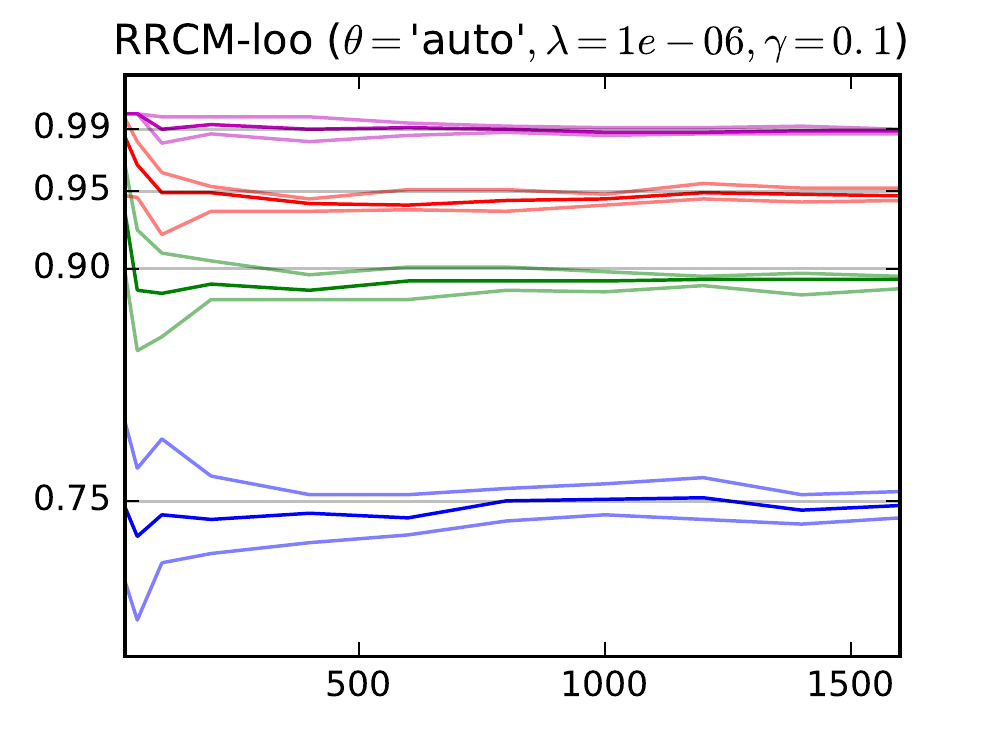}
  \end{subfigure}%
  \begin{subfigure}[b]{0.25\linewidth}
    \includegraphics[width=0.95\linewidth]{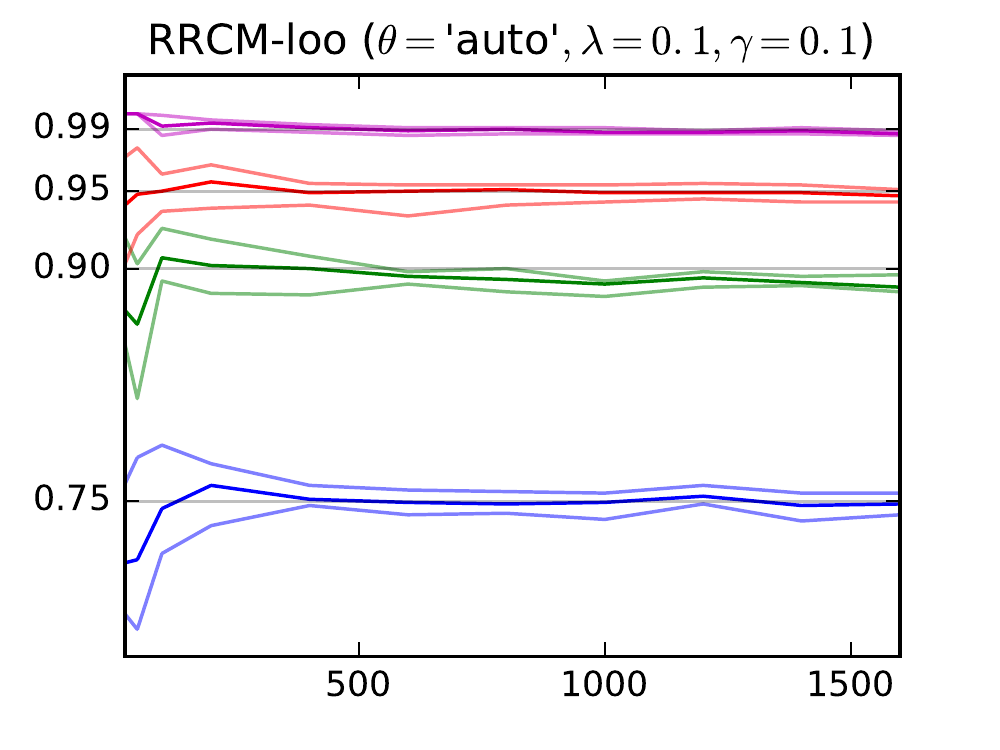}
  \end{subfigure}%
  \begin{subfigure}[b]{0.25\linewidth}
    \includegraphics[width=0.95\linewidth]{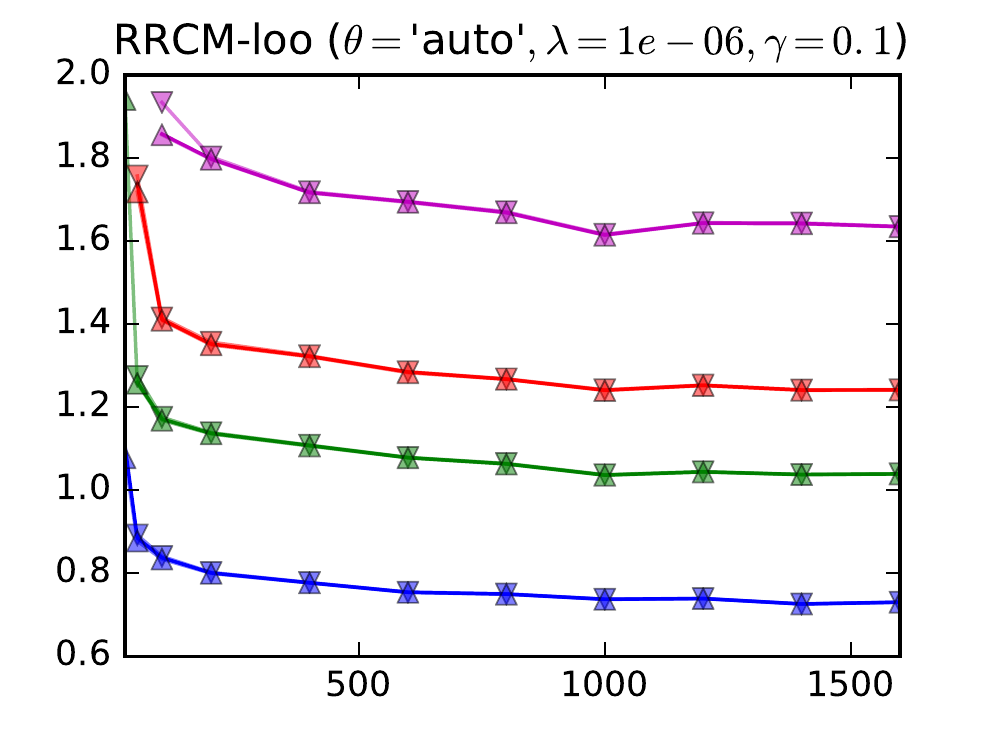}
  \end{subfigure}%
  \begin{subfigure}[b]{0.25\linewidth}
    \includegraphics[width=0.95\linewidth]{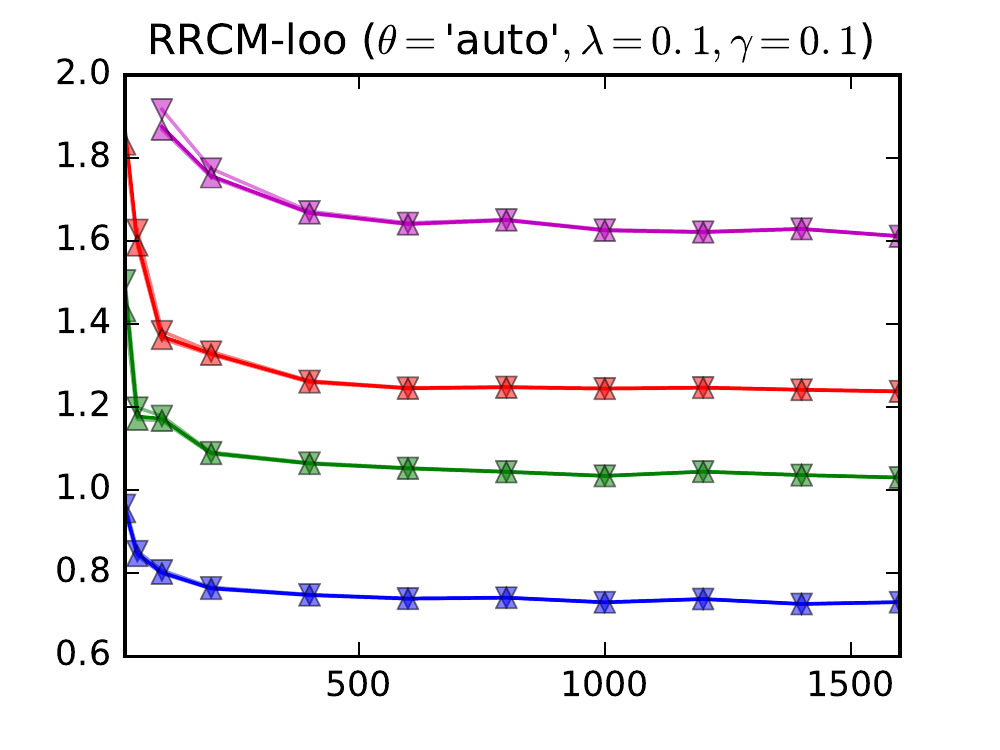}
  \end{subfigure}\\
  \begin{subfigure}[b]{0.25\linewidth}
    \includegraphics[width=0.95\linewidth]{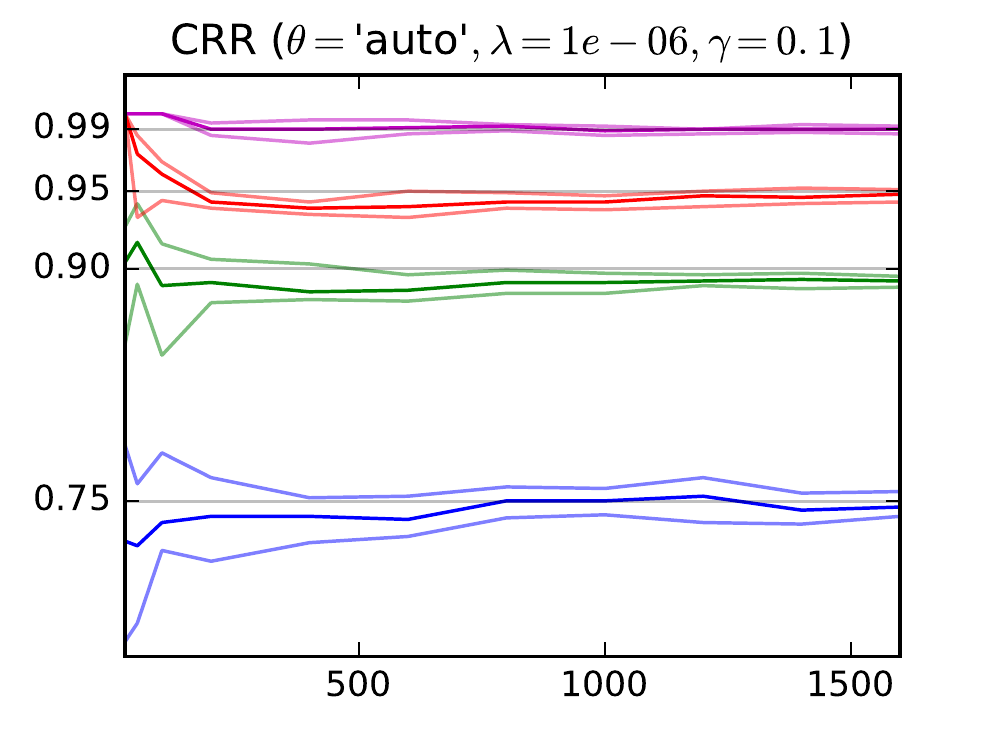}
  \end{subfigure}%
  \begin{subfigure}[b]{0.25\linewidth}
    \includegraphics[width=0.95\linewidth]{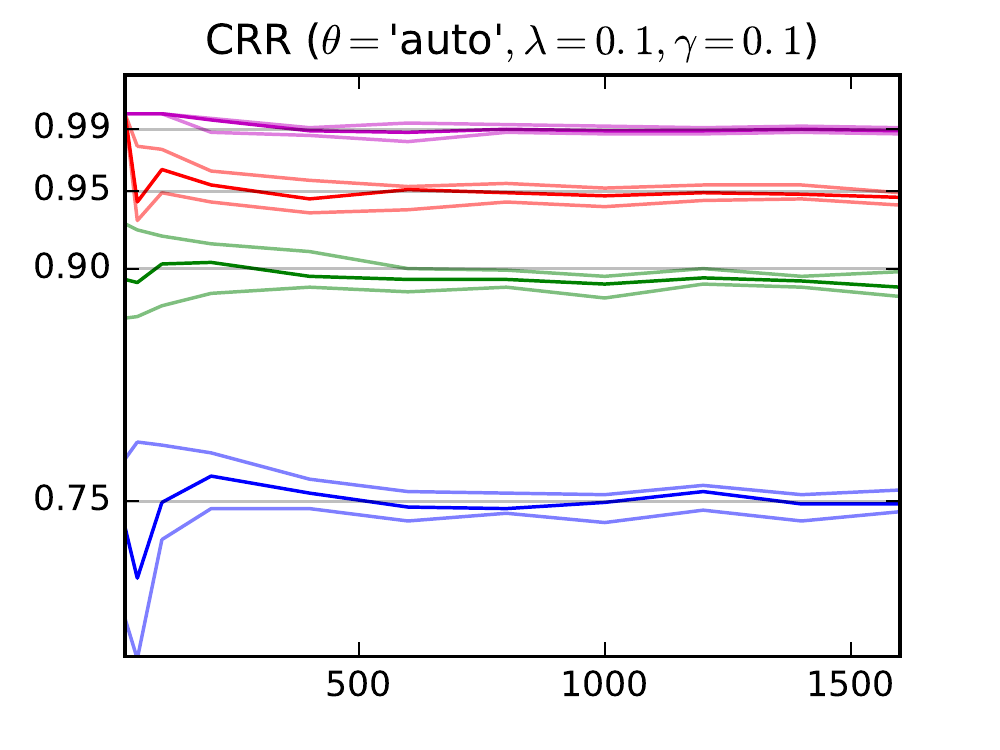}
  \end{subfigure}%
  \begin{subfigure}[b]{0.25\linewidth}
    \includegraphics[width=0.95\linewidth]{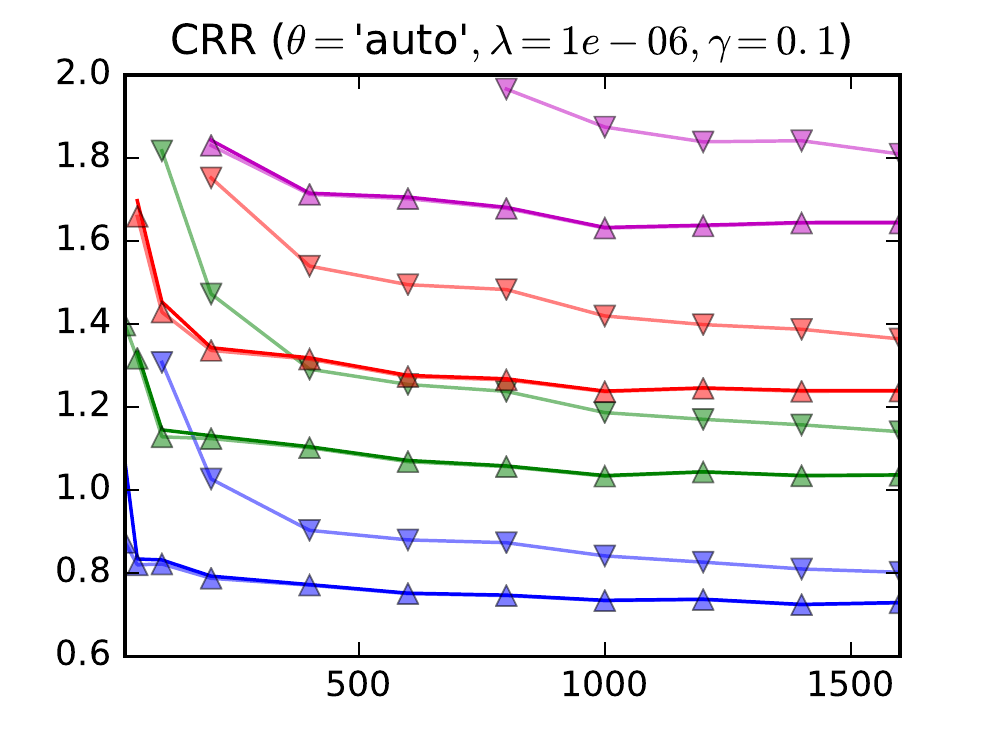}
  \end{subfigure}%
  \begin{subfigure}[b]{0.25\linewidth}
    \includegraphics[width=0.95\linewidth]{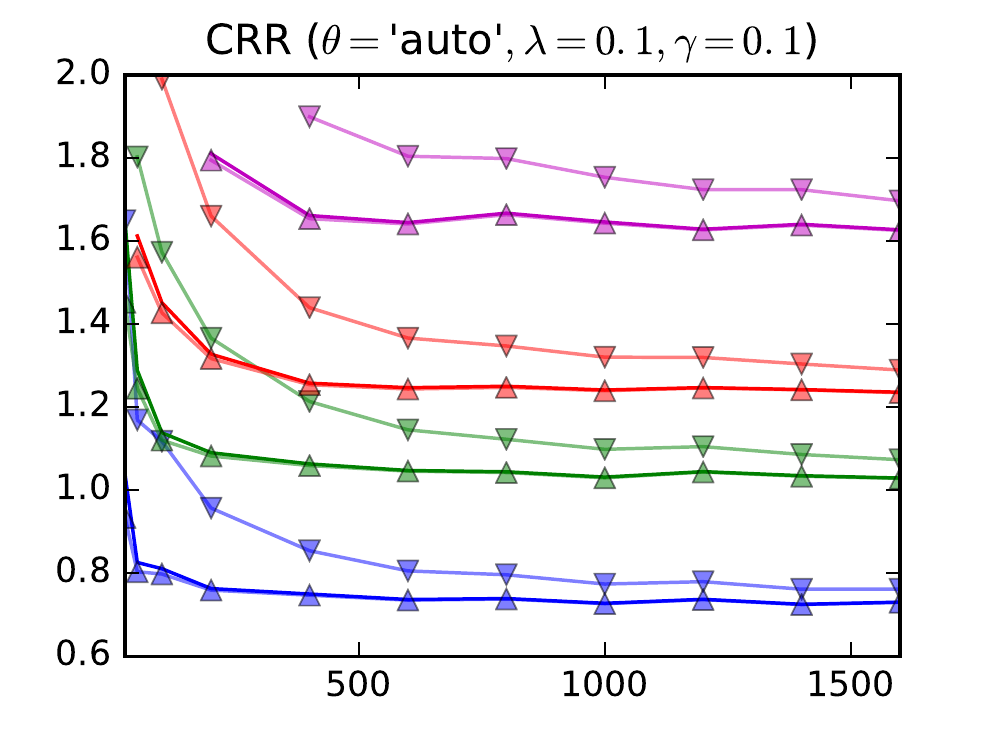}
  \end{subfigure}\\
  \begin{subfigure}[b]{0.25\linewidth}
    \includegraphics[width=0.95\linewidth]{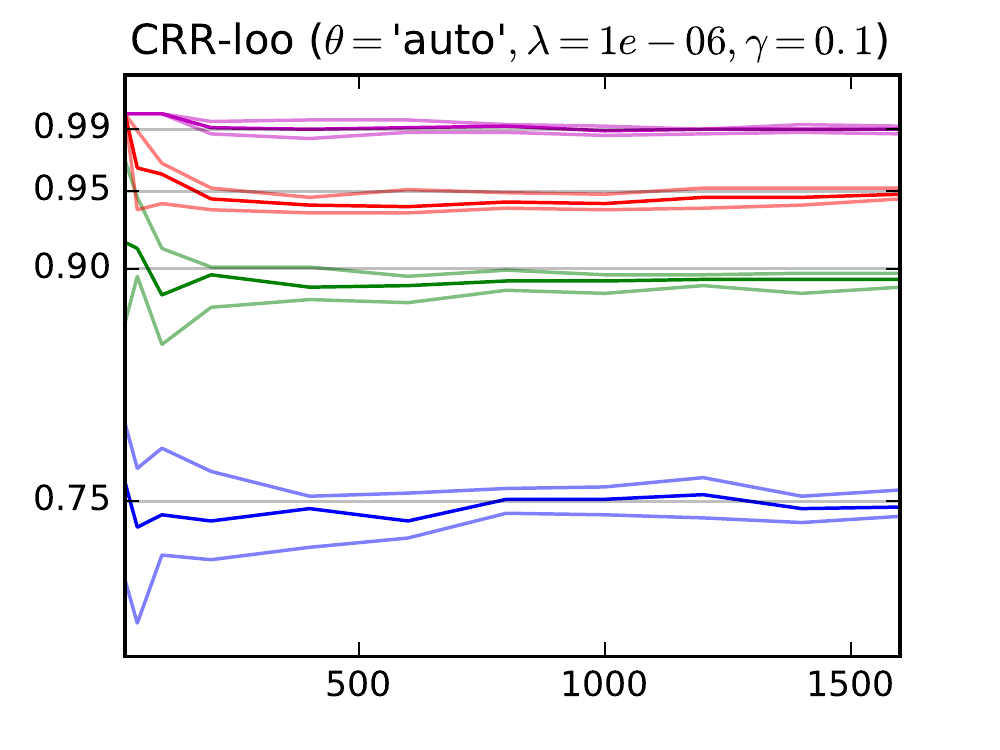}
    \caption{} \label{fig:gaussian_1d_high_noise_c3}
  \end{subfigure}%
  \begin{subfigure}[b]{0.25\linewidth}
    \includegraphics[width=0.95\linewidth]{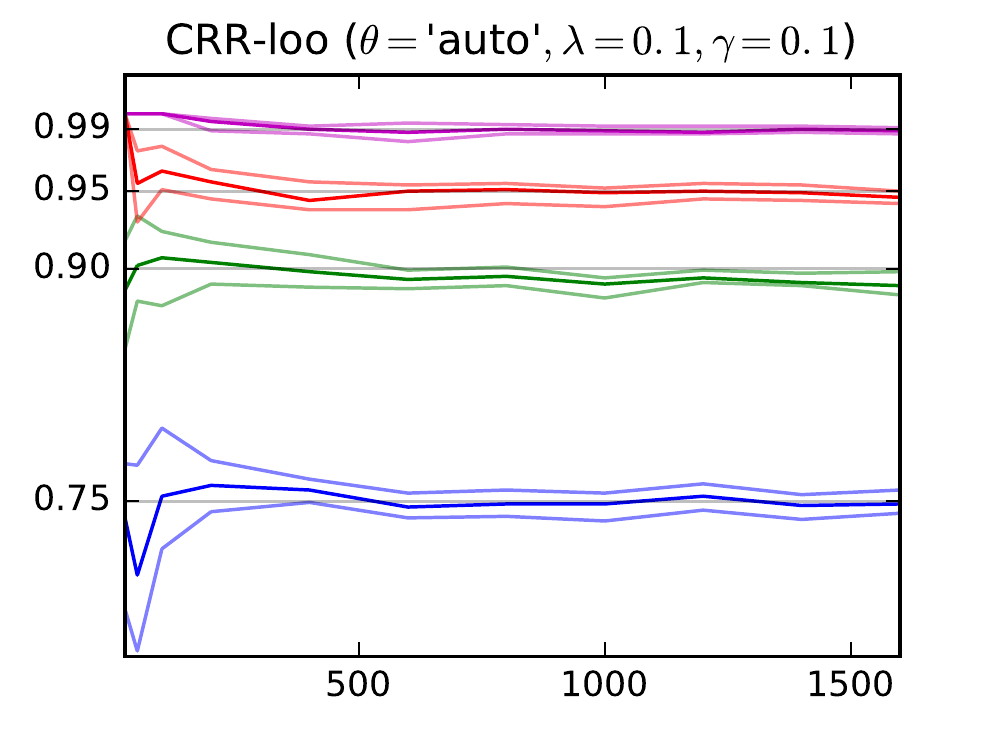}
    \caption{} \label{fig:gaussian_1d_high_noise_c4}
  \end{subfigure}%
  \begin{subfigure}[b]{0.25\linewidth}
    \includegraphics[width=0.95\linewidth]{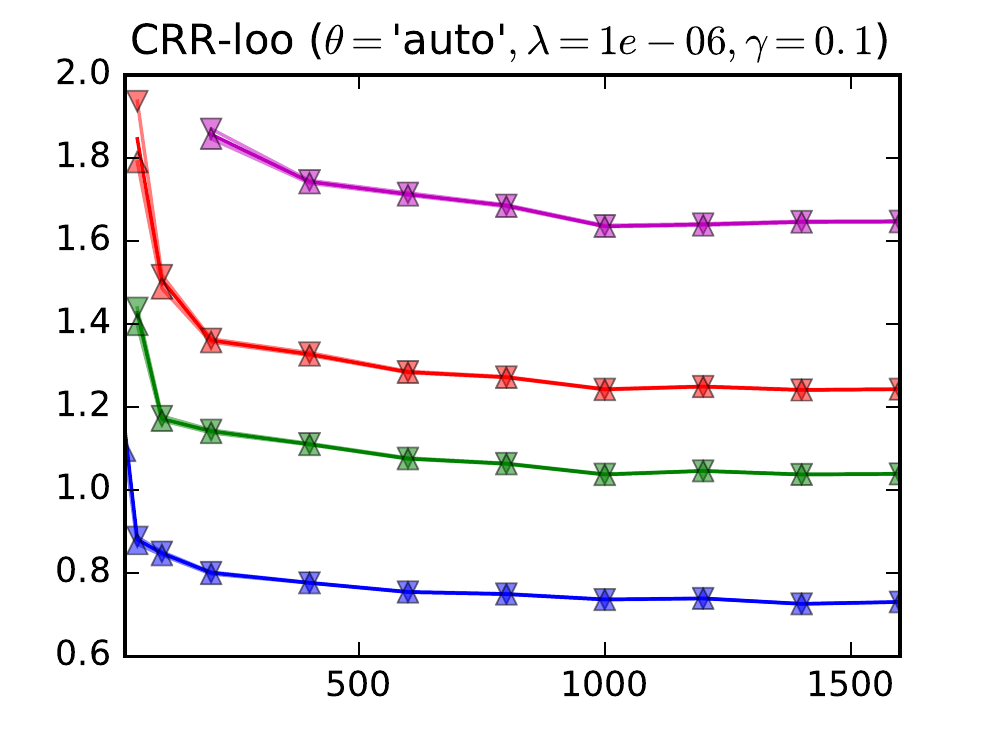}
    \caption{} \label{fig:gaussian_1d_high_noise_width_c3}
  \end{subfigure}%
  \begin{subfigure}[b]{0.25\linewidth}
    \includegraphics[width=0.95\linewidth]{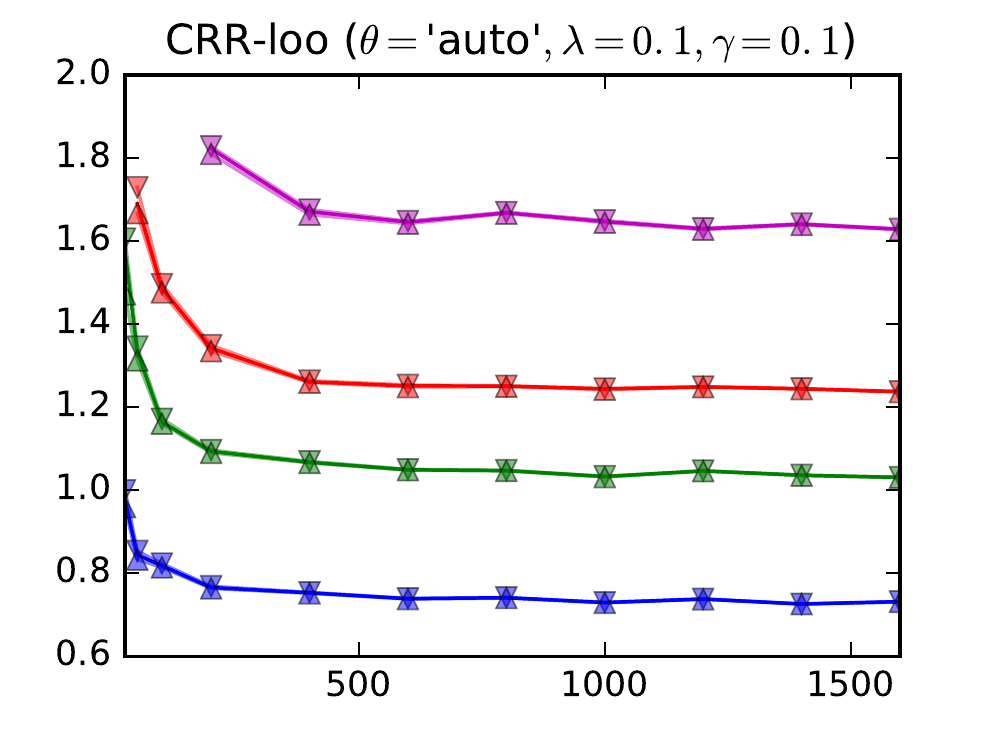}
    \caption{} \label{fig:gaussian_1d_high_noise_width_c4}
  \end{subfigure}
  \caption{Coverage rate and region size dynamics in the noisy fully Gaussian case
  with $\gamma=10^{-1}$ for different $\theta=\hat{\theta}_\text{ML}$ and $\lambda=10^{-6}$
  (\subref{fig:gaussian_1d_high_noise_c3}), and $\lambda=10^{-1}$ (\subref{fig:gaussian_1d_high_noise_c4}).
  Rows from \textit{top} to \textit{bottom}: ``GPR-f'', ``RRCM'', ``RRCM-loo'', ``CRR'', and ``CRR-loo''.}
  \label{fig:gaussian_1d_high_noise}
\end{figure}

The performance of conformal confidence regions in noisy setting ($\gamma=10^{-1}$)
is qualitatively similar to the negligible nose case, except that the bands are wider
due to higher observation noise. We report the findings for the MLE of $\theta$ only,
but the results for conformal regions with fixed $\theta$ are qualitatively similar.
In fig.~\ref{fig:gaussian_1d_high_noise} the conformal confidence regions provide the
specified level of validity regardless of the parameter $\lambda$ of the non-conformity
measure. As expected, the Bayesian confidence predictions uphold their theoretical
guarantees.

\begin{figure}
  \centering
  \begin{subfigure}[b]{0.25\linewidth}
    \includegraphics[width=0.95\linewidth]{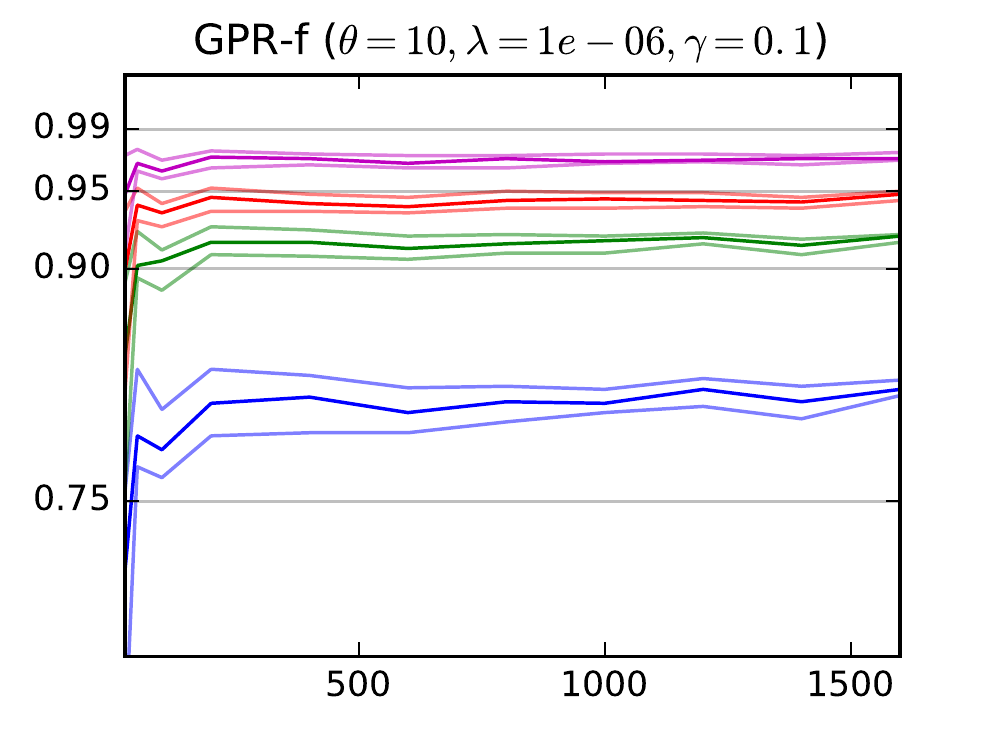}
  \end{subfigure}%
  \begin{subfigure}[b]{0.25\linewidth}
    \includegraphics[width=0.95\linewidth]{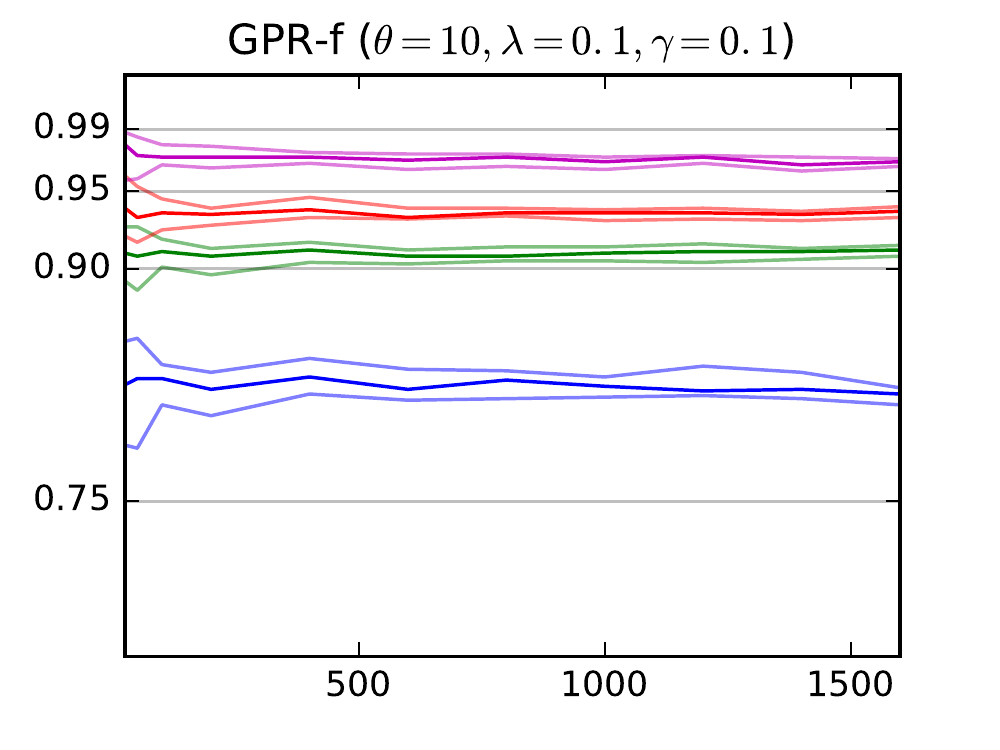}
  \end{subfigure}%
  \begin{subfigure}[b]{0.25\linewidth}
    \includegraphics[width=0.95\linewidth]{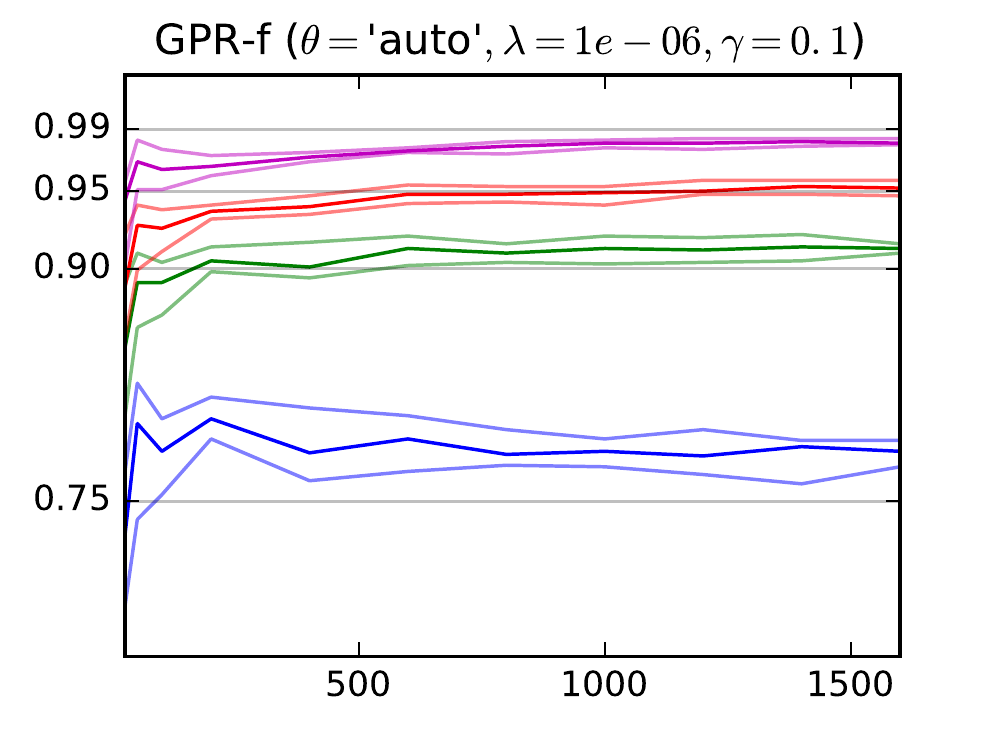}
  \end{subfigure}%
  \begin{subfigure}[b]{0.25\linewidth}
    \includegraphics[width=0.95\linewidth]{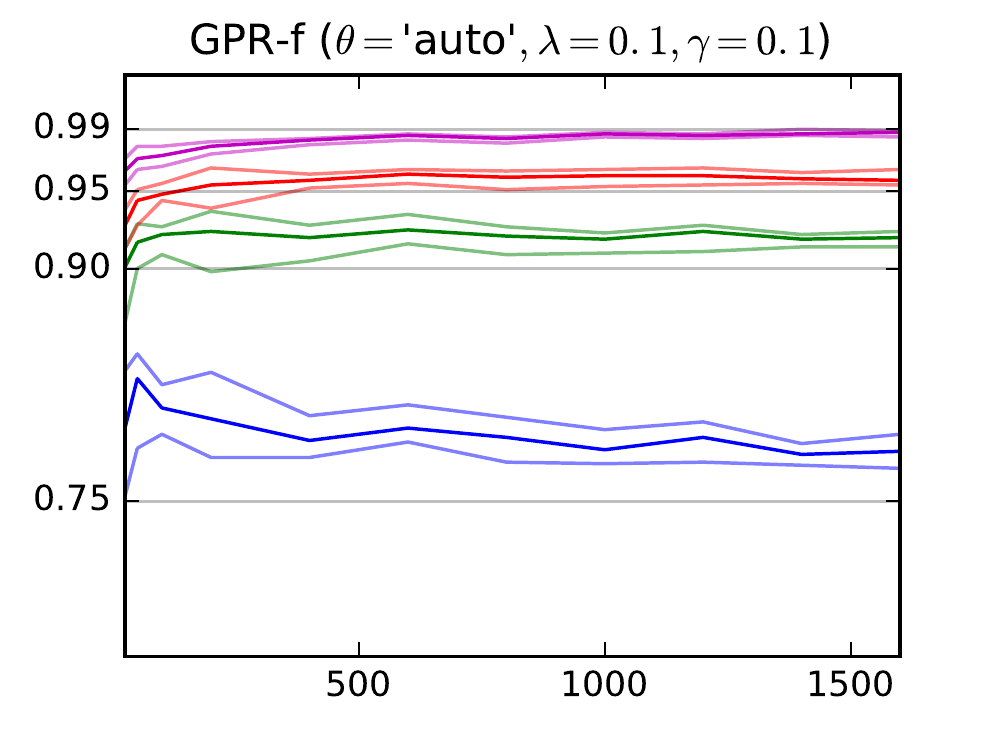}
  \end{subfigure}\\
  \begin{subfigure}[b]{0.25\linewidth}
    \includegraphics[width=0.95\linewidth]{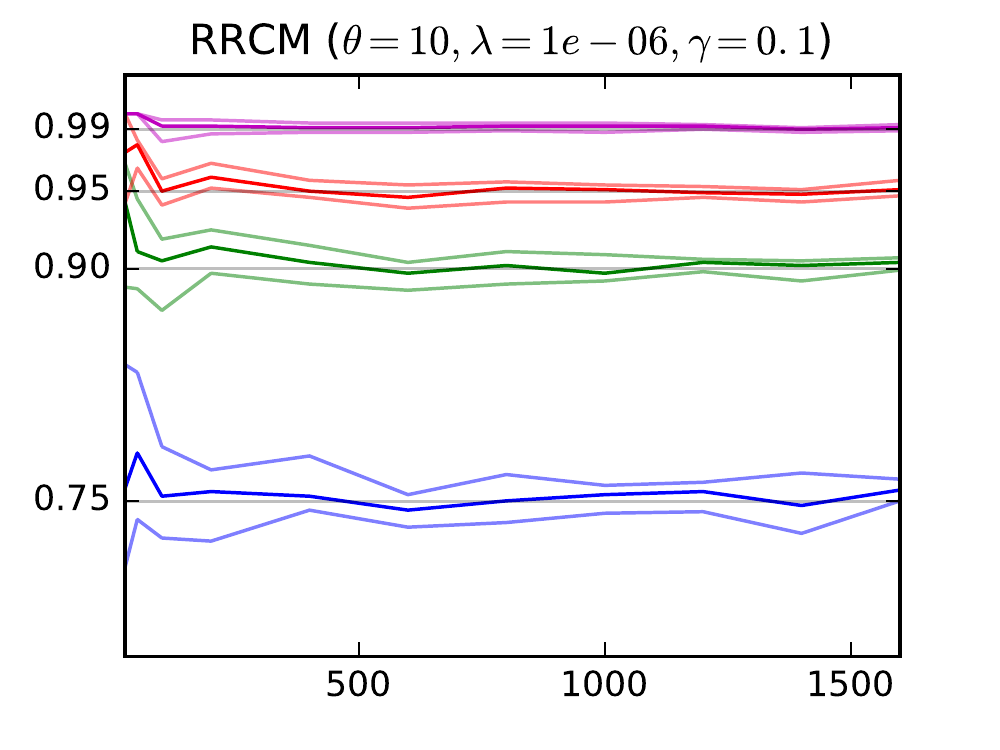}
  \end{subfigure}%
  \begin{subfigure}[b]{0.25\linewidth}
    \includegraphics[width=0.95\linewidth]{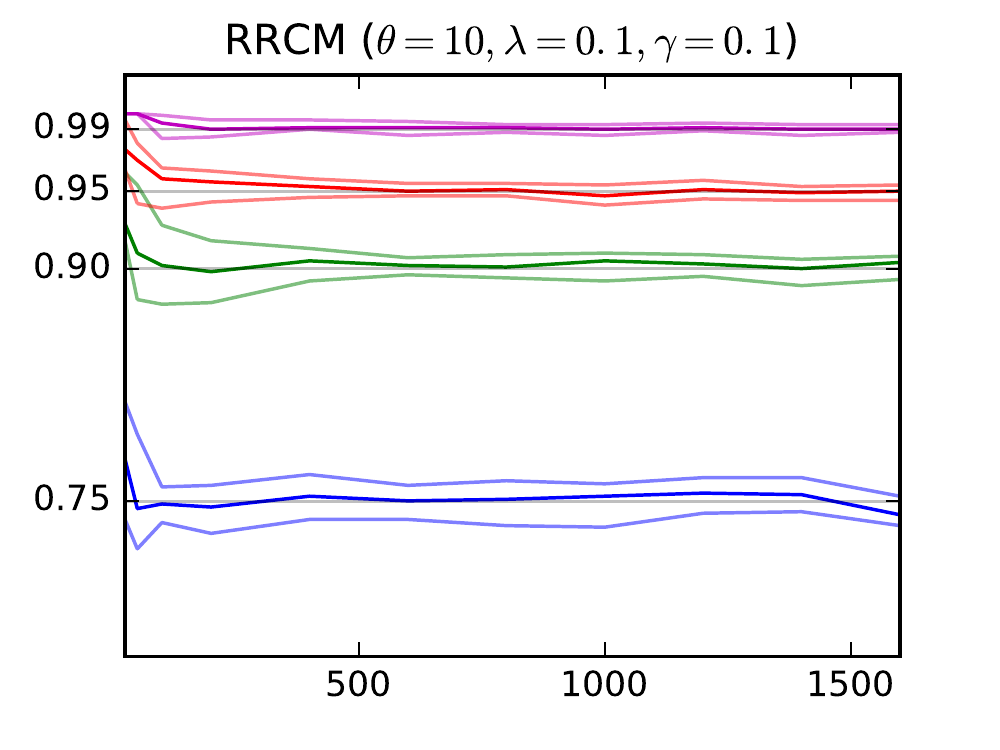}
  \end{subfigure}%
  \begin{subfigure}[b]{0.25\linewidth}
    \includegraphics[width=0.95\linewidth]{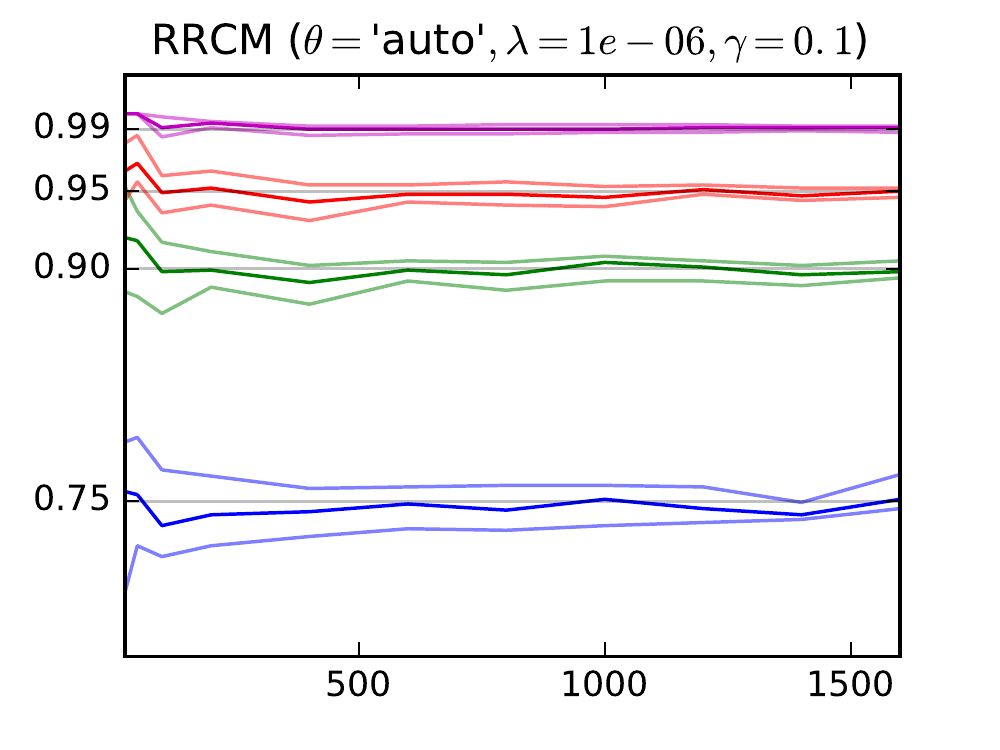}
  \end{subfigure}%
  \begin{subfigure}[b]{0.25\linewidth}
    \includegraphics[width=0.95\linewidth]{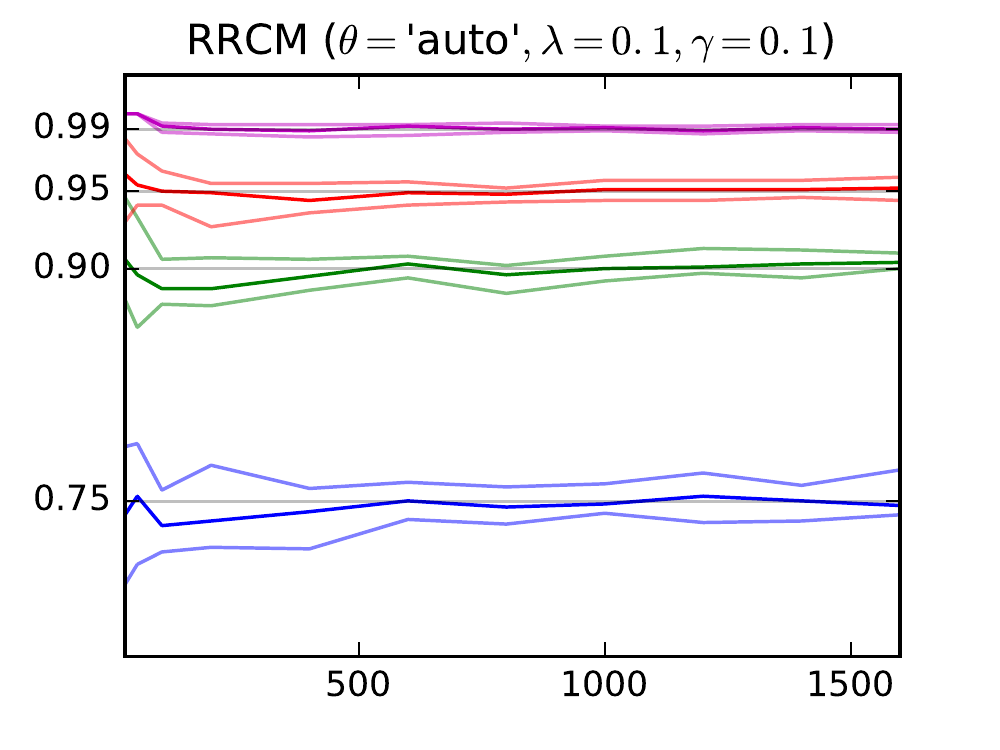}
  \end{subfigure}
  \caption{Coverage dynamics for the ``Heaviside'' ($\lambda=10^{-1}$).}
  \label{fig:heaviside_1d_high_noise}
\end{figure}

In the non-Gaussian setting with noise-to-signal ratio $\gamma=10^{-1}$, all experiments
yielded results similar to the negligible noise case: the conformal confidence sets are
asymptotically valid, whereas the Bayesian intervals are not, fig.~\ref{fig:heaviside_1d_high_noise}.

\subsection{Results: $2$-d} 
\label{sub:results_2_d}

\begin{figure}
  \centering
  \begin{subfigure}[b]{0.5\linewidth}
    \includegraphics[width=\linewidth]{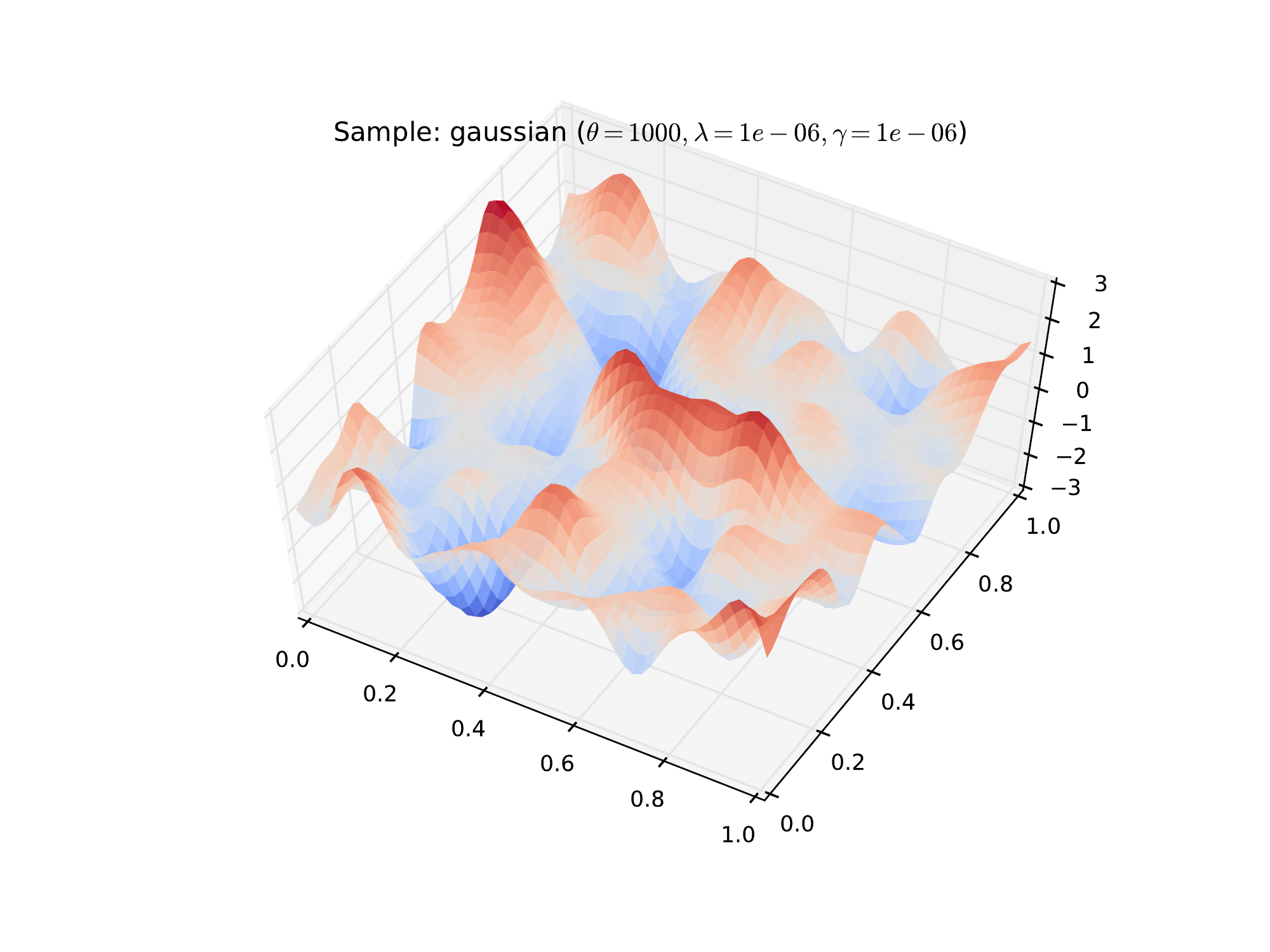}
  \end{subfigure}%
  \begin{subfigure}[b]{0.5\linewidth}
    \includegraphics[width=\linewidth]{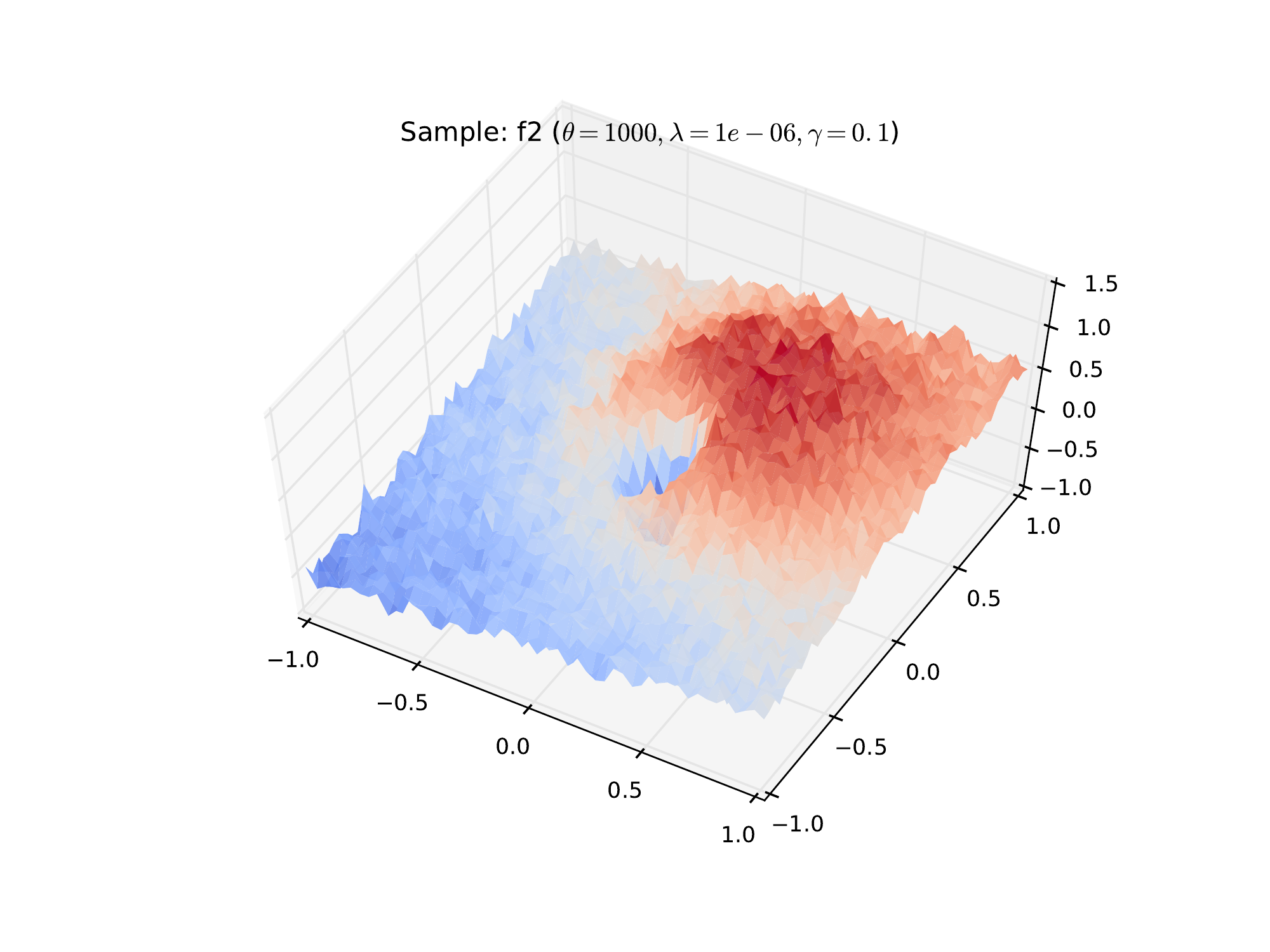}
  \end{subfigure}
  \caption{A sample path of a $2$-d Gaussian process(\textit{left} $\gamma=10^{-6}$)
  and a non-Gaussian function ``f2'' (\textit{right} $\gamma=10^{-1}$).}
  \label{fig:2d_profile}
\end{figure}

In this section we conduct experiments in the $2$-d setting $\Xcal=[-1,1]^2$, and
the experimental steps are similar to \ref{sub:results_1_d} The typical sample realisations
of the studied $2-d$ functions $f$ are depicted in fig.~\ref{fig:2d_profile}.

Table~\ref{tab:gaussian_2d_cov_gpr} shows the error rates ($y^*\notin\Bcal(x^*)$)
of the Bayesian confidence intervals on the fixed test sample. Columns 1 and 4 show
that the regions are approximately valid when the kernel and noise hyper-parameters
are known. The Bayesian intervals are more conservative for the case of low noise
($\gamma=10^{-6}$) and high regularization $\lambda=10^{-1}$. However, the validity
of the GPR confidence intervals is sensitive to misspecification of kernel precision
$\theta$.
\begin{table}
\centering
  \caption{The empirical error rate ($\%$) of the GPR confidence interval for simulated
  $2$-d Gaussian process with train size $n=1500$.}
  \label{tab:gaussian_2d_cov_gpr}
  \begin{tabular}{ll||rr|rr}
  \toprule
       & $\gamma$ & $10^{-6}$ &          & $10^{-1}$ &          \\\cline{2-2}
       & $\lambda$ & $10^{-6}$ & $10^{-1}$ & $10^{-6}$ & $10^{-1}$ \\\cline{2-2}
  $\theta$ & $\alpha(\%)$ &          &          &          &          \\
  \midrule
  $10^2$ & $1$ &     0.9 &     0.1 &     4.1 &     0.7 \\
       & $5$ &     4.5 &     0.5 &    12.0 &     4.4 \\
       & $10$ &     9.2 &     1.0 &    19.2 &     9.3 \\
       & $25$ &    23.8 &     3.3 &    36.1 &    24.5 \\\cline{2-6}
  $\hat{\theta}_\text{ML}$ & $1$ &     0.8 &     0.1 &     2.3 &     0.8 \\
       & $5$ &     4.4 &     0.5 &     4.9 &     4.5 \\
       & $10$ &     9.0 &     0.9 &     8.0 &     9.4 \\
       & $25$ &    23.6 &     2.3 &    18.9 &    24.7 \\
  \midrule
  $10^1$ & $1$ &      1.6 &      0.6 &      1.0 &      0.6 \\
       & $5$ &      5.3 &      4.2 &      5.2 &      3.7 \\
       & $10$ &      9.6 &      8.6 &     10.4 &      8.2 \\
       & $25$ &     22.5 &     23.4 &     26.3 &     22.6 \\\cline{2-6}
  $10^3$ & $1$ &      0.4 &      0.4 &      9.1 &      1.1 \\
       & $5$ &      1.2 &      1.2 &     13.1 &      4.9 \\
       & $10$ &      2.0 &      2.0 &     16.1 &      9.6 \\
       & $25$ &      4.7 &      4.5 &     24.0 &     24.0 \\
  \bottomrule
  \end{tabular}
\end{table}

In contrast to the GPR confidence intervals, the conformal regions are insensitive
to misspecification as demonstrated in table.~\ref{tab:gaussian_2d_cov_conf}, where
we show the maximal absolute deviation of the interval error rate from the specified
rate $\alpha$ across all studied significance levels (eq.~\ref{eq:mad_alpha}).
\begin{equation} \label{eq:mad_alpha}
  \mathtt{MAD}(\Gamma, A; \Theta)
  = \max_{\alpha\in A}\Bigl\lvert
    m^{-1}\#\{j\,:\, y^*_j\notin \Gamma^\alpha_n(X^*_j; \Theta)\} - \alpha
  \Bigr\rvert
    \,,
\end{equation}
where $\Theta$ is the vector of hyper-parameters of the experiment revealed to the
conformal procedure, $A = \{1\%, 5\%, 10\%, 25\%\}$, $(X^*_j, y^*_j)_{j=1}^{|X^*|}$
is the test sample.
\begin{table}
  \centering
  \caption{The maximal absolute deviation $\mathtt{MAD}(\Gamma, A; \Theta)$ ($\%$)
  of the empirical error rate from the theoretical significance level of conformal
  confidence regions for simulated $2$-d Gaussian process for $n=1500$.}
  \label{tab:gaussian_2d_cov_conf}
  \begin{tabular}{ll||rrrr}
  \toprule
       & $\gamma$ & $10^{-6}$ &          & $10^{-1}$ &          \\\cline{2-2}
       & $\lambda$ & $10^{-6}$ & $10^{-1}$ & $10^{-6}$ & $10^{-1}$ \\\cline{2-2}
  type & $\theta$ &          &          &          &          \\
  \midrule
  RRCM & $10^1$ &      0.3 &      0.2 &      0.8 &      0.4 \\
       & $10^2$ &      1.7 &      1.3 &      1.1 &      0.7 \\
       & $10^3$ &      1.1 &      2.6 &      1.2 &      0.5 \\
       & $\hat{\theta}_\text{ML}$ &      1.7 &      2.2 &      0.1 &      0.5 \\
  \midrule
  RRCM-loo & $10^1$ &      1.3 &      0.3 &      2.0 &      0.4 \\
       & $10^2$ &      2.4 &      1.7 &      2.0 &      0.4 \\
       & $10^3$ &      2.9 &      2.6 &      0.1 &      0.6 \\
       & $\hat{\theta}_\text{ML}$ &      2.6 &      2.6 &      0.8 &      0.6 \\
  \midrule
  CRR & $10^1$ &      0.3 &      0.1 &      0.8 &      0.3 \\
       & $10^2$ &      1.6 &      1.2 &      1.2 &      0.8 \\
       & $10^3$ &      0.8 &      2.2 &      1.3 &      0.5 \\
       & $\hat{\theta}_\text{ML}$ &      1.8 &      2.1 &      0.1 &      0.4 \\
  \midrule
  CRR-loo & $10^1$ &      1.2 &      0.3 &      2.0 &      0.2 \\
       & $10^2$ &      2.4 &      1.7 &      2.1 &      0.5 \\
       & $10^3$ &      2.6 &      2.4 &      0.3 &      0.5 \\
       & $\hat{\theta}_\text{ML}$ &      2.6 &      2.4 &      0.8 &      0.6 \\
  \bottomrule
  \end{tabular}
\end{table}

Typical profile of the test function used in non-Gaussian experiment is plotted
in fig.~\ref{fig:2d_profile} (p.~\pageref{fig:2d_profile}). The performance of the
conformal regions in the non-Gaussian experiments are summarized in tab.~\ref{tab:nongaussian_f2_2d_cov_conf}.
Overall, the error rates do not stray too far from the stated levels, and conformal
regions are weakly sensitive to the KRR hyper-parameters. In contrast, tab.~\ref{tab:nongaussian_f2_2d_cov_gpr}
shows that the empirical error rate of Bayesian confidence intervals depends on the
values of the precision parameter. The MLE $\theta$ produces conservatively valid
confidence intervals, and for $\gamma=10^{-1}$ the error rate becomes closer to
the specified significance level.

\begin{table}
  \centering
  \caption{The maximal absolute deviation $\mathtt{MAD}(\Gamma, A; \Theta)$ ($\%$)
  of the empirical error rate from the theoretical significance level of conformal
  confidence regions for the ``f2'' test function ($n=1500$).}
  \label{tab:nongaussian_f2_2d_cov_conf}
  \begin{tabular}{ll||rrrr}
  \toprule
       & $\gamma$ & $10^{-6}$ &          & $10^{-1}$ &          \\\cline{2-2}
       & $\lambda$ & $10^{-6}$ & $10^{-1}$ & $10^{-6}$ & $10^{-1}$ \\\cline{2-2}
  type & $\theta$ &          &          &          &          \\
  \midrule
  CRR & $10^1$ &      1.0 &      1.3 &      1.2 &      0.2 \\
       & $10^2$ &      0.7 &      2.7 &      1.4 &      0.6 \\
       & $10^3$ &      0.3 &      1.1 &      1.0 &      0.1 \\
       & $\hat{\theta}_\text{ML}$ &      1.4 &      2.2 &      0.6 &      0.5 \\
  \midrule
  CRR-loo & $10^1$ &      0.8 &      1.4 &      1.3 &      0.2 \\
       & $10^2$ &      3.0 &      3.2 &      1.9 &      0.5 \\
       & $10^3$ &      1.0 &      1.0 &      0.2 &      0.2 \\
       & $\hat{\theta}_\text{ML}$ &      2.6 &      2.6 &      0.5 &      0.2 \\
  \midrule
  RRCM & $10^1$ &      0.9 &      1.4 &      1.2 &      0.2 \\
       & $10^2$ &      0.7 &      2.6 &      1.3 &      0.6 \\
       & $10^3$ &      0.4 &      0.5 &      0.9 &      0.3 \\
       & $\hat{\theta}_\text{ML}$ &      1.3 &      2.3 &      0.8 &      0.4 \\
  \midrule
  RRCM-loo & $10^1$ &      0.8 &      1.6 &      1.2 &      0.3 \\
       & $10^2$ &      3.0 &      3.2 &      2.0 &      0.6 \\
       & $10^3$ &      0.9 &      0.6 &      0.2 &      0.2 \\
       & $\hat{\theta}_\text{ML}$ &      2.6 &      2.6 &      0.7 &      0.1 \\
  \bottomrule
  \end{tabular}
\end{table}

\begin{table}
  \centering
  \caption{The empirical error rate ($\%$) of the GPR confidence interval for the
  ``f2'' test function ($n=1500$).}
  \label{tab:nongaussian_f2_2d_cov_gpr}
  \begin{tabular}{ll||rrrr}
  \toprule
       & $\gamma$ & $10^{-6}$ &          & $10^{-1}$ &          \\\cline{2-2}
       & $\lambda$ & $10^{-6}$ & $10^{-1}$ & $10^{-6}$ & $10^{-1}$ \\\cline{2-2}
  $\theta$ & $\alpha(\%)$ &          &          &          &          \\
  \midrule
   $10^1$ & $1\%$ &      2.3 &      1.9 &      2.2 &      1.1 \\
        & $5\%$ &      3.2 &      3.0 &      7.9 &      4.9 \\
        & $10\%$ &      4.0 &      3.7 &     13.8 &      9.7 \\
        & $25\%$ &      5.8 &      5.6 &     29.9 &     24.3 \\
  \midrule
   $10^2$ & $1\%$ &      0.3 &      0.0 &     19.1 &      1.3 \\
        & $5\%$ &      0.6 &      0.1 &     28.6 &      5.9 \\
        & $10\%$ &      0.9 &      0.2 &     35.0 &     11.1 \\
        & $25\%$ &      2.6 &      1.2 &     48.3 &     26.5 \\
  \midrule
   $10^3$ & $1\%$ &      0.1 &      0.1 &      2.4 &      0.1 \\
        & $5\%$ &      1.9 &      2.1 &      4.0 &      1.9 \\
        & $10\%$ &      4.1 &      4.5 &      6.0 &      5.0 \\
        & $25\%$ &     12.9 &     13.4 &     13.9 &     16.4 \\
  \midrule
   $\hat{\theta}_\text{ML}$ & $1\%$ &      3.4 &      0.6 &      2.0 &      1.1 \\
        & $5\%$ &      4.4 &      1.1 &      3.9 &      5.1 \\
        & $10\%$ &      5.2 &      1.4 &      6.9 &     10.0 \\
        & $25\%$ &      7.1 &      2.4 &     18.1 &     24.9 \\
  \bottomrule
  \end{tabular}
\end{table}



\section{Conclusion} 
\label{sec:conclusion}

Experiments in sec.~\ref{sec:numerical_study} provide evidence suggesting that
conformal procedures are insensitive to the choice of the core NCM and are asymptotically
equivalent in terms of coverage and efficiency, despite being applied in the off-line
batch learning setting. Furthermore, the results indicate that both Bayesian and
conformal confidence intervals possess the asymptotic validity guarantees, when
the Gaussian assumptions hold, and the conformal procedure yields asymptotically
efficient regions.

Further research shall focus on establishing theoretical foundations for the obtained
experimental results for the KRR with Gaussian kernel, or isolating special cases
when it holds, and studying the cases when it fails, and generalizing the efficiency
result in \cite{burnaevV14}.


\section*{Acknowledgements} 
\label{sec:acknowledgements}
\noindent The research, presented in Section \ref{sec:numerical_study} of this paper, was supported by the RFBR grants 16-01-00576 A and 16-29-09649 ofi\_m; the research, presented in other sections, was conducted in IITP RAS and supported solely by the Russian Science Foundation grant (project 14-50-00150).


\bibliographystyle{IEEEtran}
\bibliography{references,references_icmla}

\end{document}